\documentclass{article}

\PassOptionsToPackage{numbers, compress}{natbib}

\usepackage[final]{neurips_2025}

\usepackage[utf8]{inputenc} 
\usepackage[T1]{fontenc}    
\usepackage{hyperref}       
\usepackage{url}            
\usepackage{booktabs}       
\usepackage{amsfonts}       
\usepackage{nicefrac}       
\usepackage{microtype}      
\usepackage{xcolor}         

\usepackage{amsmath}
\usepackage{amssymb}
\usepackage{mathtools}
\usepackage{amsthm}
\usepackage{mathabx}
\usepackage{bbm}
\usepackage{mathrsfs}
\usepackage{blkarray}
\usepackage{bm}
\usepackage{cancel}

\usepackage{abbreviation_definitions}

\usepackage{enumerate}
\usepackage{enumitem}

\usepackage{caption}
\usepackage{graphicx}
\usepackage{subcaption}
\usepackage{multirow}
\usepackage{wrapfig}

\usepackage{graphicx}
\usepackage{mwe}

\usepackage{tikz}
\usetikzlibrary{shapes,arrows,positioning,fit,decorations.pathreplacing,calc}

\usepackage{todonotes}

\usepackage{algorithm}
\usepackage{algpseudocode}

\usepackage[toc,page,header]{appendix}
\usepackage{minitoc}

\title{Preference-based Reinforcement Learning beyond\\Pairwise Comparisons: Benefits of Multiple Options}

\author{%
  Joongkyu Lee \\
  Seoul National University\\
  \texttt{jklee0717@snu.ac.kr} \\
  \And
  Seouh-won Yi \\
  Seoul National University\\
  \texttt{uniqueseouh@snu.ac.kr} \\
  \And
  Min-hwan Oh \\
  Seoul National University\\
  \texttt{minoh@snu.ac.kr} \\
}

\doparttoc 
\faketableofcontents 

\begin{document}

\maketitle

\begin{abstract}
    We study online preference-based reinforcement learning (PbRL) with the goal of improving sample efficiency.
    While a growing body of theoretical work has emerged—motivated by PbRL’s recent empirical success, particularly in aligning large language models (LLMs)—most existing studies focus only on pairwise comparisons.
    A few recent works~\citep{zhu2023principled, mukherjee2024optimal, thekumparampil2024comparing} have explored using multiple comparisons and ranking feedback, but their performance guarantees fail to improve—and can even deteriorate—as the feedback length increases, despite the richer information available.
    To address this gap, we adopt the Plackett–Luce (PL) model for ranking feedback over action subsets and propose \AlgName{}, an algorithm that selects multiple actions by maximizing the average uncertainty within the offered subset.
    We prove that \AlgName{} achieves a suboptimality gap of $\tilde{\mathcal{O}}\!\left( \frac{d}{T}\! \sqrt{ \sum_{t=1}^T \frac{1}{|S_t|}} \right)$, where $T$ is the total number of rounds, $d$ is the feature dimension, and $|S_t|$ is the size of the subset at round $t$.
    This result shows that larger subsets directly lead to improved performance and, notably, the bound avoids the exponential dependence on the unknown parameter’s norm, which was a fundamental limitation in most previous works.
    Moreover, we establish a near-matching lower bound of $\Omega \Big( \frac{d}{K \sqrt{T}} \Big)$, where $K$ is the maximum subset size.
    To the best of our knowledge, this is the first theoretical result in PbRL with ranking feedback that explicitly shows improved sample efficiency as a function of the subset size.
\end{abstract}

\section{Introduction}
\label{sec:Introduction}
The framework of \textit{Preference-based Reinforcement Learning} (PbRL)~\citep{busa2014preference, wirth2016model, wirth2017survey, saha2023dueling} was introduced to address the difficulty of designing effective reward functions, which often demands substantial and complex engineering effort~\citep{wirth2013preference, wirth2017survey}.
PbRL has been successfully applied in diverse domains, including robot training, stock prediction, recommender systems, and clinical trials~\citep{jain2013learning, sadigh2017active, christiano2017deep, kupcsik2017learning, novoseller2020dueling}.
Notably, PbRL also serves as a foundational framework for Reinforcement Learning from Human Feedback (RLHF) when feedback is provided in the form of preferences rather than explicit scalar rewards. 
This preference-based approach has proven highly effective in aligning Large Language Models (LLMs) with human values and preferences~\citep{christiano2017deep, ouyang2022training, rafailov2023direct}.

Given its practical success, the field has also seen significant theoretical advances~\citep{chen2022human, mehta2023sample, saha2023dueling, zhu2023principled, xiong2023iterative, zhan2023provable, zhan2023provableoffline, wu2023making, sekhari2023contextual, munos2023nash, cen2024value,  chowdhury2024provably, dong2024rlhf, das2024active, mukherjee2024optimal, swamy2024minimaximalist, scheid2024optimal, thekumparampil2024comparing, xie2025exploratory, chen2025avoiding, kveton2025active}.
However, despite this progress, most existing models remain limited to handling only \textit{pairwise} comparison feedback.
A few works~\citep{zhu2023principled, mukherjee2024optimal, thekumparampil2024comparing} explore the more general setting of \textit{multiple} comparisons, offering a strict extension beyond the pairwise case.
\citet{zhu2023principled} study the offline setting, where a dataset of questions (or contexts) along with corresponding ranking feedback over $K$ answers (or actions), labeled by human annotators, is available.
\citet{mukherjee2024optimal} investigate the online learning-to-rank problem~\citep{radlinski2008learning}, where a dataset of questions with $K$ candidate answers is provided, but no feedback is initially available.
\citet{thekumparampil2024comparing} consider a context-free setting (i.e., a singleton context), and the goal is to learn the ranking of $N \geq K$ answers based on ranking feedback obtained from subsets of size $K$.
However, all of their theoretical performance guarantees fail to show that using multiple comparisons provides any advantage over the pairwise setting (see Table~\ref{tab:existing_works}). 
This is counterintuitive, as ranking feedback is inherently more informative than pairwise feedback. Specifically, since a ranking over $K$ actions provides $\binom{K}{2}$ pairwise comparisons, it should, in principle, enable faster learning and lead to stronger performance guarantees.
\begin{table}[t]
\caption{Comparisons of settings and theoretical guarantees in related works on PbRL with ranking feedback.
Here, $T$ denotes the number of rounds (or the number of data points in the offline setting),  
$K$ is the (maximum) size of the offered action set (i.e., \textit{assortment}),  
and $d$ is the feature dimension.
$\rho$ represents the unknown context distribution.
Here, $\BigOTilde$ hides logarithmic factors and polynomial dependencies on $B$.
``Pred. Error'' refers to the prediction error.
}
\resizebox{\textwidth}{!}{
\begin{tabular}{lccccc}
\toprule
\multicolumn{1}{l}{}   
                                                                         & Setting     & Context                  &  Assortment        & Measure          &  Result  \\
\midrule
      \citet{zhu2023principled}                                          & Offline      & Accessible $\Xcal$      &  Given             & Suboptimality    &  $\BigOTilde \left( \textcolor{red}{e^B K^2 } \sqrt{\frac{d}{T}} \right)$ \\
      \citet{mukherjee2024optimal}                                       & Online       & Accessible $\Xcal$      &  Given             & Pred. Error   &  $\BigOTilde \left( \textcolor{red}{e^BK^3 } \frac{  d}{ \sqrt{T}} \right)$ \\
      \citet{thekumparampil2024comparing}                                & Online       & No context              &  Select $K$         & Pred. Error   &  $\BigOTilde \left( \textcolor{red}{e^B K^3 } \frac{  d}{ \sqrt{T}} \right)$ \\
     \textbf{This work }(Theorem~\ref{thm:main_PL},~\ref{thm:main_RB})   & Online       & Sampled $x \sim \rho$   &  Select $\leq K$    & Suboptimality   &   $\BigOTilde \left( \frac{d}{T} \sqrt{\sum_{t=1}^T \frac{1}{\textcolor{blue}{ |S_t|} }} \right)$     \\
\midrule
     \textbf{This work } (Theorem~\ref{thm:lower_bound})                 & Lower Bound  & Sampled $x \sim \rho$   &  Select $\leq K$    & Suboptimality   &    $\Omega \left( \frac{d}{\textcolor{blue}{K} \sqrt{T}} \right)$ \\
\bottomrule
\end{tabular}
}
\label{tab:existing_works}
\end{table}
%
Thus, the following fundamental question remains open:
\begin{center}
    \textit{Can we design an algorithm that achieves a \underline{strictly better theoretical guarantee} under \underline{multiple-option feedback} compared to the pairwise comparisons
    in the online PbRL setting?
    }
\end{center}
In this paper, we assume that the ranking feedback follows the Plackett-Luce (PL) model~\citep{plackett1975analysis, luce1959individual}, where, in each round, the learner receives ranking feedback over a subset of up to $K$ actions (with $K \leq N$) selected from a universe of $N$ actions.
This problem setup is closely related to that of \citet{thekumparampil2024comparing}; however, unlike their work, which focuses solely on a context-free setting (or equivalently, a fixed singleton context), we study a more general setting where contexts are diverse and drawn from an unknown distribution.

Under this problem setup,
we provide an affirmative answer to the above question by introducing a novel algorithm, 
\textit{Maximizing Average Uncertainty for Preference Optimization}
(\AlgName{}), which explicitly exploits the richer information available from ranking feedback under the Plackett–Luce (PL) model.
\AlgName{} selects action subsets by maximizing \textit{average uncertainty} and achieves a suboptimality gap that strictly improves upon what is attainable with pairwise comparisons.
In particular, we show that its suboptimality gap decreases with longer ranking feedback.

Furthermore, our suboptimality gap eliminates the \textit{exponential} dependence on the parameter norm bound, $\BigO(e^B)$, in the leading term.
This improvement stems from \textit{analytically} dividing the total rounds into \textit{warm-up} and \textit{non–warm-up} phases (see the proof sketch in Section~\ref{subsec:sub_gap_main}).
This represents a significant improvement over most prior works, where performance guarantees depend on $\BigO(e^B)$~\citep{saha2021optimal, saha2023dueling, zhu2023principled, xiong2023iterative, zhan2023provable, das2024active, xie2025exploratory, thekumparampil2024comparing, kveton2025active}.
Very recently, a few works~\citep{chen2025avoiding, di2024nearly} have successfully avoided the $\BigO(e^B)$ dependency by relying on auxiliary techniques or additional information—such as specialized sampling schemes~\citep{chen2025avoiding} or prior knowledge of $\kappa$~\citep{di2024nearly}—which, however, are often impractical.
Moreover, their methods are limited to pairwise comparison settings.
In contrast, our approach eliminates the $\BigO(e^B)$ dependency without using any auxiliary techniques and
considers more general ranking feedback beyond pairwise comparisons.
Our main contributions are summarized as follows:
\begin{itemize}
    \item \textbf{Improved sample efficiency via larger subsets:}
    We propose \AlgName{}, a novel algorithm for online PbRL (or RLHF) with PL ranking feedback, 
    which achieves a suboptimality gap of $\BigOTilde\!\left( \frac{d}{T} \sqrt{ \sum_{t=1}^T \frac{1}{|S_t|}} \right)$, 
    where $|S_t|$ is the size of the action subset offered at round $t$.
    This result provides the first rigorous theoretical guarantee that larger subsets directly improve sample efficiency.
    To the best of our knowledge, this is the first theoretical work in PbRL that explicitly demonstrates performance improvements as a function of the subset size $|S_t|$.

    \item \textbf{Improvement on $\BigO(e^B)$ dependency:}
    Our result eliminates the exponential dependence on the parameter norm bound, $\BigO(e^B)$, in the leading term—without relying on any auxiliary techniques—through a \textit{fully} refined analysis.
    This shows that the $\BigO(e^B)$ dependence widely observed in PbRL (or RLHF) and dueling bandit analyses is \textit{not inherent} but rather an artifact of loose analysis.
    Moreover, our key technique for removing the $\BigO(e^B)$ dependence—by dividing the rounds into warm-up and non–warm-up phases only in the analysis—can be seamlessly incorporated into existing PbRL or dueling bandit analyses, including regret-minimization frameworks without altering the underlying algorithms (see Appendix~\ref{app_subsec:regret}).
    To the best of our knowledge, this is the first PbRL work with ranking feedback beyond pairwise comparisons that avoids the $\BigO(e^B)$ dependence.

    \item \textbf{Efficiency of rank-breaking (RB):}
    We present both naive PL loss–based and rank-breaking (RB) loss–based learning approaches, and show that the RB formulation achieves superior computational efficiency and empirical performance compared to the PL counterpart.

    \item \textbf{Lower bound:} We establish a near-matching lower bound of $\Omega\!\left( \frac{d}{K \sqrt{T}} \right)$ under the PL model with ranking feedback, matching our upper bound up to a $K$ factor.
    This shows that leveraging richer ranking information (larger $K$) provably improves sample efficiency.

    \item \textbf{Experiment:} We empirically evaluate \AlgName{} on both synthetic and real-world datasets, showing its improved performance for larger $K$ and its superiority over existing baselines.
    
\end{itemize}
%
\section{Related Works}
\label{sec:related}
Fueled by the remarkable success of LLMs~\citep{christiano2017deep, ouyang2022training, rafailov2023direct},
the theoretical study of PbRL has rapidly emerged as a central focus within the research community.
Early work in this area traces back to the dueling bandits literature~\citep{yue2012k, zoghi2015copeland, saha2018battle, bengs2021preference}.

\textbf{Dueling bandits.}
The dueling bandit framework, introduced by~\citet{yue2012k}, departs from the classical multi-armed bandit setting by requiring the learner to select two arms and observe only their pairwise preference.
For general preferences, a single best arm that is globally dominant may not exist.
To address this, various alternative winners have been proposed, including the Condorcet winner~\citep{zoghi2014relative, komiyama2015regret}, Copeland winner~\citep{zoghi2015copeland, wu2016double, komiyama2016copeland}, Borda winner~\citep{jamieson2015sparse, falahatgar2017maxing, heckel2018approximate, saha2021adversarial, wu2023borda}, and von Neumann winner~\citep{ramamohan2016dueling, dudik2015contextual, balsubramani2016instance}, each with its own corresponding performance metric.

To address scalability and contextual information, \citet{saha2021optimal} proposed a structured contextual dueling bandit setting in which preferences are modeled using a Bradley–Terry–Luce (BTL) model~\citep{bradley1952rank} based on the unknown intrinsic rewards of each arm.
In a similar setting, \citet{bengs2022stochastic} studied a contextual linear stochastic transitivity model, and \citet{di2023variance} proposed a layered algorithm that achieves variance-aware regret bounds.
However, most prior dueling bandit works suffer from an $\mathcal{O}(e^B)$ dependence.
Only a few recent studies~\citep{di2024nearly, chen2025avoiding} have succeeded in removing this $\BigO(e^B)$ term, either by introducing additional complex subroutines~\citep{chen2025avoiding} or by relying on prior knowledge of $\kappa$~\citep{di2024nearly}.

\textbf{Preference-based reinforcement learning (PbRL).}
Building upon this line of work, subsequent research has extended the dueling bandit framework to the RL,
considering both online~\citep{xu2020preference, novoseller2020dueling, chen2022human, saha2023dueling, wu2023making}
and offline settings~\citep{zhu2023principled, zhan2023provable, liu2024provably}.
More recently, under the active learning framework—where the full set of contexts $\Xcal$ is accessible—many studies aim to improve sample efficiency
by selecting prompts either based on the differences in estimated rewards for their responses~\citep{muldrew2024active}
or through D-optimal design methods~\citep{mehta2023sample, scheid2024optimal, das2024active, mukherjee2024optimal, thekumparampil2024comparing, kveton2025active}.
However, most of these works focus exclusively on pairwise preference feedback and cannot be extended to more general ranking feedback cases.
\citet{mukherjee2024optimal} study the online learning-to-rank problem when prompts are given along with $K$ candidate answers, 
while \citet{thekumparampil2024comparing} investigate learning to rank $N \geq K$ answers from partial rankings over $K$ answers, but under a context-free setting.
In this paper, we consider a stochastic contextual setting (more general than~\citet{thekumparampil2024comparing}), where contexts are sampled from an unknown but fixed distribution, and aim to minimize the suboptimality gap using ranking feedback of up to length $K$.
For further related work, see Appendix~\ref{app_sec:More_Related}.

\section{Problem Setting and Preliminaries}
\label{sec:problem_setting}
\textbf{Notations.}
Given a set $\Xcal$, we use $\lvert \Xcal \rvert$ to denote its cardinality.  
For a positive integer $n$, we denote $[n] := \{1, 2, \ldots, n\}$.  
For a real-valued matrix $A$, we let $\|A\|_2 := \sup_{x:\|x\|_2=1}\|Ax\|_2$ 
which is the maximum singular value of $A$.  
We write $A \succeq A'$ if $A-A'$ is positive semidefinite.
For a univariate function $f$, we denote $\dot{f}$ as its derivative.

We have a set of contexts (or prompts), denoted by $\Xcal$, and a set of possible actions (or answers), denoted by $\Acal := \{a_1, \dots, a_N \}$.\footnote{
For simplicity, we assume a stationary action space $\Acal$, though it may depend on the context $x \in \Xcal$.
}. 
We consider preference feedback in the form of partial rankings over subsets of $\Acal$, and model this feedback using the Plackett-Luce (PL) distribution:
\begin{definition} [PL model] \label{def:PL_model}
    Let $\Scal := \{ S \subseteq \Acal \mid 2 \leq |S| \leq K \}$ be the collection of all action subsets whose sizes range from $2$ to $K$.
    For any $S \in \Scal$, let $\sigma$ denote the labeler’s ranking feedback—that is, a permutation of the elements in $S$.
    We write $\sigma_j$ for the $j$-th most preferred action under $\sigma$.
    We model the distribution of such rankings using the Plackett-Luce (PL) model~\citep{plackett1975analysis,luce1959individual}, defined as:
    \begin{align*}
        \PP (\sigma | x, S ; \thetab^\star)
        = \prod_{j=1}^{|S|}
        \frac{\exp \left( r_{\thetab^\star} ( x, \sigma_j )  \right)}{
            \sum_{k=j}^{|S|} \exp \left( 
                r_{\thetab^\star} ( x, \sigma_k )
            \right)
        },
        \quad \text{where}  \,\, (x, S) \in \Xcal \times \Scal
        .
        \numberthis \label{eq:PL_model}
    \end{align*}
    Here, $r_{\thetab^\star}$ represents a reward model parameterized by the unknown parameter $\thetab^\star$.
\end{definition}
When $K\!=\!2$, this reduces to the pairwise comparison framework considered in the Bradley-Terry-Luce (BTL) model~\citep{bradley1952rank}.
The probability that $a$ is preferred to $a'$ given $x$ can be expressed as:
\begin{align*}
    \PP ( a \succ a' | x ; \thetab^\star )
    = \frac{\exp \left( r_{\thetab^\star} ( x, a )  \right)}
    {
    \exp \left( r_{\thetab^\star} ( x, a )  \right)
    + \exp \left( r_{\thetab^\star} ( x, a' )  \right)
    }
    =\mu \left( r_{\thetab^\star} ( x, a ) - r_{\thetab^\star} ( x, a' )  \right)
    ,
    \numberthis \label{eq:BT_model}
\end{align*}
where $\mu(w) = \frac{1}{1 + e^{-w}}$ is the sigmoid function.
In this work, we assume a linear reward model:
\begin{assumption} \label{assum:linear_reward}
    Let $\phi: \Xcal \times \Acal \rightarrow \RR^d$ be a known feature map satisfying  $\max_{x,a} \| \phi(x,a) \|_2 \leq 1$, and let $\thetab^\star \in \RR^d$ denote the true but \textit{unknown} parameter. 
    The reward is assumed to follow a linear structure given by $r_{\thetab^\star}(x,a) = \phi(x,a)^\top \thetab^\star$.
    We further assume realizability, i.e., $\thetab^\star \in \Theta := \{ \thetab \in \RR^d \mid \| \thetab \|_2 \leq B \}$.
    Without loss of generality, we assume $B \geq 1$.
\end{assumption}

At each round $t \in [T]$, a context $x_t \in \Xcal$ is drawn from a fixed but unknown distribution $\rho$. 
Given the context $x_t$, the learning agent selects a subset of actions $S_t \in \Scal$—referred to as an \textit{assortment} throughout the paper—and receives a ranking over $S_t$ as feedback, generated according to the PL model.
Let $\pi^\star(x) = \argmax_a r_{\thetab^\star} (x,a)$ be the  optimal policy under the true reward $r_{\thetab^\star}$.
After $T$ rounds of interaction with the labeler, the goal is to output a policy $\widehat{\pi}_T : \Xcal \rightarrow \Acal$ that minimizes the \textit{suboptimality gap}, defined as:
\begin{align*}
    \SubOpt(T) 
        := 
        \EE_{x \sim \rho} \left[
            r_{\thetab^\star} \left( x, \pi^\star(x) \right)
            - r_{\thetab^\star} \left( x, \widehat{\pi}_T(x) \right)
        \right].
\end{align*}
%
%
\subsection{Loss Functions and Rank-Breaking}
\label{subsec:loss_and_RB}
In this paper, we consider two different losses for estimating the parameter: 
one directly induced by the PL model, and the other obtained by splitting the ranking feedback into pairwise comparisons.

\textbf{Plackett-Luce (PL) loss.} 
The PL loss function for round $t$ is defined as follows:
\begin{align*}
    \ell_t(\thetab) 
    := 
    \sum_{j=1}^{|S_t|} \ell^{(j)}_t (\thetab),
    \quad \text{where }\,
    \ell^{(j)}_t (\thetab) \!:=\! - \log \left( 
                \frac{\exp \left( \phi(x_t, \sigma_{tj})^\top \thetab   \right)}{
                \sum_{k=j}^{|S_t|} \exp \left( 
                     \phi(x_t, \sigma_{tk})^\top \thetab )
                \right)
                }
            \right).
    \numberthis \label{eq:PL_loss}
\end{align*}
Here, $\ell^{(j)}_t(\thetab)$ denotes the negative log-likelihood loss under the Multinomial Logit (MNL) model~\citep{mcfadden1977modelling}, conditioned on the assortment being the remaining actions in $S_t$ after removing the previously selected actions ${\sigma_{t1}, \dots, \sigma_{t(j-1)}}$—that is, over the set $S_t \setminus \{\sigma_{t1}, \dots, \sigma_{t(j-1)} \}$.

\textbf{Rank-Breaking (RB) loss.}
In addition to this standard approach, one can replace the full $|S_t|$-action ranking with its $\binom{|S_t|}{2}$ pairwise comparisons.
This technique, referred to as \textit{rank-breaking} (RB), decomposes (partial) ranking data into individual pairwise comparisons, treating each comparison as independent~\citep{azari2013generalized, khetan2016data, jang2017optimal, saha2024stop}.
Thus, the RB loss is defined as:
\begin{align*}
    \ell_t(\thetab) 
    &:= \!\!
    \sum_{j=1}^{|S_t|-1} \!\! \sum_{k=j+1}^{|S_t|} \ell^{(j,k)}_t (\thetab),
    \quad \text{where }\,
    \ell^{(j,k)}_t (\thetab) \!:= 
    \! -
    \log \left( 
        \frac{\exp \left( \phi(x_t, \sigma_{tj})^\top \thetab \right)}{
        \sum_{m \in \{j,k \} }
        \exp \left( \phi(x_t, \sigma_{tm})^\top \thetab \right) 
        }
    \right)
    .
    \numberthis \label{eq:RB_loss}
\end{align*}
This approach is applied in the current RLHF for
LLM (e.g.,~\citet{ouyang2022training}) and is also studied in the theoretical RLHF paper~\citep{zhu2023principled} under the offline setting.

\begin{figure}[t!]
    \begin{minipage}[t]{0.48\textwidth}
        \floatname{algorithm}{Procedure}
        \begin{algorithm}[H]
        \caption{\texttt{OMD-PL}, OMD for PL Loss}
        \label{alg:update_PL}
        \begin{algorithmic}
        \State \textbf{Input:} $\widehat{\thetab}_{t}^{(1)}, S_t, H_t$
            \For{$j = 1$ to $|S_t|$}
                \State Update $\tilde{H}_{t}^{(j)}$, $\widehat{\thetab}_{t}^{(j+1)}$ via~\eqref{eq:online_update}
            \EndFor
            \State \Return  $\widehat{\thetab}_{t}^{(|S_t|+1)}$
        \end{algorithmic}
        \end{algorithm}
    \end{minipage}
    \hfill
    \begin{minipage}[t]{0.48\textwidth}
        \floatname{algorithm}{Procedure}
        \begin{algorithm}[H]
        \caption{\texttt{OMD-RB}, OMD for RB Loss}
        \label{alg:update_RB}
        \begin{algorithmic}
        \State \textbf{Input:} $\widehat{\thetab}_{t}^{(1,2)}, S_t, H_t$
            \For{each $(j,k)$ such that $j < k \le |S_t|$}
                \State Update $\tilde{H}_{t}^{(j,k)}$, $\widehat{\thetab}_{t}^{(j,k+1)}$ via~\eqref{eq:online_update_RB}
            \EndFor
            \State \Return  $\widehat{\thetab}_{t}^{(|S_t|-1, |S_t|+1)}$
        \end{algorithmic}
        \end{algorithm}
    \end{minipage}
\end{figure}
\subsection{Online Parameter Estimation}
\label{subsec:online_param_estimatation}
Motivated by recent advances in Multinomial Logit (MNL) bandits~\citep{zhang2024online, lee2024nearly, lee2025improved}, we adopt an online mirror descent (OMD) algorithm to estimate the underlying parameter $\thetab^\star$, instead of relying on maximum likelihood estimation (MLE).
This enables a constant per-round computational cost, in contrast to the MLE-based approach, whose cost grows linearly with the number of rounds $t$.

\textbf{OMD update for PL loss.}
For the the PL loss~\eqref{eq:PL_loss}, we estimate the true parameter $\thetab^\star$ as follows:
\begin{align*}
    \widehat{\thetab}_{t}^{(j+1)} 
    &= \argmin_{\thetab \in \Theta }   \, \langle \nabla \ell_t^{(j)} (\widehat{\thetab}_{ t }^{(j)} ), \thetab \rangle
    + \frac{1}{2 \eta} \| \thetab - \widehat{\thetab}_t^{(j)} \|_{\tilde{H}_{t}^{(j)}}^2,
    \quad \, j = 1, \dots , |S_t|,
    \numberthis \label{eq:online_update}
\end{align*} 
where we write
$\widehat{\thetab}_{t}^{(|S_t|+1)}= \widehat{\thetab}_{t+1}^{(1)} $, and $\eta$ is the step-size parameter to be specified later.
The matrix $\tilde{H}_{t}^{(j)}$ is given by
$\tilde{H}_{t}^{(j)} := H_{t} + \eta \sum_{j'=1}^{j} \nabla^2 \ell_{t}^{(j')}(\widehat{\thetab}^{(j')}_t)$, where
\begin{align*}
    H_{t}:=  \sum_{s =1}^{t-1} \sum_{j=1}^{|S_s|} \nabla^2 \ell_s^{(j)} (\widehat{\thetab}_{s}^{(j+1)}) + \lambda \Ib_d,
    \quad
     \lambda >0
    .
    \numberthis \label{eq:H_update_PL}
\end{align*}
The optimization problem~\eqref{eq:online_update} can be solved using a single projected gradient step~\citep{orabona2019modern}, 
which enjoys a computational cost of only $\BigO(K d^3)$—independent of $t$~\citep{mhammedi2019lipschitz}, unlike MLE—and requires only $\BigO(d^2)$ storage, thanks to the incremental updates of $\tilde{H}_{t}^{(j)}$ and $H_t$.

\textbf{OMD update for RB loss.} Similarly, for the RB loss~\eqref{eq:RB_loss}, 
we estimate the underlying parameter as:
\begin{align*}
    \widehat{\thetab}_{t}^{(j, k+1)} 
    &= \argmin_{\thetab \in \Theta }   \, \langle \nabla \ell_t^{(j,k)} (\widehat{\thetab}_{ t }^{(j,k)} ), \thetab \rangle
    + \frac{1}{2 \eta} \| \thetab - \widehat{\thetab}_t^{(j,k)} \|_{\tilde{H}_{t}^{(j,k)}}^2,
    \quad \, 1 \leq j < k \leq |S_t|,
    \numberthis \label{eq:online_update_RB}
\end{align*} 
where we set $\widehat{\thetab}_{t}^{(j, |S_t|+1)} = \widehat{\thetab}_{t}^{(j+1, j+2)}$ for all $j < |S_t| - 1$ and for the final pair, let
$\widehat{\thetab}_{t}^{(|S_t|-1, |S_t|+1)} = \widehat{\thetab}_{t+1}^{(1,2)} $.
The matrix $\tilde{H}_{t}^{(j,k)}$ is defined as 
$\tilde{H}_{t}^{(j,k)} \!:= H_{t} + \eta \sum_{(j',k') \leq (j,k)}  \!\ \nabla^2 \ell_{t}^{(j',k')}(\widehat{\thetab}^{(j',k')}_t)$
\footnote{We write $(j',k') \le (j,k)$ to indicate lexicographic order, i.e., $j' < j$ or $j' = j$ and $k' \le k$.}
, where
\begin{align*}
    H_{t}:= \sum_{s =1}^{t-1} \sum_{j=1}^{|S_s|-1} 
    \!\! \sum_{k=j+1}^{|S_s|} 
    \nabla^2 \ell_s^{(j,k)} (\widehat{\thetab}_{s}^{(j,k+1)}) + \lambda \Ib_d, 
    \quad \lambda > 0.
    \numberthis \label{eq:H_update_RB}
\end{align*}
\begin{remark} [Computational cost of OMD] \label{remark:comp_cost_OMD}
    The per-round computational cost of the PL parameter update is $\BigO(K^2 d^3)$, since the parameter is updated $|S_t| \leq K$ times per round.
    Similarly, the cost for the RB parameter update is $\BigO(K^3 d^3)$, as the parameter is updated $\binom{|S_t|}{2}$ times per round.
\end{remark}

\begin{algorithm}[t!]
\caption{\AlgName{}: \textbf{M}aximizing \textbf{A}verage \textbf{U}ncertainty for \textbf{P}reference \textbf{O}ptimization}
\label{alg:main}
\begin{algorithmic}[1]
\State \textbf{Inputs:} maximum assortment size $K$, regularization parameter $\lambda$, step size $\eta$
\State \textbf{Initialize:} $H_1 = \lambda \mathbf{I}_d$, $\widehat{\thetab}_1 \in \Theta$
\For{round $t = 1$ to $T$}
    \State Observe $x_t \sim \rho$ and select $ S_t$ via~\eqref{eq:S_t_selection_greedy}
    \label{eq:alg_S_t}
    \State Observe ranking feedback $\sigma_t$ for $S_t$ 
    \label{eq:alg_feedback}
    \State $\widehat{\thetab}_{t+1} \gets \Call{\texttt{OMD-PL}}{\widehat{\thetab}_t, S_t, H_t}$ (Proc.~\ref{alg:update_PL}) \Comment{or $\Call{\texttt{OMD-RB}}{\widehat{\thetab}_t, S_t, H_t}$ (Proc.~\ref{alg:update_RB}) if \underline{RB loss}}
    \label{eq:alg_update_param}
    \State Update $H_{t+1} \leftarrow H_{t} + 
     \sum_{j=1}^{|S_t|} \nabla^2 \ell_t^{(j)} (\widehat{\thetab}_{t}^{(j+1)})
    $ via~\eqref{eq:H_update_PL} \Comment{or via~\eqref{eq:H_update_RB} if \underline{RB loss}}
    \label{eq:alg_final_policy}
\EndFor
\State \textbf{Return:} $\widehat{\pi}_T(x) \leftarrow \argmax_{a \in \Acal} \phi(x,a)^\top \widehat{\thetab}_{T+1}$
\end{algorithmic}
\end{algorithm}
\section{\AlgName{}: Maximizing Average Uncertainty}
\label{sec:algorithm}
In this section, we propose a new algorithm,~\AlgName{}, which selects an assortment that maximizes the \textit{average uncertainty} of $S_t$, thereby exploiting the potential benefits of a larger $K$.
For clarity of presentation, we first define the MNL probability~\citep{mcfadden1977modelling} for a given assortment $S$ at round $t$ as follows:
\begin{align*}
    P_t( a | S ; \thetab) := \frac{\exp \left( \phi(x_t, a)^\top \thetab \right)}{
    \sum_{a' \in S} \exp \left(
        \phi(x_t, a')^\top \thetab
    \right)
    }
    ,
    \quad 
    \forall
    a \in S.
    \numberthis \label{eq:MNL}
\end{align*}
Given an assortment $S$ and ranking feedback $\sigma$, let $S^{(j)}_\sigma := \{ \sigma_{j}, \dots, \sigma_{|S|} \}$ denote the remaining actions in $S$ after removing the first $j-1$ actions.
Let $\PP_t (\sigma | S ; \thetab) = \PP (x_t, \sigma | S ; \thetab)$, for simplicity.
Then, the PL model in Equation~\eqref{eq:PL_model} can be expressed as follows:
\begin{align*}
    \PP_t (\sigma | S ; \thetab)
    &=
    P_t(\sigma_{1} |  S ; \thetab)
    \cdot 
    P_t(\sigma_{2} | S\setminus \{\sigma_{1}\} ; \thetab)
    \cdot
    \;\dots\;
    \cdot
    P_t(\sigma_{ |S| } | \{\sigma_{ |S|}\} ; \thetab)
    =
    \prod_{j=1}^{|S|}
    P_t(\sigma_{j} | S^{(j)}_\sigma ; \thetab ),
\end{align*}

\textbf{Greedy assortment selection for PL loss.}
Given $H_t$ and $\widehat{\thetab}_t$, we define the function $f_t(S)$ as follows:
\begin{align*}
    f_t(S) :=
    \frac{1}{|S|}
    \sum_{j=1}^{|S|}
        \EE_{
        \substack{
        \sigma \sim \PP_t(\cdot | S ;\widehat{\thetab}_t )
        \\
        a \sim P_t(\cdot | S^{(j)}_\sigma; \widehat{\thetab}_t)
        }
        }
        \left[
            \left\| 
                \phi(x_t, a) - \EE_{a' \sim P_t(\cdot | S^{(j)}_\sigma; \widehat{\thetab}_t)} [\phi(x_t, a')]
            \right\|_{H_t^{-1}}^2
        \right].
        \numberthis \label{eq:S_t_selection_exact}
\end{align*}
%
%
Intuitively, finding $S$ that exactly maximizes $f_t(S)$ amounts to maximizing the \textit{average (mean-centered) uncertainty}—that is, the Mahalanobis dispersion of the feature vectors evaluated under $H_t^{-1}$.
However, computing the exact maximizer of Equation~\eqref{eq:S_t_selection_exact} is generally NP-hard.
In our setting, it is sufficient to add actions sequentially in a \textit{greedy manner}.
This is because our analysis centers on the suboptimality gap between two actions—one optimal and one chosen by the policy.
As a result, the suboptimality gap is (approximately) upper bounded by a term involving $f_t(S)$ for pairwise action sets (i.e., $|S|=2$).
Therefore, greedily adding actions that increase $f_t(S)$ is sufficient.

We initialize $S$ with a pair of actions ($|S|=2$) that maximizes the average information gain, as defined in Equation~\eqref{eq:S_t_selection_exact}.
Then, 
iteratively add one action at a time by
\begin{align*}
    a^\star \in \argmax_{a\in\Acal\setminus S}\ 
    \Delta_t(a\mid S),
    \qquad
    \text{where }\,
    \Delta_t(a\mid S):= f_t(S  \cup \{a\}) - f_t(S),
\numberthis \label{eq:S_t_selection_greedy}
\end{align*}
and accept $a^\star$ if $\Delta_t(a\mid S) \geq 0$, until $|S|=K$ or no non-negative gain remains.
This selection rule is central to our algorithm, as it facilitates a rapid decrease in the reward estimation error by favoring assortments that provide more informative feedback, especially when the assortment size $|S_t|$ is large.
Notably, the greedy variant of the selection rule can be implemented efficiently with a computational cost of $\mathcal{O}(N^2 d^3 +  L N K^2 d^2)$\footnote{$\mathcal{O}(N^2 d^3)$ to choose the first pair, followed by at most $K-2$ additions, each of which requires $\mathcal{O}(L N K d^2)$.} , where $L$ denotes the (approximate) expectation cost over the ranking $\sigma$.
This implementation avoids enumerating all $\binom{N}{K}$ possible subsets.

\textbf{Greedy assortment selection for RB loss.}
Analogous to Equation~\eqref{eq:S_t_selection_exact}, we define the following function to select an assortment that maximizes the average uncertainty:
\begin{align*}
    f_t(S) &:=
    \frac{1}{|S|}
    \sum_{a, a' \in S}
        \EE_{
        \bar{a} \sim P_t(\cdot | \{ a, a' \} ; \widehat{\thetab}_t)
        }
        \left[
            \left\| 
                \phi(x_t, \bar{a}) - \EE_{\tilde{a} \sim  P_t(\cdot | \{ a, a' \} ; \widehat{\thetab}_t)} [\phi(x_t, \tilde{a})]
            \right\|_{H_t^{-1}}^2
        \right]
    \\
    &= \frac{1}{2 |S|} 
     \sum_{a, a' \in S}
     \dot{\mu}\left( (\phi(x_t, a) - \phi(x_t, a'))^\top \widehat{\thetab}_t \right)
     \left\| 
                \phi(x_t, a) - \phi(x_t, a')
            \right\|_{H_t^{-1}}^2.
        \numberthis \label{eq:S_t_selection_exact_RB}
\end{align*}
We then perform greedy assortment selection according to Equation~\eqref{eq:S_t_selection_greedy}.
Note that, unlike in the PL loss case, assortment selection under the RB loss does not require taking expectations over rankings, as all pairs within $S$ are directly compared.
Therefore, the RB-based assortment selection is both \textit{exact} and \textit{computationally efficient}, without any additional expectation-approximation cost~$L$, and incurs a total computational cost of $\mathcal{O}\left(N^2d^3 + NK^3 d^2\right)$\footnote{$\mathcal{O}(N^2 d^3)$ to choose the first pair, followed by up to $K-2$ additional steps, each costing $\mathcal{O}(NK^2 d^3)$.}.

Once the assortment $S_t$ is selected, the algorithm receives the ranking feedback $\sigma_t$ from the labeler (Line~\ref{eq:alg_feedback}) and updates the parameter using Procedure~\ref{alg:update_PL} for the PL loss or Procedure~\ref{alg:update_RB} for the RB loss (Line~\ref{eq:alg_update_param}).
After $T$ rounds, the algorithm returns the final policy $\widehat{\pi}_T$, which selects actions by maximizing the estimated reward under the final parameter estimate $\widehat{\thetab}_{T+1}$ (Line~\ref{eq:alg_final_policy}).

\section{Main Results}
\label{sec:main_results}
%
%
%
\subsection{Suboptimality Gap of \AlgName{}}
\label{subsec:sub_gap_main}
We begin by presenting the online confidence bound for the PL loss, derived by extending the results of~\citet{lee2025improved}, who analyzed the MNL model~\citep{mcfadden1977modelling}.
Since the PL model constructs ranking probabilities as a product of MNL probabilities, their confidence bound can be directly applied to our setting by replacing the round $t$ with the cumulative number of updates $\sum_{s=1}^{t} |S_s|$.
\begin{corollary} [Online confidence bound for PL loss]
    \label{cor:online_CB_PL_main}
    Let $\delta \in (0, 1]$.
    We set $\eta = (1+ 3\sqrt{2} B)/2$ and $\lambda =  \max \{ 12 \sqrt{2} B \eta, 144 \eta d, 2 \} $.
    Then, under Assumption~\ref{assum:linear_reward}, with probability at least $1 - \delta$, we have
    \begin{align*}
        \| \widehat{\thetab}_t^{(j)} -\thetab^\star \|_{H_t^{(j)}} 
        \leq \beta_t (\delta)
        = \BigO\left(
        B
        \sqrt{
            d \log (t K/\delta)
            }
            + B \sqrt{\lambda}
        \right),
        \quad \forall t \geq 1, j \leq |S_t|
        ,
    \end{align*}
    where $H_t^{(j)} := H_t +  \sum_{j'=1}^{j-1} \nabla^2 \ell_s^{(j')} (\widehat{\thetab}_{s}^{(j'+1)}) + \lambda \Ib_d$.
\end{corollary}
This confidence bound is free of any polynomial dependency on $K$, which is primarily made possible by the improved self-concordant-like properties proposed by~\citet{lee2025improved}.
Moreover, for the RB loss, we can derive a confidence bound of the same order (see Proposition \ref{prop:online_CB_RB}).
Based on this confidence bound, we derive the suboptimality gap for~\AlgName{}, with the proof deferred to Appendix~\ref{app_sec:proof_main_PL}.
\begin{theorem} [Suboptimality gap for PL loss]
\label{thm:main_PL}
    Let $\delta \in (0, 1]$.
    Set $\lambda = \Omega\big(d \log (KT/\delta) + \eta(B + d) \big)$ and $\eta = \frac{1}{2}(1 + 3\sqrt{2} B)$.
    Define $\kappa := e^{-6B}$.
    If Assumption~\ref{assum:linear_reward} holds, then for any $T \ge e^{K/d}$, with probability at least $1-\delta$, \textup{\AlgName{}} (Algorithm~\ref{alg:main}) achieves the following suboptimality gap:
    \begin{align*}
        \SubOpt(T) = 
        \BigOTilde 
            \left(
                \frac{d }{T}
                \sqrt{
                    \sum_{t=1}^T \frac{1}{|S_t|}
                }
                + \frac{ d^2 K^4}{\kappa T}
            \right).
    \end{align*}
\end{theorem}
\textbf{Discussion of Theorem~\ref{thm:main_PL}.}
For sufficiently large $T$, the second (non-leading) term becomes negligible, and Theorem~\ref{thm:main_PL} shows that the suboptimality gap of~\AlgName{} decreases as the assortment size $|S_t|$ increases.
This establishes a strict advantage of receiving ranking feedback over larger assortments.
%
Moreover, our result does not involve any $\BigO(e^B)$ dependency in the leading term, a harmful dependency that commonly appears in prior works~\citep{saha2021optimal, saha2023dueling, zhu2023principled, xiong2023iterative, zhan2023provable, das2024active, thekumparampil2024comparing, kveton2025active}.
Although very recent studies~\citep{di2024nearly, chen2025avoiding} also achieve $\BigO(e^B)$-free performance in the leading term, they rely on auxiliary techniques and are restricted to pairwise preference feedback.
To the best of our knowledge, this is the first theoretical study that simultaneously establishes (i) the performance benefits of utilizing richer ranking feedback over larger assortments, and (ii) the elimination of the $\mathcal{O}(e^B)$ dependence in the leading term of the PbRL framework when accommodating multiple (i.e., more than two) options.

\textbf{Proof sketch of Theorem~\ref{thm:main_PL}.}
We provide a proof sketch for Theorem~\ref{thm:main_PL}.
For simplicity, we define $\psi_{t, a, a'} := \phi(x_t, a) - \phi(x_t, a')$.
We begin by defining the set of \textit{warm-up rounds}, denoted by $\WarmupRounds$, which consists of rounds with large (non-centered) uncertainty:
\begin{align*}
    \WarmupRounds
    &:= \left\{
        t \in [T]:
        \max_{a, a' \in \Acal} \|\psi_{t, a, a'} \|_{H_t^{-1} } \geq 
        1/\left(3 \sqrt{2} K \beta_{T+1}(\delta) \right)
    \right\}.
\end{align*} 
\textbf{1) Regret decomposition and assortment selection.}
The proof begins by decomposing the suboptimality gap into two components: the \textit{realized regrets} and a \textit{martingale difference sequence (MDS)}.
Since the MDS term can be readily bounded using the Azuma–Hoeffding inequality, the analysis focuses on bounding the realized regrets.

\begin{align*}
    \SubOpt(T)
    &= \frac{1}{T} \sum_{t=1}^T 
        \underbrace{
            \big(\psi_{t, \pi^\star(x_t), \widehat{\pi}_T(x_t)}\big)^\top 
            \thetab^\star}_{\text{realized regret of $\widehat{\pi}_T$ at round } t}
            +
            \underbrace{\frac{1}{T} \sum_{t=1}^T \text{MDS}_t}_{ = \BigOTilde(1/\sqrt{T})}
     \\
     &\lesssim 
     \frac{1}{T}\sum_{t \notin \WarmupRounds}
            \|
            \psi_{t, \pi^\star(x_t), \widehat{\pi}_T(x_t)}
            \|_{H_t^{-1}}
            \underbrace{
            \| \thetab^\star
                - \widehat{\thetab}_{T+1}
            \|_{H_T}
            }_{\leq \beta_t(\delta) = \BigOTilde(\sqrt{d}) }
        +   \BigOTilde\left(\frac{1}{\sqrt{T}} + \frac{ d^2 K^4}{\kappa T}\right)
        .
\end{align*}
In the inequality, we first use the fact that $-\big(\psi_{t, \pi^\star(x_t), \widehat{\pi}_T(x_t)}\big)^\top 
             \widehat{\thetab}_{T+1}
             \geq  0$, which follows
from definition of $\widehat{\pi}_T$.
We then apply Hölder’s inequality together with the inequality $H_{T+1} \succeq H_t$.
Moreover, we invoke Lemma~\ref{lemma:bound_Tw_PL}, which states that the regret incurred during the warm-up rounds  $\WarmupRounds$ is a non-leading term.
Finally, by applying Corollary~\ref{cor:online_CB_PL_main}, we bound
$\| \thetab^\star
    - \widehat{\thetab}_{T+1}
\|_{H_T}$ by $\BigOTilde(\sqrt{d} )$.

\textbf{2) Avoiding $\BigO(e^B )$ for non-warm-up rounds.}
For $t \notin \WarmupRounds$, we have 
\begin{align*}
    \big(\psi_{t, \pi^\star(x_t), \widehat{\pi}_T(x_t)} \big)^\top \thetab^\star
    \leq 
            \|
               \psi_{t, \pi^\star(x_t), \widehat{\pi}_T(x_t)}
            \|_{H_t^{-1}}
             \| \thetab^\star
                - \widehat{\thetab}_{T+1}
            \|_{H_T}
    \leq \frac{\beta_t(\delta)}{3\sqrt{2}K\beta_{T+1}(\delta)} \leq 1,
\end{align*}
which implies that $1/\dot{\mu} \big(
    \big(\psi_{t, \pi^\star(x_t), \widehat{\pi}_T(x_t)} \big)^\top \thetab^\star
        \big) \leq (1+e)^2$.
Thus, we obtain
\begin{align*}
    \frac{1}{T}\sum_{t \notin \WarmupRounds}
        \|
             \psi_{t, \pi^\star(x_t), \widehat{\pi}_T(x_t)}
        \|_{H_t^{-1}}
    &\leq
     \frac{(1+e)^2}{T}\sum_{t \notin   \WarmupRounds}
       \dot{\mu} \big(
            \big(\psi_{t, \pi^\star(x_t), \widehat{\pi}_T(x_t)} \big)^\top \thetab^\star
                \big)
          \|
             \psi_{t, \pi^\star(x_t), \widehat{\pi}_T(x_t)}
        \|_{H_t^{-1}}
    \\
    &\leq 
     \frac{e(1+e)^2}{T}\sum_{t \notin   \WarmupRounds}
        \dot{\mu} \big(
            \big(\psi_{t, \pi^\star(x_t), \widehat{\pi}_T(x_t)} \big)^\top \widehat{\thetab}_{t+1}
                \big)
        \| 
             \psi_{t, \pi^\star(x_t), \widehat{\pi}_T(x_t)}
        \|_{H_t^{-1} }
    ,
\end{align*}
where the last inequality follows from Lemma~\ref{lemma:dot_sigmoid_bound}, together with the fact that for any $a, a' \in \Acal$ and $t \notin \WarmupRounds$, 
$
|\psi_{t,a,a'}^\top (\thetab^\star
        - \widehat{\thetab}_{t+1})|
    \leq
    \|
      \psi_{t,a,a'}
    \|_{H_t^{-1}}
     \| \thetab^\star
        - \widehat{\thetab}_{t+1}
    \|_{H_T}
    \leq\beta_t(\delta)/\beta_{T+1}(\delta) = 1
$.
Then, by the Cauchy–Schwarz inequality, and ignoring constants, we have
\begin{align*}
    &\frac{1}{T}
    \sqrt{
    \sum_{t \notin   \WarmupRounds}
    \frac{1}{|S_t|}
    }
        \sqrt{ 
        \sum_{t \notin   \WarmupRounds}
            |S_t| \cdot
            \dot{\mu} \big(
                \big(\psi_{t, \pi^\star(x_t), \widehat{\pi}_T(x_t)} \big)^\top \widehat{\thetab}_{t+1}
                    \big)
            \| 
                 \psi_{t, \pi^\star(x_t), \widehat{\pi}_T(x_t)}
            \|_{H_t^{-1} }^2
        }
    \\
    &\lesssim \frac{1}{T}
    \sqrt{\sum_{t=1}^T
    \frac{1}{|S_t|}
    }
        \sqrt{ 
        \sum_{t \notin   \WarmupRounds}
        |S_t| 
        \cdot
        f_t( \{\pi^\star(x_t), \widehat{\pi}_T(x_t)  \})
        }
    \leq 
    \frac{1}{T}
    \sqrt{\sum_{t=1}^T
    \frac{1}{|S_t|}
    }
        \sqrt{ 
        \sum_{t \notin   \WarmupRounds}
        |S_t| 
        \cdot
        f_t( S_t)
        }
    \tag{$S_t$ selection}
    \\
    &\lesssim
    \sqrt{\sum_{t = 1}^T
            \frac{1}{|S_t|}
        }
    \sqrt{
        \sum_{t \notin   \WarmupRounds}
        \sum_{j=1}^{|S_t|}
        \EE_{
            \substack{
            \PP_t(\cdot | S_t ; \thetab^\star )
            \\
            P_t(\cdot | S_t^{(j)}; \widehat{\thetab}_t)}
            }
            \left[
                \left\| 
                    \phi(x_t, a) - \EE_{ P_t(\cdot | S_t^{(j)}; \widehat{\thetab}_t)} [\phi(x_t, a')]
                \right\|_{M_t^{-1} }^2
            \right]
        },
        \tag{Lemma~\ref{eq:M_t_def}}
\end{align*}
where $M_t =  \sum_{s \in [t-1] \setminus \Tcal^w} \sum_{j=1}^{|S_s|}
            \EE_{\sigma \sim \PP_s(\cdot | S_s, \thetab^\star) } \left[ \nabla^2 \ell^{(j)}_{s, \sigma} (\widehat{\thetab}_s) \right]
            + \lambda \Ib_d$ is  the expected version of $H_t$.
Finally, applying the elliptical potential lemma (Lemma~\ref{lemma:EPL_assortment}), we concludes the proof.
\begin{remark}[Generality of our technique for avoiding $\BigO(e^B)$]
    Our approach of dividing the total rounds into warm-up and non–warm-up phases to improve the $\BigO(e^B)$ dependency is broadly applicable.  
    In particular, this technique can be readily incorporated into most existing PbRL and dueling bandit algorithms without modifying their original algorithms.
    See Appendix~\ref{app_subsec:regret} for more details.
\end{remark}
Furthermore, under the rank-breaking (RB) parameter update~\eqref{eq:online_update_RB} and assortment selection rule~\eqref{eq:S_t_selection_exact_RB}, we establish a comparable suboptimality gap for the RB loss.
The proof is provided in Appendix~\ref{app_sec:proof_upper_RB}.
\begin{theorem}  [Online confidence bound for RB loss]
\label{thm:main_RB}
    Under the same setting as Theorem~\ref{thm:main_PL}, if the parameter is updated and the assortment is selected according to the RB-loss setting and $K^2 \leq (2 \sqrt{6} + 12 \sqrt{3} B)d$, 
    then for any $T \ge e^{K^2/d}$, with probability at least $1-\delta$, \textup{\AlgName{}} (Algorithm~\ref{alg:main}) satisfies
    \begin{align*}
        \SubOpt(T) = 
        \BigOTilde \left(
                \frac{ d  }{T}
                \sqrt{
                    \sum_{t=1}^T \frac{1}{|S_t|}
                }
                + \frac{d^2}{\kappa T}
            \right).
    \end{align*}
\end{theorem}
\textbf{Discussion of Theorem~\ref{thm:main_RB}.}
For sufficiently large $T$, the suboptimality gap in Theorem~\ref{thm:main_RB} matches the leading-order term of Theorem~\ref{thm:main_PL}, while its second (non-leading) term is tighter by a factor of $\BigO(K^4)$.
In addition, the assortment selection cost for RB~\eqref{eq:S_t_selection_exact_RB} is lower, since it avoids the expensive expectation over rankings required in the PL-loss case (Equation~\eqref{eq:S_t_selection_exact}).
This yields an important insight: under ranking feedback, \emph{breaking the ranking} and updating pairs independently can be more computationally efficient—and even more stable—than learning directly from the PL loss.
Although the bound is stated for the small-$K$ regime, the assumption appears mild in practice, since in LLM applications the dimension $d$ is typically much larger than $K$.
Moreover, this result provides a theoretical explanation for the empirical success of RLHF in LLMs (e.g.,~\citet{ouyang2022training}), where ranking feedback is commonly decomposed into pairwise comparisons for parameter estimation.

\subsection{Lower Bound}
\label{subsec:lower_bound}
%
%
\begin{theorem} [Lower bound] \label{thm:lower_bound}
    Suppose $T \geq d^2/(8 K^2 )$.
    Define the feature space as $\Phi := \mathcal{S}^{d-1}$, the unit sphere in $\mathbb{R}^d$, and let the parameter space be $\Theta = \{ -\mu, \mu \}^d$, where $\mu = \sqrt{d/(8K^2 T)}$.
    Then, for any final policy $\widehat{\pi}_T \in \triangle_{\Phi}$ returned after collecting $T$ samples (using any sampling policy), the expected suboptimality gap is lower bounded as:
    \begin{align*}
        \SubOpt(T) = \Omega \left( \frac{d}{K \sqrt{T}} \right).
    \end{align*}
\end{theorem}
\textbf{Discussion of Theorem~\ref{thm:lower_bound}.}
The proof is deferred to Appendix~\ref{app_sec:proof_lower}.
Theorem~\ref{thm:lower_bound} provides theoretical support for our upper bounds, especially in their dependence on $K$; in particular, it matches the upper bound in Theorem~\ref{thm:main_RB} when $|S_t|=K$.
To the best of our knowledge, this is the first lower bound on the suboptimality gap that incorporates PL ranking feedback in PbRL and formally shows that the suboptimality gap can diminish as $K$ grows, highlighting the advantage of utilizing ranking feedback over simple pairwise comparisons.
%

\section{Numerical Experiments}
\label{sec:experiments}
We conduct two sets of experiments to empirically validate our theoretical findings:
(i) one using synthetic data (Subsection~\ref{subsec:exp_synthetic}), and
(ii) another using two real-world datasets (Subsection~\ref{subsec:real}).
We compare our proposed algorithm, \AlgName{}, against three baselines:
(i) \texttt{DopeWolfe}~\citep{thekumparampil2024comparing}, which selects $K$ actions in a non-contextual setting;
(ii) \texttt{Uniform}, which uniformly samples assortments of size $K$ at random; and
(iii) \texttt{Best\&Ref} constructs an action pair ($|S_t|=2$) by combining the action that maximizes the current reward estimate with another sampled from a reference policy (e.g., uniform random or SFT), following the setup in Online GSHF~\citep{xiong2023iterative} and XPO~\citep{xie2025exploratory}.
In our experiments, the reference policy for \texttt{Best\&Ref} is set to the uniform random policy.
\subsection{Synthetic Data}
\label{subsec:exp_synthetic}
In the synthetic data experiment, for each instance, we sample the underlying parameter $\thetab^\star \sim \mathcal{N}(0, I_d)$ and normalize it to ensure that $\|\thetab^\star\|_2 \leq 1$.
At every round $t$, a context $x \in \Xcal$ is 
drawn uniformly at random, and its feature vector $\phi(x, \cdot)$ lies within the unit ball. 
We set $d=5$, $|\Acal| = N = 100$, and $|\Xcal| = 100$. 
We measure the suboptimality gap every $25$ rounds and report the mean over $20$ independent runs, together with one standard error.

The first two plots in Figure~\ref{fig:synthetic_experiment} show the suboptimality gap of~\AlgName{} under both the PL loss~\eqref{eq:PL_loss} and RB loss~\eqref{eq:RB_loss} as the maximum assortment size $K$ varies.
The results clearly show that performance improves as $K$ increases, supporting our theoretical findings.
In the third plot of Figure~\ref{fig:synthetic_experiment}, we compare the performance of~\AlgName{} with three baseline methods at the final round for $K=5$, showing that our algorithm significantly outperforms all baselines.
While \texttt{DopeWolfe} also considers the selection of $K$ actions from $N$ actions, it treats each context $x$ independently and is specifically designed for the context-free setting (i.e., a singleton context).
As a result, \texttt{DopeWolfe} cannot leverage information sharing across varying contexts and performs poorly in our setting.
Furthermore, \AlgName{} outperforms naive assortment selection strategies such as \texttt{Uniform} and \texttt{Best\&Ref}, as it explicitly chooses assortments that maximize the average uncertainty, thereby achieving more efficient exploration.
Finally, the results also show that the RB version of~\AlgName{} empirically outperforms the PL-based approach, since RB does not incur approximation errors when computing expectations.
See Appendix~\ref{app_subsec:exp_synthetic} for additional experimental results.

\subsection{Real-World Dataset}
\label{subsec:real}
We also conduct experiments using real-world datasets from TREC Deep Learning (TREC-DL)\footnote{https://microsoft.github.io/msmarco/TREC-Deep-Learning} and NECTAR\footnote{https://huggingface.co/datasets/berkeley-nest/Nectar}.
The TREC-DL dataset provides 100 candidate answers for each question, while the NECTAR dataset offers 7 candidate answers per question.
We sample $|\Xcal|=5000$ prompts from each dataset, with the corresponding set of actions (100 or 7 actions, respectively).

\begin{figure*}[t]
    \centering
    \begin{subfigure}[b]{0.325\textwidth}
        \includegraphics[width=\textwidth, trim=0mm 0mm 0mm 0mm, clip]{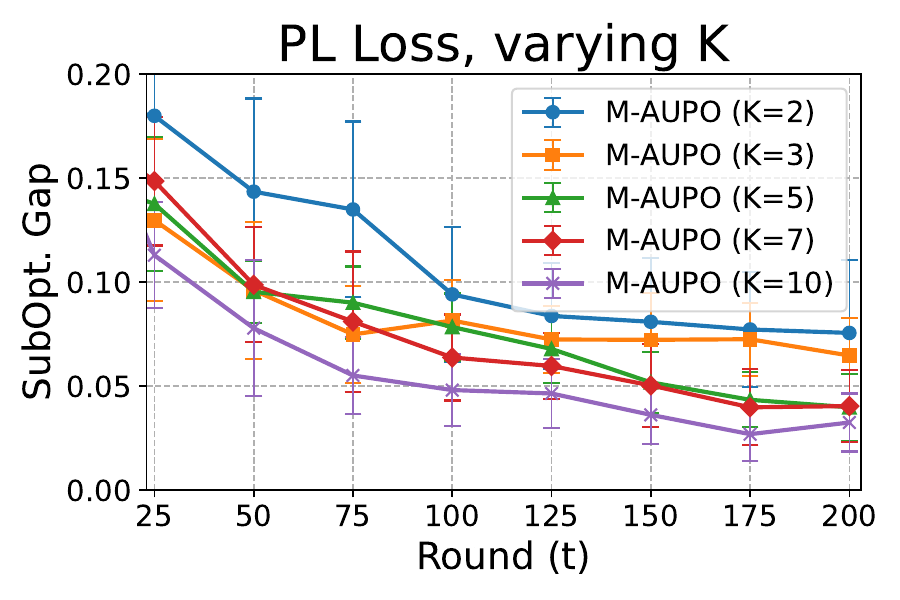}
        \label{fig:stochastic_context_pl_loss_varying_K}
    \end{subfigure}
    \hfill
    \begin{subfigure}[b]{0.325\textwidth}
        \includegraphics[width=\textwidth, trim=0mm 0mm 0mm 0mm, clip]{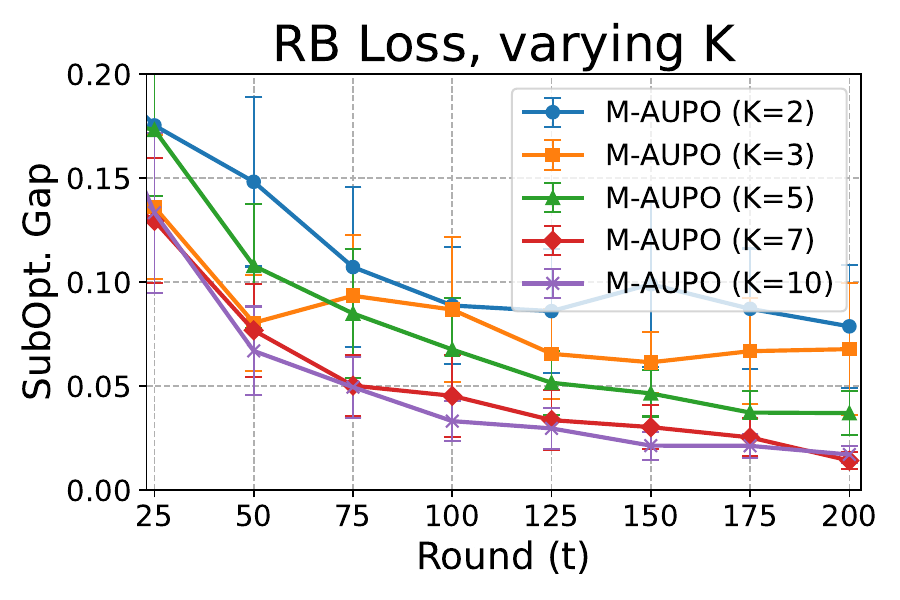}
        \label{fig:stochastic_context_rb_loss_varying_K}
    \end{subfigure}
    \hfill
    \begin{subfigure}[b]{0.325\textwidth}
        \includegraphics[width=\textwidth, trim=0mm 0mm 0mm 0mm, clip]{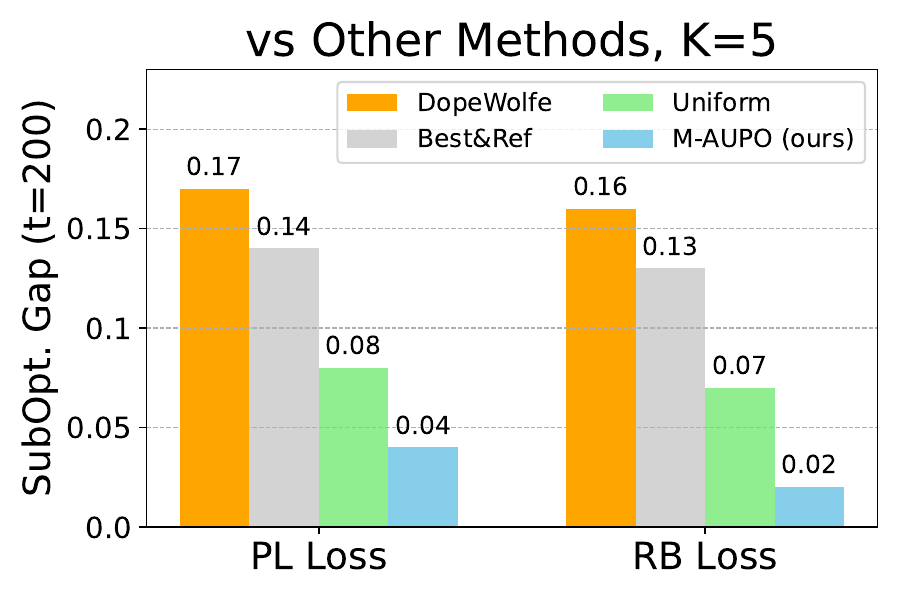}
        \label{fig:stochastic_context_pl_loss_vs_others}
    \end{subfigure}
    \caption{Synthetic data experiment: suboptimality gap of \AlgName{} under varying $K$, evaluated with PL loss (left) and RB loss (middle), along with comparison against other baselines (right).}
    \label{fig:synthetic_experiment}
\end{figure*}

We use the gemma-2b\footnote{https://huggingface.co/google/gemma-2b-it}~\citep{team2024gemma} LLM to construct the feature $\phi(x, a)$. 
Specifically, $\phi(x, a)$ is obtained by extracting the embedding of the concatenated prompt and response from the last hidden layer of the LLM, with size $d = 2048$. 
Additionally, we use the Mistral-7B~\citep{jiang2023mistral7b} reward model\footnote{https://huggingface.co/Ray2333/reward-model-Mistral-7B-instruct-Unified-Feedback} as the true reward model $r_{\thetab^\star}$ to generate ranking feedback and compute the suboptimality gap accordingly.
We measure the suboptimality gap every 2,500 rounds and report the average over 10 independent runs, along with the standard error.
In these experiments, we report results only for the RB version of~\AlgName{}, as it is computationally more efficient and empirically performs better, as shown in Figure~\ref{fig:synthetic_experiment}.

\begin{figure*}[t]
    \centering
    \begin{subfigure}[b]{0.325\textwidth}
        \includegraphics[width=\textwidth, trim=0mm 0mm 0mm 0mm, clip]{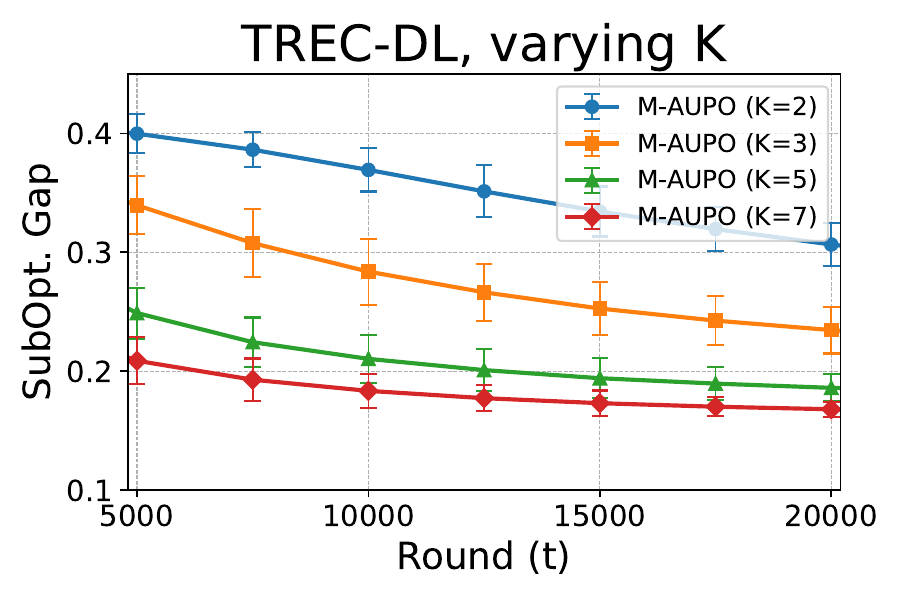}
        \label{fig:trec_varying_K}
    \end{subfigure}
    \hfill
    \begin{subfigure}[b]{0.325\textwidth}
        \includegraphics[width=\textwidth, trim=0mm 0mm 0mm 0mm, clip]{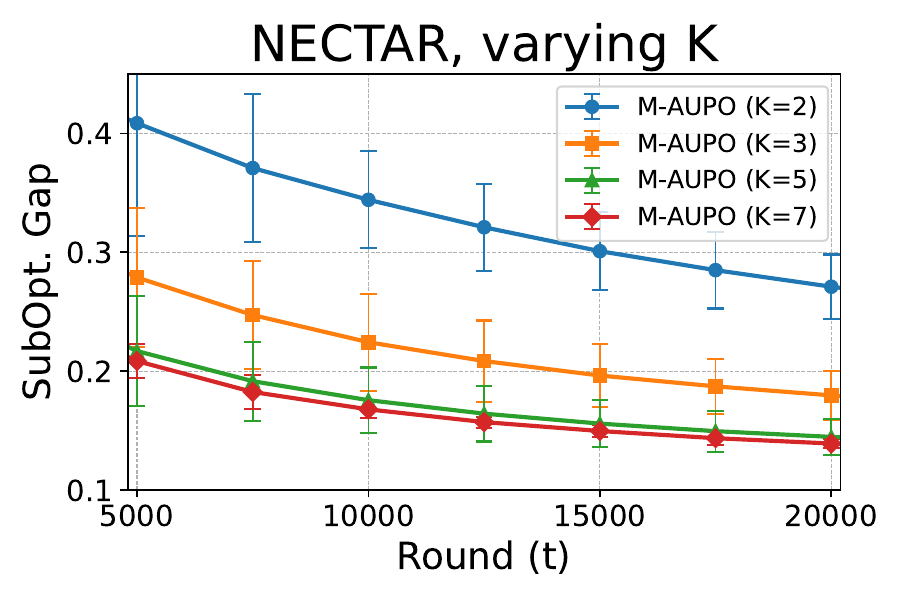}
        \label{fig:nectar_varying_K}
    \end{subfigure}
    \hfill
    \begin{subfigure}[b]{0.325\textwidth}
        \includegraphics[width=\textwidth, trim=0mm 0mm 0mm 0mm, clip]{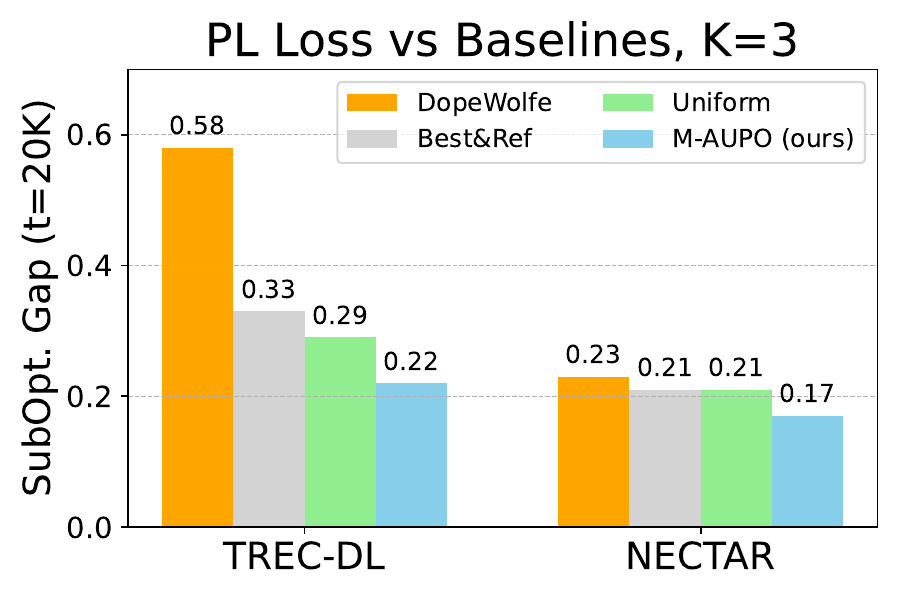}
        \label{fig:ms_nectar_pl_loss_vs_baselines}
    \end{subfigure}
    \caption{Real-world dataset experiment: suboptimality gap of \AlgName{} under varying $K$ on the TREC-DL dataset (left) and the NECTAR dataset (middle), along with comparison against other baselines (right).
    The results are rescaled to align the performances between the two datasets.
    }
    \label{fig:LLM_experiment}
\end{figure*}

The first two plots in Figure~\ref{fig:LLM_experiment} show that  the performance improves as $K$ increases on two real-world datasets
In the third plot of Figure~\ref{fig:LLM_experiment}, we compare the performance of~\AlgName{} with other baselines with $K=3$ at the final round, showing that \AlgName{} outperforms baselines by a large margin.
See Appendix~\ref{app_subsec:exp_LLM_dataset} for additional experimental details and results.

\section{Conclusion}
\label{sec:conclusion}
To the best of our knowledge, this work provides the first theoretical result in online PbRL showing that the suboptimality gap decreases as more options are revealed to the labeler for ranking feedback. 
Our analysis also removes the $\BigO(e^B)$ dependency in the leading term without modifying the algorithm, implying that existing PbRL and dueling bandit methods can similarly avoid this dependence through our refined analysis. 
Together, these results advance the theoretical understanding of PbRL, revealing both the value of richer feedback and the opportunity for more refined and efficient analysis.


\section*{Acknowledgements}
This work was supported by the National Research Foundation of Korea~(NRF) grant funded by the Korea government~(MSIT) (No.  RS-2022-NR071853, RS-2023-00222663, and RS-2025-25420849), by the Institute of Information \& communications Technology Planning \& Evaluation~(IITP) grant funded by the Korea government~(MSIT) (No. RS-2025-02263754 and RS-2025-25463302), by the AI-Bio Research Grant through Seoul National University,
and by the 2025 Global Google PhD Fellowship funded with support from Google.org.

\bibliography{main_bib}
\bibliographystyle{plainnat}


\newpage
\appendix
\onecolumn
\counterwithin{table}{section}
\counterwithin{lemma}{section}
\counterwithin{corollary}{section}
\counterwithin{theorem}{section}
\counterwithin{algorithm}{section}
\counterwithin{assumption}{section}
\counterwithin{figure}{section}
\counterwithin{equation}{section}
\counterwithin{condition}{section}
\counterwithin{remark}{section}
\counterwithin{definition}{section}
\counterwithin{proposition}{section}

\addcontentsline{toc}{section}{Appendix} 
\part{Appendix} 
\parttoc 
\section{Further Related Work}
\label{app_sec:More_Related}
In this section, we provide additional related work that complements Section~\ref{sec:related}.

\textbf{Logistic and MNL bandits.}
Our work is also closely related to logistic bandits and multinomial logit (MNL) bandits.
The logistic bandit problem~\citep{dong2019performance, faury2020improved, abeille2021instance, faury2022jointly, lee2024improved, lee2024unified} is a special case of the MNL bandit model in which the agent offers only a single item (i.e., $K = 1$) at each round and receives binary feedback indicating whether the item was selected (1) or not (0).
\citet{faury2020improved} examined how the regret in logistic bandits depends on the non-linearity parameter $\kappa$ of the logistic link function and proposed the first algorithm whose regret bound eliminates explicit dependence on $1/\kappa = \mathcal{O}(e^B)$.
\citet{abeille2021instance} further improved the theoretical dependency on $1/\kappa$ and established a matching, problem-dependent lower bound.
Building on this, \citet{faury2022jointly} developed a computationally efficient algorithm whose regret still matches the lower bound established by \citet{abeille2021instance}.

Multinomial logit (MNL) bandits tackle a more sophisticated problem than logistic bandits. 
Instead of offering a single item and observing binary feedback, the learner chooses a subset of items—underscoring the combinatorial nature of the task—and receives non-uniform rewards driven by an MNL choice model \citep{agrawal2019mnl, agrawal2017thompson, ou2018multinomial, chen2020dynamic, oh2019thompson, oh2021multinomial, perivier2022dynamic, agrawal2023tractable, lee2024nearly, lee2025improved}.
A recent breakthrough by \citet{lee2024nearly} closed a long-standing gap by providing a computationally efficient algorithm that attains the minimax-optimal regret for this setting.
Building on this result, \citet{lee2025improved} further reduced the regret bound by a factor polynomial in $B$ and logarithmic in $K$, and established the first variance-dependent regret bounds for MNL bandits.

Our work extends the online confidence bound analysis of \citet{lee2025improved} to the Plackett–Luce (PL) model. 
This extension is natural because the PL probability distribution decomposes into a sequence of MNL probabilities over successive choices.
Crucially, we leverage their key insight—that the MNL loss exhibits an $\ell_{\infty}$-self-concordant property—to eliminate the harmful $\BigO(e^B)$ dependence. 
This is one of the main contributions of our work (see Lemma~\ref{lemma:H_t_Lambda_PL}).

\textbf{RL with MNL models.}
Recent work has extended the Multinomial Logit (MNL) framework beyond bandit formulations to reinforcement learning.
\citet{lee2025combinatorial} introduced \textit{combinatorial RL with preference feedback}, a framework in which an agent learns to select subsets of items so as to maximize long-term cumulative rewards.

Another line of research incorporates MNL models directly into the transition dynamics.
\citet{hwang2022model} proposed MNL-MDPs, a class of Markov decision processes whose transition probabilities follow an MNL parameterization.
Building upon this formulation, \citet{cho2024randomized} improved the regret bounds by improving the exponential dependence on $B$, and \citet{park2025infinite} extended the analysis to the infinite-horizon setting.

\section{Notation}
\label{app_sec:notation}
Let $T$ denote the total number of rounds, with $t \in [T]$ representing the current round. 
We use $N$ for the total number of items, $K$ for the maximum assortment size, $d$ for the feature vector dimension, and $B$ as an upper bound on the norm of the unknown parameter. 
For notational convenience, we provide Table~\ref{table_symbols}.

For clarity, we derive the first- and second-order derivatives (i.e., gradients and Hessians) of the loss functions.
For the PL loss at round $t$ for the $j$'th ranking, let $y_{ti}^{(j)} =1$ if $i=j$, and $y_{ti}^{(j)} =0$ for otherwise.
Then, we have
\begin{align*}
    \ell^{(j)}_t (\thetab) 
        &= - \log \left( 
                \frac{\exp \left( \phi(x_t, \sigma_{tj})^\top \thetab   \right)}{
                \sum_{k=j}^{|S_t|} \exp \left( 
                     \phi(x_t, \sigma_{tk})^\top \thetab )
                \right)
                }
            \right)
       = - \sum_{i=j}^{|S_t|} 
            y_{ti}^{(j)}
            \log \Bigg( 
                \underbrace{
                    \frac{\exp \left( \phi(x_t, \sigma_{ti})^\top \thetab   \right)}{
                    \sum_{k=j}^{|S_t|} \exp \left( 
                         \phi(x_t, \sigma_{tk})^\top \thetab )
                    \right)
                    }
                }_{=: P_{t,\thetab}^{(j)}(\sigma_{ti})}
            \Bigg)
        \\
        &= - \sum_{i=j}^{|S_t|} 
            y_{ti}^{(j)}
            \log 
                P_{t,\thetab}^{(j)}(\sigma_{ti} )
            ,
    \\
    \nabla \ell^{(j)}_t (\thetab) 
    &= \sum_{i=j}^{|S_t|}
    \left(
        P_{t,\thetab}^{(j)}(\sigma_{ti} ) - y_{ti}^{(j)}
    \right)
    \phi(x_t, \sigma_{ti}),
    \\
    \nabla^2 \ell^{(j)}_t (\thetab) 
    &= \sum_{i=j}^{|S_t|}  P_{t,\thetab}^{(j)}(\sigma_{ti})
    \phi(x_t, \sigma_{ti}) \phi(x_t, \sigma_{ti})^\top
    \!- \sum_{i=j}^{|S_t|} \sum_{k=j}^{|S_t|}
     P_{t,\thetab}^{(j)}(\sigma_{ti} )
    P_{t,\thetab}^{(j)}(\sigma_{tk} )
    \phi(x_t, \sigma_{ti}) \phi(x_t, \sigma_{tk})^\top
    \\
    &= \frac{1}{2}
    \sum_{i=j}^{|S_t|} \sum_{k=j}^{|S_t|}
     P_{t,\thetab}^{(j)}(\sigma_{ti} )
    P_{t,\thetab}^{(j)}(\sigma_{tk} )
    \big(
        \phi(x_t, \sigma_{ti}) - \phi(x_t, \sigma_{tk})
    \big)
    \big(
        \phi(x_t, \sigma_{ti}) - \phi(x_t, \sigma_{tk})
    \big)^\top.
\end{align*}

\begin{table}[!ht]
\centering
    \caption{Symbols}
    \label{table_symbols}
    \resizebox{\textwidth}{!}{
    \begin{tabular}{ll}
         \toprule
         $\Xcal, \Acal, \Scal$   & context (prompt) space, action (answer) space, assortment space       \\[0.1cm]
         $\phi(x,a) \in \RR^d$       &      feature representation of  context-action pair $(x,a)$ \\[0.1cm]
         $z_{t j k}$            & $:= \phi(x_t, \sigma_{tj}) - \phi(x_t, \sigma_{tk})$, feature difference between $\sigma_{tj}$ and $\sigma_{tk}$ under context $x_t$ \\[0.1cm]
         $S_t$       &      assortment chosen by an algorithm at round $t$ \\[0.1cm]   
         $\ell^{(j)}_t (\thetab)$       &   $:= - \log \left( 
                                        \frac{\exp \left( \phi(x_t, \sigma_{tj})^\top \thetab   \right)}{
                                        \sum_{k=j}^{|S_t|} \exp \left( 
                                             \phi(x_t, \sigma_{tk})^\top \thetab )
                                        \right)
                                        }
                                    \right)$, PL loss at round $t$ for  $j$'th ranking \\[0.1cm]
        $ \ell^{(j,k)}_t (\thetab)$       &   $:= -
                                        \log \left( 
                                            \frac{\exp \left( \phi(x_t, \sigma_{tj})^\top \thetab \right)}{
                                            \sum_{m \in \{j,k \} }
                                            \exp \left( \phi(x_t, \sigma_{tm})^\top \thetab \right) 
                                            }
                                        \right)$, RB loss at round $t$ for comparison $\sigma_{tj}$ vs $\sigma_{tk}$ \\[0.1cm]
         $\nabla^2 \ell^{(j)}_t (\thetab)$       &    $= 
                                            \sum_{k=j}^{|S_t|}
                                            \sum_{k'=j}^{|S_t|}
                                            \frac{\exp\left( 
                                                \left(
                                                    \phi(x_t, \sigma_{tk})
                                                    + \phi(x_t, \sigma_{tk'})
                                                \right)^\top \thetab 
                                            \right)}{
                                            2 \left( 
                                                    \sum_{k'=j}^{|S_t|} \exp\left(
                                                    \phi(x_t, \sigma_{t k'})^\top \thetab          
                                                \right)
                                              \right)^2
                                            }
                                            \cdot
                                            z_{t k k'} z_{t k k'}^\top$ \\[0.1cm] 
         $\nabla^2 \ell_t^{(j,k)} (\thetab)$       &    $= 
                                            \dot{\mu} \left(
                                                z_{tjk}^\top \thetab
                                            \right)
                                            z_{tjk}
                                            z_{tjk}^\top$, where $\mu(w) = \frac{1}{1 + e^{-w}}$ is sigmoid function \\[0.1cm] 
         $\widehat{\thetab}^{(j+1)}_t$       &    online parameter estimate using PL loss at round $t$, after $j$'th update \\[0.1cm] 
         $\widehat{\thetab}^{(j,k+1)}_t$       &    online parameter estimate using RB loss at round $t$, after $(j,k)$'th comparison update \\[0.1cm] 
         $\eta$       &  $:= \frac{1}{2}(1+ 3\sqrt{2} B)$,  step-size parameter  \\[0.1cm] 
         $\lambda$       &  $:=  \Omega\big(d \log (KT/\delta) + \eta(B + d) \big)$,  regularization parameter \\[0.1cm]   
         $H_t$       &    $:=  \sum_{s =1}^{t-1} \sum_{j=1}^{|S_s|} \nabla^2 \ell_s^{(j)} (\widehat{\thetab}_{s}^{(j+1)}) + \lambda \Ib_d$
         (or $\sum_{s =1}^{t-1} \sum_{j=1}^{|S_s|-1} 
            \!\! \sum_{k=j+1}^{|S_s|} \!\!
            \nabla^2 \ell_s^{(j,k)} (\widehat{\thetab}_{s}^{(j,k+1)}) + \lambda \Ib_d$)
         \\[0.1cm] 
         $\tilde{H}^{(j)}_{t}$       &  $:= H_{t} + \eta \sum_{j'=1}^{j} \nabla^2 \ell_{t}^{(j')}(\widehat{\thetab}^{(j')}_t)$ \,\,(for PL loss)\\[0.1cm] 
         $\tilde{H}^{(j,k)}_{t}$       &  $:= H_{t} + \eta \sum_{(j',k') \leq (j,k)}  \!\ \nabla^2 \ell_{t}^{(j',k')}(\widehat{\thetab}^{(j',k')}_t)$ \,\,(for RB loss)\\[0.1cm] 
         $\beta_t(\delta)$       &    $:=\BigO\left(
                                B
                                \sqrt{
                                    d \log (t K/\delta)
                                    }
                                    + B \sqrt{\lambda}
                                \right)$,  confidence radius for $\thetab_t$ at round $t$ \\[0.1cm] 
        $\WarmupRounds$     &  $:=  \left\{
                                    t \in [T]:
                                    \max_{a, a' \in \Acal} \| \phi(x_t, a) - \phi(x_t, a') \|_{H_t^{-1} } \geq 
                                    \frac{1}{3 \sqrt{2} K \beta_{T+1}(\delta)}
                                \right\}$, warm-up rounds  \\[0.1cm]              
        $M_t $     &  $:= \sum_{s \in [t-1] \setminus \Tcal^w} \sum_{j=1}^{|S_s|}
                    \EE_{\sigma \sim \PP_s(\cdot | S_s, \thetab^\star) } \left[ \nabla^2 \ell^{(j)}_s (\widehat{\thetab}_s) \right]
                    + \lambda \Ib_d$ \\[0.1cm]    
         \bottomrule
    \end{tabular}
}
\end{table}

For the RB loss at round $t$ for the pairwise comparison between $\sigma_{tj}$ and $\sigma_{tk}$, let $y_{ti}^{(j,k)} =1$ if $i=j$, and $y_{ti}^{(j,k)} =0$ for otherwise (i.e., when $i=k$).
Then, 
we have
\begin{align*}
     \ell^{(j,k)}_t (\thetab)
     &=  - \log \left( 
            \frac{\exp \left( \phi(x_t, \sigma_{tj})^\top \thetab \right)}{
            \exp \left( \phi(x_t, \sigma_{tj})^\top \thetab \right) 
            + 
            \exp \left( \phi(x_t, \sigma_{tk})^\top \thetab \right) 
            }
        \right)
    \\
    &= - \log \mu \!\left( 
            \big(
                \phi(x_t, \sigma_{tj}) - \phi(x_t, \sigma_{tk})
            \big)^\top \thetab
        \right),
        \quad \text{where }\, \mu(w) = \frac{1}{1 + e^{-w}},
    \\
    \nabla \ell^{(j,k)}_t (\thetab)
    &= 
    \left(
        \mu \! \left( 
            \big(
                \phi(x_t, \sigma_{tj}) - \phi(x_t, \sigma_{tk})
            \big)^\top \thetab
        \right)  - 1
    \right)
    \big(
        \phi(x_t, \sigma_{tj}) - \phi(x_t, \sigma_{tk})
    \big),
    \\
    \nabla^2 \ell^{(j,k)}_t (\thetab)
    &= 
    \dot{\mu}\!
    \left( 
            \big(
                \phi(x_t, \sigma_{tj}) - \phi(x_t, \sigma_{tk})
            \big)^\top \thetab
        \right)
    \big(
        \phi(x_t, \sigma_{tj}) - \phi(x_t, \sigma_{tk})
    \big)
    \big(
        \phi(x_t, \sigma_{tj}) - \phi(x_t, \sigma_{tk})
    \big)^\top.
\end{align*}


\section{Proof of Theorem~\ref{thm:main_PL}}
\label{app_sec:proof_main_PL}
In this section, we present the proof of Theorem~\ref{thm:main_PL}.

\subsection{Main Proof of Theorem~\ref{thm:main_PL}} \label{app_subsec:main_proof_thm:main_PL}
\textbf{PL loss and OMD.}
We begin by recalling the loss function and the parameter update rule.
Specifically, we use the PL loss defined in Equation~\eqref{eq:PL_loss} and update the parameter according to Equation~\eqref{eq:online_update}.
\begin{align*}
    \ell_t(\thetab) 
    := 
    \sum_{j=1}^{|S_t|} 
    \underbrace{ - \log \left( 
                \frac{\exp \left( \phi(x_t, \sigma_{tj})^\top \thetab   \right)}{
                \sum_{k=j}^{|S_t|} \exp \left( 
                     \phi(x_t, \sigma_{tk})^\top \thetab )
                \right)
                }
            \right)}_{=: \ell^{(j)}_t (\thetab)}
    =  \sum_{j=1}^{|S_t|} \ell^{(j)}_t (\thetab),
\end{align*}
and
\begin{align*}
    \widehat{\thetab}_{t}^{(j+1)} 
    &= \argmin_{\thetab \in \Theta }   \, \langle \nabla \ell_t^{(j)} (\widehat{\thetab}_{ t }^{(j)} ), \thetab \rangle
    + \frac{1}{2 \eta} \| \thetab - \widehat{\thetab}_t^{(j)} \|_{\tilde{H}_{t}^{(j)}}^2,
    \quad \, j = 1, \dots , |S_t|,
\end{align*} 
where $\widehat{\thetab}_{t}^{(|S_t|+1)}= \widehat{\thetab}_{t+1}^{(1)} $, and $\eta := \frac{1}{2}(1 + 3 \sqrt{2}B)$ is the step-size parameter.
The matrix $\tilde{H}_{t}^{(j)}$ is given by
$\tilde{H}_{t}^{(j)} := H_{t} + \eta \sum_{j'=1}^{j} \nabla^2 \ell_{t}^{(j')}(\widehat{\thetab}^{(j')}_t)$, where
\begin{align*}
    H_{t}:=  \sum_{s =1}^{t-1} \sum_{j=1}^{|S_s|} \nabla^2 \ell_s^{(j)} (\widehat{\thetab}_{s}^{(j+1)}) + \lambda \Ib_d,
    \quad \lambda >0
    .
\end{align*}

\textbf{Online confidence bound for PL loss.}
Now, we present the confidence bound for online parameter estimation in MNL models, as recently proposed by~\citet{lee2025improved}.
\begin{lemma} [Online confidence bound, Theorem 4.2 of~\citealt{lee2025improved}]
\label{lemma:online_CB}
    Let $\delta \in (0, 1]$.
    We set $\eta = (1+ 3\sqrt{2} B)/2$ and $\lambda = \max \{ 12 \sqrt{2} B \eta, 144 \eta d, 2 \}$.
    Then, under Assumption~\ref{assum:linear_reward}, with probability at least $1 - \delta$, we have
    \begin{align*}
        \| \widehat{\thetab}_t -\thetab^\star \|_{H_t} 
        \leq \beta_t (\delta)
        = \BigO\left(
        B
        \sqrt{
            d \log (t/\delta)
            }
            + B \sqrt{\lambda}
        \right),
        \quad \forall t \geq 1
        .
    \end{align*}
\end{lemma}
We now extend this result to our setting.
Since the total number of updates up to round $t$ is $\sum_{s=1}^t |S_s|$,
the corresponding confidence bound can be expressed as follows:
\begin{corollary} [Restatement of Corollary~\ref{cor:online_CB_PL_main}, Online confidence bound for PL loss]
    \label{cor:online_CB_PL}
    Let $\delta \in (0, 1]$.
    We set $\eta = (1+ 3\sqrt{2} B)/2$ and $\lambda = \max \{ 12 \sqrt{2} B \eta, 144 \eta d, 2 \}$.
    Then, under Assumption~\ref{assum:linear_reward}, with probability at least $1 - \delta$, we have
    \begin{align*}
        \| \widehat{\thetab}_t^{(j)} -\thetab^\star \|_{H_t^{(j)}} 
        \leq \beta_t (\delta)
        = \BigO\left(
        B
        \sqrt{
            d \log (t K/\delta)
            }
            + B \sqrt{\lambda}
        \right),
        \quad \forall t \geq 1, j \leq |S_t|
        ,
    \end{align*}
    where $H_t^{(j)} := H_t +  \sum_{j'=1}^{j-1} \nabla^2 \ell_s^{(j')} (\widehat{\thetab}_{s}^{(j'+1)}) + \lambda \Ib_d$ and 
    $\widehat{\thetab}^{(1)}_t = \widehat{\thetab}_t$.
\end{corollary}

\textbf{Useful definitions.}
For clarity, we write $\ell_{t,\sigma}(\thetab)$ to make explicit that the loss function is random with respect to the ranking $\sigma$.
When the ranking is realized as $\sigma_t$, we simply denote it by $\ell_t(\thetab)$.
We define the set of \textit{warm-up rounds}, denoted by $\WarmupRounds$, which consists of rounds with large uncertainty, as:
\begin{align*}
    \WarmupRounds
    &:= \left\{
        t \in [T]:
        \max_{a, a' \in \Acal} \| \phi(x_t, a) - \phi(x_t, a') \|_{H_t^{-1} } \geq 
        \frac{1}{3 \sqrt{2} K \beta_{T+1}(\delta)}
    \right\}, 
    \numberthis \label{eq:warmup_rounds}
\end{align*} 
where $\beta_{T+1}(\delta)$ denotes the confidence radius as defined in Corollary~\ref{cor:online_CB_PL}.

Furthermore, we define the expected version of the Hessian matrix (taken with respect to the randomness of the ranking feedback) as follows:
\begin{align*}
    M_t =  \sum_{s \in [t-1] \setminus \Tcal^w} \sum_{j=1}^{|S_s|}
            \EE_{\sigma \sim \PP_s(\cdot | S_s, \thetab^\star) } \left[ \nabla^2 \ell^{(j)}_{s, \sigma} (\widehat{\thetab}_s) \right]
            + \lambda \Ib_d
    \numberthis \label{eq:M_t_def}
\end{align*}
Note that  $\nabla^2 \ell^{(j)}_{s, \sigma} (\widehat{\thetab}_s)$ is a random matrix conditional on the assortment $S_s$, where the randomness arises from the ranking feedback $\sigma$.

\textbf{Key lemmas.}
We now present key lemmas needed to prove Theorem~\ref{thm:main_PL}.
The following lemma, one of our main contributions, is crucial for avoiding the $1/ \kappa = \mathcal{O}(e^{B})$ dependency in the leading term.
\begin{lemma} [Empirical-to-expected Hessian lower bound]
\label{lemma:H_t_Lambda_PL}
    Let $M_t$ be defined as in Equation~\eqref{eq:M_t_def}.
    Set $\lambda = \Omega(d \log (KT/\delta))$.
    Then, for all $t \in [T]$, 
    with probability at least $1-\delta$,
    we have 
    \begin{align*}
        H_t \succeq  \frac{1}{3 e^2}
            M_t.
    \end{align*}
\end{lemma}
The proof is deferred to Appendix~\ref{app_subsubsec:proof_of_lemma:H_t_Lambda_PL}.

The following lemma is the elliptical potential lemma adapted to our setting.
\begin{lemma} [Elliptical potential for expected mean-centered uncertainty]
\label{lemma:EPL_assortment} 
    Let $M_t$ be defined as in Equation~\eqref{eq:M_t_def}. 
    Let $P_t(a \mid S_t^{(j)}; \thetab)$ denote the MNL probability for the subset of remaining actions in $S_t$ after removing the first $j-1$ actions according to the ranking $\sigma$.
    Then, for any $\lambda > 0$, we have 
    \begin{align*}
     \sum_{t \notin  \WarmupRounds}
     \min &\left\{1, 
        \sum_{j=1}^{|S_t|}
        \EE_{
            \substack{
            \sigma \sim \PP_t(\cdot | S_t ; \thetab^\star )
            \\
            a \sim P_t(\cdot | S_t^{(j)}; \widehat{\thetab}_t)}
            }
            \left[
                \left\| 
                    \phi(x_t, a) - \EE_{a \sim  P_t(\cdot | S_t^{(j)}; \widehat{\thetab}_t)} [\phi(x_t, a')]
                \right\|_{M_t^{-1}}^2
            \right]
     \right\}
     \\
        &\leq 2 d
        \log \left(
            1 + \frac{KT}{d \lambda}
        \right).
    \end{align*}        
\end{lemma}
The proof is deferred to Appendix~\ref{app_subsubsec:proof_of_lemma:EPL_assortment}.

The size of the set $\WarmupRounds$ is bounded as described in the following lemma.
\begin{lemma} 
\label{lemma:bound_Tw_PL} 
    Let
        $\WarmupRounds
        = \big\{
            t \in [T]:
            \max_{a, a' \in \Acal} \| \phi(x_t, a) - \phi(x_t, a') \|_{H_t^{-1} } \geq 
            \frac{1}{3 \sqrt{2} K \beta_{T+1}(\delta)}
        \big\}$.
    Define $\kappa := e^{-6B}$.
    Set $\lambda \geq 1$.
    Then, the size of the set $\WarmupRounds $ is bounded as follows:
    \begin{align*}
         \left| \WarmupRounds \right|
         \leq \frac{288 K^4}{\kappa} \beta_{T+1}(\delta)^2
         d \log \left(
            1 + \frac{2 KT}{d \lambda}
         \right)
         .
    \end{align*}
\end{lemma}
The proof is deferred to Appendix~\ref{app_subsubsec:proof_of_lemma:bound_Tw_PL}.

We are now ready to provide the proof of Theorem~\ref{thm:main_PL}.
%
\begin{proof} [Proof of Theorem~\ref{thm:main_PL}]
    To begin, we define a martingale difference sequence (MDS) $\zeta_t$ as follows:
    \begin{align*}
        \zeta_{t} := &\EE_{x \sim \rho} \left[
        \big(
            \phi \left( x, \pi^\star(x) \right)
            - \phi \left( x, \widehat{\pi}_T(x) \right)
        \big)^\top 
            \thetab^\star
    \right] 
    - 
        \big(
            \phi \left( x_t, \pi^\star(x_t) \right)
            - \phi \left( x_t, \widehat{\pi}_T(x_t) \right)
        \big)^\top 
            \thetab^\star,
    \end{align*}
    which satisfies $|\zeta_{t}| \leq 2 B$.
    Then, by the definition of the suboptimality gap, we have
    \begin{align*}
        \SubOpt(T) 
        &= \EE_{x \sim \rho} \left[
            \big(
                \phi \left( x, \pi^\star(x) \right)
                - \phi \left( x, \widehat{\pi}_T(x) \right)
            \big)^\top 
            \thetab^\star
        \right]
        \\
        &= \frac{1}{T} \sum_{t=1}^T \big(
                \phi \left( x_t, \pi^\star(x_t) \right)
                - \phi \left( x_t, \widehat{\pi}_T(x_t) \right)
            \big)^\top 
            \thetab^\star
            +
            \frac{1}{T} \sum_{t=1}^T \zeta_t
        \tag{Def. of $\zeta_t$}
        \\
        &\leq 
         \frac{1}{T} \sum_{t=1}^T \big(
                \phi \left( x_t, \pi^\star(x_t) \right)
                - \phi \left( x_t, \widehat{\pi}_T(x_t) \right)
            \big)^\top 
            \left( \thetab^\star
                - \widehat{\thetab}_{T+1}
            \right)
            +
            \frac{1}{T} \sum_{t=1}^T \zeta_t
        \tag{$\widehat{\pi}_T(x_t) = \argmax_{a \in \Acal}\phi(x_t,a)^\top \widehat{\thetab}_{T+1} $}
        \\
        &\leq
        \frac{1}{T}\sum_{t=1}^T
        \big(
                \phi \left( x_t, \pi^\star(x_t) \right)
                - \phi \left( x_t, \widehat{\pi}_T(x_t) \right)
            \big)^\top 
            \left( \thetab^\star
                - \widehat{\thetab}_{T+1}
            \right)
        +  \BigOTilde \left(\frac{1}{\sqrt{T}} \right),
         \numberthis \label{eq:main_proof_major_term}
    \end{align*}
    where the last inequality follows from the Azuma–Hoeffding inequality. Specifically, for any $T \geq 1$, with probability at least $1 - \delta$, we have
    \begin{align*}
         \frac{1}{T} \sum_{t=1}^T \zeta_{t}
         \leq \frac{ 1}{T}  \sqrt{8 B^2 T \log(1/ \delta)}
         = \BigOTilde \left( \frac{1}{\sqrt{T}} \right).
    \end{align*}
    To complete the proof, it remains to bound the first term in Equation~\eqref{eq:main_proof_major_term}.
    \begin{align*}
        \frac{1}{T}& \sum_{t=1}^T
        \big(
            \phi \left( x_t, \pi^\star(x_t) \right)
            - \phi \left( x_t, \widehat{\pi}_T(x_t) \right)
        \big)^\top 
        \left( \thetab^\star
            - \widehat{\thetab}_{T+1}
        \right)
        \\
        &= \frac{1}{T} \sum_{t \in \WarmupRounds}
        \big(
            \phi \left( x_t, \pi^\star(x_t) \right)
            - \phi \left( x_t, \widehat{\pi}_T(x_t) \right)
        \big)^\top 
        \left( \thetab^\star
            - \widehat{\thetab}_{T+1}
        \right)
        \\
        &\quad+
        \frac{1}{T} \sum_{t \notin \WarmupRounds}
        \big(
            \phi \left( x_t, \pi^\star(x_t) \right)
            - \phi \left( x_t, \widehat{\pi}_T(x_t) \right)
        \big)^\top 
        \left( \thetab^\star
            - \widehat{\thetab}_{T+1}
        \right)
        \\
        &\leq  
        \frac{4 B}{T} |\WarmupRounds |
        +
        \frac{1}{T} \sum_{t \notin \WarmupRounds}
        \big(
            \phi \left( x_t, \pi^\star(x_t) \right)
            - \phi \left( x_t, \widehat{\pi}_T(x_t) \right)
        \big)^\top 
        \left( \thetab^\star
            - \widehat{\thetab}_{T+1}
        \right)
        \tag{Assumption~\ref{assum:linear_reward}}
        \\
        &\leq  
        \tag{Lemma~\ref{lemma:bound_Tw_PL}}
        \BigO \left(\frac{ B K^4}{ \kappa T} \beta_{T+1}(\delta)^2
         d \log \left(
            1 + \frac{ T}{d \lambda}
         \right)
         \right)
        \\
        &\quad+
        \frac{1}{T} \sum_{t \notin   \WarmupRounds}
        \big(
            \phi \left( x_t, \pi^\star(x_t) \right)
            - \phi \left( x_t, \widehat{\pi}_T(x_t) \right)
        \big)^\top 
        \left( \thetab^\star
            - \widehat{\thetab}_{T+1}
        \right).
        \numberthis \label{eq:main_proof_major_term_low_EP}
    \end{align*}
    To further bound the last term of Equation~\eqref{eq:main_proof_major_term_low_EP}, we get
    \begin{align*}
        \frac{1}{T} \sum_{t \notin   \WarmupRounds}
        &\big(
            \phi \left( x_t, \pi^\star(x_t) \right)
            - \phi \left( x_t, \widehat{\pi}_T(x_t) \right)
        \big)^\top 
        \left( \thetab^\star
            - \widehat{\thetab}_{T+1}
        \right)
        \\
        &\leq \frac{1}{T} \sum_{t \notin   \WarmupRounds}
        \left\| 
            \phi \left( x_t, \pi^\star(x_t) \right)
            - \phi \left( x_t, \widehat{\pi}_T(x_t) \right)
        \right\|_{H_{T+1}^{-1} }
        \left\|
            \thetab^\star
            - \widehat{\thetab}_{T+1}
        \right\|_{H_{T+1}}
        \tag{Hölder's ineq.}
        \\
        &\leq \frac{1}{T} \sum_{t \notin   \WarmupRounds}
        \left\| 
            \phi \left( x_t, \pi^\star(x_t) \right)
            - \phi \left( x_t, \widehat{\pi}_T(x_t) \right)
        \right\|_{H_t^{-1} }
        \left\|
            \thetab^\star
            - \widehat{\thetab}_{T+1}
        \right\|_{H_{T+1}}
        \tag{$H_{T+1} \succeq H_t$}  
        \\
        &\leq \frac{  \beta_{T+1} (\delta) }{T}   \sum_{t \notin   \WarmupRounds}
        \left\| 
            \phi \left( x_t, \pi^\star(x_t) \right)
            - \phi \left( x_t, \widehat{\pi}_T(x_t) \right)
        \right\|_{H_t^{-1} }
        \tag{Corollary~\ref{cor:online_CB_PL}, with prob. $1-\delta$}
        .
    \end{align*}
    Define the sigmoid function $\mu(w) = \frac{1}{1 + e^{-w}}$.
    Then, we have
    \begin{align*}
        &\sum_{t \notin   \WarmupRounds}
        \left\| 
            \phi \left( x_t, \pi^\star(x_t) \right)
            - \phi \left( x_t, \widehat{\pi}_T(x_t) \right)
        \right\|_{H_t^{-1} }
        \\
        &= 
        \sum_{t \notin   \WarmupRounds}
        \frac{\dot{\mu} \left(\big(
            \phi \left( x_t, \pi^\star(x_t) \right)
            - \phi \left( x_t, \widehat{\pi}_T(x_t) \right)
        \big)^\top \thetab^\star \right)  }{\dot{\mu} \left(\big(
            \phi \left( x_t, \pi^\star(x_t) \right)
            - \phi \left( x_t, \widehat{\pi}_T(x_t) \right)
        \big)^\top \thetab^\star \right) }
        \left\| 
            \phi \left( x_t, \pi^\star(x_t) \right)
            - \phi \left( x_t, \widehat{\pi}_T(x_t) \right)
        \right\|_{H_t^{-1} }
        \\
        &= 
        (1+e)^2 \!\! \sum_{t \notin   \WarmupRounds}
        \dot{\mu} \left(\big(
            \phi \left( x_t, \pi^\star(x_t) \right)
            - \phi \left( x_t, \widehat{\pi}_T(x_t) \right)
        \big)^\top \thetab^\star \right) 
        \left\| 
            \phi \left( x_t, \pi^\star(x_t) \right)
            - \phi \left( x_t, \widehat{\pi}_T(x_t) \right)
        \right\|_{H_t^{-1} },
    \end{align*}
    where the last equality holds due to the fact that
    \begin{align*}
        &\frac{1 }{\dot{\mu} \left(\big(
            \phi \left( x_t, \pi^\star(x_t) \right)
            - \phi \left( x_t, \widehat{\pi}_T(x_t) \right)
        \big)^\top \thetab^\star \right) }
        \\
        &\quad\quad= \frac{\Big(1 + e^{\big(
            \phi \left( x_t, \pi^\star(x_t) \right)
            - \phi \left( x_t, \widehat{\pi}_T(x_t) \right)
        \big)^\top \thetab^\star} \Big)^2}{e^{\big(
            \phi \left( x_t, \pi^\star(x_t) \right)
            - \phi \left( x_t, \widehat{\pi}_T(x_t) \right)
        \big)^\top \thetab^\star}}
        \\
        &\quad\quad\leq \Big(1 + e^{\big(
            \phi \left( x_t, \pi^\star(x_t) \right)
            - \phi \left( x_t, \widehat{\pi}_T(x_t) \right)
        \big)^\top \thetab^\star} \Big)^2
        \tag{$\big(\phi \left( x_t, \pi^\star(x_t) \right)
            - \phi \left( x_t, \widehat{\pi}_T(x_t) \right)
        \big)^\top \thetab^\star \geq 0$}
        \\
        &\quad\quad\leq \left(1 + e^{\big(
            \phi \left( x_t, \pi^\star(x_t) \right)
            - \phi \left( x_t, \widehat{\pi}_T(x_t) \right)
        \big)^\top (\thetab^\star - \widehat{\thetab}_{T+1})
        } \right)^2
        \tag{$\widehat{\pi}_T(x_t) = \argmax_{a \in \Acal}\phi(x_t,a)^\top \widehat{\thetab}_{T+1} $}
        \\
        &\quad\quad\leq 
        \left(1 + e^{ \|
            \phi \left( x_t, \pi^\star(x_t) \right)
            - \phi \left( x_t, \widehat{\pi}_T(x_t) \right)
            \|_{H_{t}^{-1}} \| \thetab^\star - \widehat{\thetab}_{T+1} \|_{H_{T+1}}
        }\right)^2
        \tag{Hölder's ineq. and $H_{T+1} \succeq H_t$}
        \\
        &\quad\quad\leq 
        \left(1 + e^{ \frac{\beta_{T+1}(\delta)}{3\sqrt{2} K \beta_{T+1}(\delta)} 
        }\right)^2
        \tag{$ t \notin \WarmupRounds$ and Corollary~\ref{cor:online_CB_PL}}
        \\
        &\quad\quad= (1 + e^{1/K})^2
        \leq (1 + e)^2
        . 
    \end{align*}
    Recall that $S^{(j)}_\sigma := \{ \sigma_{j}, \dots, \sigma_{|S|} \}$ denotes the subset of remaining actions in $S$ after removing the first $j-1$ actions, given $S$ and $\sigma$.
    We denote the Multinomial Logit (MNL) model at round $t$ by $P_t(a \mid S; \thetab)$ (see Equation~\eqref{eq:MNL}).
    Then, by our assortment selection rule in Equation~\eqref{eq:S_t_selection_greedy}, we obtain

    \begin{align*}
        \dot{\mu} &\left(\big(
            \phi \left( x_t, \pi^\star(x_t) \right)
            - \phi \left( x_t, \widehat{\pi}_T(x_t) \right)
        \big)^\top \thetab^\star \right) 
        \left\| 
            \phi \left( x_t, \pi^\star(x_t) \right)
            - \phi \left( x_t, \widehat{\pi}_T(x_t) \right)
        \right\|_{H_t^{-1} }
        \\
        &\leq 
        \dot{\mu} \left(\big(
            \phi \left( x_t, \pi^\star(x_t) \right)
            - \phi \left( x_t, \widehat{\pi}_T(x_t) \right)
        \big)^\top \widehat{\thetab}_{t}\right) 
        e^{\left|  \phi \left( x_t, \pi^\star(x_t) \right)
            - \phi \left( x_t, \widehat{\pi}_T(x_t) \right)^\top \big( \widehat{\thetab}_{t} - \thetab^\star \big) \right|}
            \\
        &\quad \quad\cdot \left\| 
            \phi \left( x_t, \pi^\star(x_t) \right)
            - \phi \left( x_t, \widehat{\pi}_T(x_t) \right)
        \right\|_{H_t^{-1} }
        \tag{Lemma~\ref{lemma:dot_sigmoid_bound}}
        \\
        &\leq e \cdot \dot{\mu} \left(\big(
            \phi \left( x_t, \pi^\star(x_t) \right)
            - \phi \left( x_t, \widehat{\pi}_T(x_t) \right)
        \big)^\top \widehat{\thetab}_{t} \right) 
        \left\| 
            \phi \left( x_t, \pi^\star(x_t) \right)
            - \phi \left( x_t, \widehat{\pi}_T(x_t) \right)
        \right\|_{H_t^{-1} }
        \tag{Eqn.~\ref{eq:bound_error}}
        \\
        &= \frac{e}{2}  \cdot
        \EE_{a \sim P_t(\cdot | S^\star_t; \widehat{\thetab}_t)} 
        \left[
            \left\| 
                \phi(x_t, a) - \EE_{a' \sim P_t(\cdot | S^\star_t; \widehat{\thetab}_t)} [\phi(x_t, a')]
            \right\|_{H_t^{-1}}
        \right]
        \tag{$\dot{\mu}(z) = \mu(z)(1-\mu(z))$ and let $S^\star_t := \{ \pi^\star(x_t), \widehat{\pi}_T(x_t) \}$}
        \\
        &=  \frac{e}{|S^\star_t|}
        \EE_{\sigma \sim \PP_t(\cdot | S^\star_t ;\widehat{\thetab}_t )}
        \sum_{j=1}^{|S^\star_t|}
        \EE_{a \sim P_t(\cdot | S^{\star, (j)}_t; \widehat{\thetab}_t)} 
        \left[
            \left\| 
                \phi(x_t, a) - \EE_{a' \sim P_t(\cdot | S^{\star, (j)}_t; \widehat{\thetab}_t)} [\phi(x_t, a')]
            \right\|_{H_t^{-1}}
        \right]
        \tag{$|S^\star_t| = 2$}
        ,
    \end{align*}
    where the second inequality holds since, for any $t \notin \WarmupRounds$ and $a, a' \in \Acal$, the following property holds:
    \begin{align*}
        \Big| \big( & \phi(x_t, a) - \phi(x_t, a') \big)^\top 
         \left( \widehat{\thetab}_{t} - \thetab^\star \right) \Big|
         \\
        &\leq \big\| \phi(x_t, a) - \phi(x_t, a') \big\|_{H_t^{-1} }
        \big\| \widehat{\thetab}_{t} - \thetab^\star \big\|_{H_t}
        \tag{Hölder's inequality}
        \\
        &\leq \frac{1}{3\sqrt{2} K \beta_{T+1}(\delta)}  
        \big\| \widehat{\thetab}_t - \thetab^\star \big\|_{H_t}
        \tag{$t \neq \WarmupRounds$}
        \\
        &\leq \frac{\beta_t(\delta)}{ K \beta_{T+1}(\delta)} 
        \tag{Corollary~\ref{cor:online_CB_PL}}
        \\
        &\leq \frac{1}{K}.
        \numberthis \label{eq:bound_error}
    \end{align*}
    Combining the above results, we get
    \begin{align*}
         &
        \sum_{t \notin   \WarmupRounds}
        \left\| 
            \phi \left( x_t, \pi^\star(x_t) \right)
            - \phi \left( x_t, \widehat{\pi}_T(x_t) \right)
        \right\|_{H_t^{-1} }
        \\
         &\leq
         e(1+e)^2
            \sum_{t \notin   \WarmupRounds}
            \frac{1}{|S^\star_t|}
            \sum_{j=1}^{|S^{\star}_t|}
            \EE_{
            \substack{
            \PP_t(\cdot | S^{\star}_t ;\widehat{\thetab}_t )
            \\
            P_t(\cdot | S^{\star, (j)}_t; \widehat{\thetab}_t)}
            }
            \left[
                \left\| 
                    \phi(x_t, a) - \EE_{ P_t(\cdot | S^{\star, (j)}_t; \widehat{\thetab}_t)} [\phi(x_t, a')]
                \right\|_{H_t^{-1}}
            \right]
        \numberthis  \label{eq:main_proof_major_term_before_S_selection}
        \\
        &\leq 
        e(1+e)^2
         \sqrt{\sum_{t = 1}^T
            \frac{1}{|S_t|}
        }
        \sqrt{
        \sum_{t \notin   \WarmupRounds}
        \frac{|S_t|}{|S^\star_t|}
        \sum_{j=1}^{|S^{\star}_t|}
            \EE_{
            \substack{
            \PP_t(\cdot | S^{\star}_t ;\widehat{\thetab}_t )
            \\
            P_t(\cdot | S^{\star, (j)}_t; \widehat{\thetab}_t)}
            }
            \left[
                \left\| 
                    \phi(x_t, a) - \EE_{ P_t(\cdot | S^{\star, (j)}_t; \widehat{\thetab}_t)} [\phi(x_t, a')]
                \right\|_{H_t^{-1}}^2
            \right]
        }
        \tag{Cauchy-Schwartz ineq., and $\WarmupRounds \subseteq [T]$}
        \\
         &\leq 
        e(1+e)^2 
         \sqrt{\sum_{t = 1}^T
            \frac{1}{|S_t|}
        }
        \sqrt{
        \sum_{t \notin   \WarmupRounds}
        \frac{\cancel{|S_t|}}{\cancel{|\textcolor{blue}{S_t}|}}
        \sum_{j=1}^{|\textcolor{blue}{S_t}|}
            \EE_{
            \substack{
            \PP_t(\cdot | \textcolor{blue}{S_t} ;\widehat{\thetab}_t )
            \\
            P_t(\cdot | \textcolor{blue}{S^{(j)}_t}; \widehat{\thetab}_t)}
            }
            \left[
                \left\| 
                    \phi(x_t, a) - \EE_{ P_t(\cdot | \textcolor{blue}{S^{(j)}_t}; \widehat{\thetab}_t)} [\phi(x_t, a')]
                \right\|_{H_t^{-1}}^2
            \right]
        }
        \tag{Assortment selection rule, Equation~\eqref{eq:S_t_selection_greedy}}
        \\
        &\leq 
        e^2(1+e)^2
         \sqrt{\sum_{t = 1}^T
            \frac{1}{|S_t|}
        }
        \sqrt{
        \sum_{t \notin   \WarmupRounds}
        \sum_{j=1}^{|S_t|}
        \EE_{
            \substack{
            \PP_t(\cdot | S_t ; \textcolor{red}{\thetab^\star} )
            \\
            P_t(\cdot | S_t^{(j)}; \widehat{\thetab}_t)}
            }
            \left[
                \left\| 
                    \phi(x_t, a) - \EE_{ P_t(\cdot | S_t^{(j)}; \widehat{\thetab}_t)} [\phi(x_t, a')]
                \right\|_{H_t^{-1}}^2
            \right]
        }
        \tag{Eqn.~\eqref{eq:PP_hat_to_star}}
        \\
        &\leq \sqrt{3} e^3(1+e)^2
         \sqrt{\sum_{t = 1}^T
            \frac{1}{|S_t|}
        }
        \sqrt{
        \sum_{t \notin   \WarmupRounds}
        \sum_{j=1}^{|S_t|}
        \EE_{
            \substack{
            \PP_t(\cdot | S_t ; \thetab^\star )
            \\
            P_t(\cdot | S_t^{(j)}; \widehat{\thetab}_t)}
            }
            \left[
                \left\| 
                    \phi(x_t, a) - \EE_{ P_t(\cdot | S_t^{(j)}; \widehat{\thetab}_t)} [\phi(x_t, a')]
                \right\|_{\textcolor{orange}{M_t^{-1}} }^2
            \right]
        }
        \tag{Lemma~\ref{lemma:H_t_Lambda_PL}, with prob. $1-\delta$}
        ,
    \end{align*}
    where the second-to-last inequality holds because, for any $t \notin \WarmupRounds$, $S \in \Scal$, and $a \in S$, we have
    \begin{align*}
        P_t(a | S ; \widehat{\thetab}_t)
        &= \frac{\exp \left( \phi(x_t, a)^\top \widehat{\thetab}_t \right)}{
        \sum_{a' \in S} \exp \left(
            \phi(x_t, a')^\top \widehat{\thetab}_t
        \right)
        }
        \\
        &= 
        \frac{\exp \left( \big(\phi(x_t, a) - \phi(x_t, \bar{a}) \big)^\top \widehat{\thetab}_t \right)}{ 1+
        \sum_{a' \in S \setminus \{\bar{a}\}} \exp \left(
            \big(\phi(x_t, a') - \phi(x_t, \bar{a}) \big)^\top \widehat{\thetab}_t
        \right)
        }
        \tag{any $\bar{a} \in S$}
        \\
        &\leq \frac{\exp \left( \big(\phi(x_t, a) - \phi(x_t, \bar{a}) \big)^\top \thetab^\star + 1/K \right)}{ 1+
        \sum_{a' \in S \setminus \{\bar{a}\}} \exp \left(
            \big(\phi(x_t, a') - \phi(x_t, \bar{a}) \big)^\top \thetab^\star - 1/K
        \right)
        }
        \tag{Eqn.~\eqref{eq:bound_error}}
        \\
        &= e^{2/K} P_t(a | S ; \thetab^\star),
    \end{align*}
    which implies that, for the ranking $\sigma = (\sigma_1, \dots, \sigma_{|S|})$, 
    \begin{align*}
        \PP_t (\sigma | S ; \widehat{\thetab}_t)
        &= \prod_{j=1}^{|S|}
        P_t(\sigma_{j} | S^{(j)}_\sigma ; \widehat{\thetab}_t )
        \leq e^{2 |S|/ K} \prod_{j=1}^{|S|}
        P_t(\sigma_{j} | S^{(j)}_\sigma ; \thetab^\star )
        \leq e^2 \prod_{j=1}^{|S|}
        P_t(\sigma_{j} | S^{(j)}_\sigma ; \thetab^\star )
        \\
        &= e^2 \PP_t (\sigma | S ; \thetab^\star)
        \numberthis \label{eq:PP_hat_to_star}
        .
    \end{align*}
    Moreover, by setting $\lambda = d \log T$ and $ T \geq e^{K/d}$, for any $t$, we have
    \begin{align*}
        \sum_{j=1}^{|S_t|}
        \EE_{
            \substack{
            \PP_t(\cdot | S_t ; \thetab^\star )
            \\
            P_t(\cdot | S_t^{(j)}; \widehat{\thetab}_t)}
            }
            \left[
                \left\| 
                    \phi(x_t, a) - \EE_{ P_t(\cdot | S_t^{(j)}; \widehat{\thetab}_t)} [\phi(x_t, a')]
                \right\|_{M_t^{-1}}^2
            \right]
        &\leq
        \frac{2K}{\lambda}
        \leq 2.
        \numberthis \label{eq:samll_EP}
    \end{align*}
    Hence, by the elliptical potential for expected mean-centered uncertainty (Lemma~\ref{lemma:EPL_assortment}), we obtain
    \begin{align*}
    &\frac{1}{T} \sum_{t \notin   \WarmupRounds}
        \big(
            \phi \left( x_t, \pi^\star(x_t) \right)
            - \phi \left( x_t, \widehat{\pi}_T(x_t) \right)
        \big)^\top 
        \left( \thetab^\star
            - \widehat{\thetab}_{T+1}
        \right)
        \\
        &\leq 
         \frac{  \beta_{T+1} (\delta) }{T}   \sum_{t \notin   \WarmupRounds}
        \left\| 
            \phi \left( x_t, \pi^\star(x_t) \right)
            - \phi \left( x_t, \widehat{\pi}_T(x_t) \right)
        \right\|_{H_t^{-1} }
        \\
        &\leq 
        \BigO \!\left(
            \frac{\beta_{T+1} (\delta) }{T} 
             \sqrt{\sum_{t = 1}^T
                \frac{1}{|S_t|}
            }
            \sqrt{
            \sum_{t \notin   \WarmupRounds}
            \sum_{j=1}^{|S_t|}
            \EE_{
                \substack{
                \PP_t(\cdot | S_t ; \thetab^\star )
                \\
                P_t(\cdot | S_t^{(j)}; \widehat{\thetab}_t)}
                }
                \left[
                    \left\| 
                        \phi(x_t, a) - \EE_{ P_t(\cdot | S_t^{(j)}; \widehat{\thetab}_t)} [\phi(x_t, a')]
                    \right\|_{M_t^{-1}}^2
                \right]
            }
        \right)
        \\
        &=
        \BigO \!\left(
            \frac{\beta_{T+1} (\delta) }{T} 
             \sqrt{\sum_{t = 1}^T
                \frac{1}{|S_t|}
            }
            \sqrt{
                d \log \left(1 + \frac{KT}{ d\lambda} \right)
            }
        \right)
        \tag{Lemma~\ref{lemma:EPL_assortment} with Eqn.~\eqref{eq:samll_EP}}
        \\
        &= \BigO \!
        \left(
            \frac{\beta_{T+1} (\delta)  }{T}
            \sqrt{
                \sum_{t=1}^T \frac{1}{|S_t|}
            }
            \cdot 
            \sqrt{d \log \left( KT \right)}
        \right).
        \numberthis \label{eq:main_proof_major_term_final_PL}
    \end{align*}
    By combining Equations~\eqref{eq:main_proof_major_term},~\eqref{eq:main_proof_major_term_low_EP}, and~\eqref{eq:main_proof_major_term_final_PL}, and setting $\beta_{T+1}(\delta) = \mathcal{O}\big( B \sqrt{d \log (KT)} + B\sqrt{\lambda}\big)$, 
    we derive that, with probability at least $1 - 3\delta$ (omitting logarithmic terms and polynomial dependencies on $B$ for brevity),
    \begin{align*}
        \SubOpt(T) 
        &= 
        \BigOTilde \left(
            \frac{ d} {T}
            \sqrt{
                \sum_{t=1}^T \frac{1}{|S_t|}
            }
            + \frac{ d^2 K^4}{\kappa T}
        \right).
    \end{align*}
    Substituting $\delta \leftarrow \frac{\delta}{3} $, we conclude the proof of Theorem~\ref{thm:main_PL}.

\end{proof}
\subsection{Proofs of Lemmas for Theorem~\ref{thm:main_PL}} 
\label{app_subsec:useful_lemmas_thm:main_PL}
\subsubsection{Proof of Lemma~\ref{lemma:H_t_Lambda_PL}}
\label{app_subsubsec:proof_of_lemma:H_t_Lambda_PL}
\begin{proof}[Proof of Lemma~\ref{lemma:H_t_Lambda_PL}]
    Recall the definition of $H_t$.
    \begin{align*}
        H_t 
        &= \sum_{s=1}^{t-1} \sum_{j=1}^{|S_s|}
        \nabla^2 \ell^{(j)}_s (\widehat{\thetab}^{(j+1)}_{s})
        + \lambda \Ib_d
        \succeq
        \!\sum_{s \in [t-1] \setminus \Tcal^w} \sum_{j=1}^{|S_s|}
        \nabla^2 \ell^{(j)}_s (\widehat{\thetab}^{(j+1)}_{s})
        + \lambda \Ib_d
    \end{align*}
    Here, we can equivalently express the MNL loss at step $j$ and round $s$, denoted by $ \nabla^2 \ell^{(j)}_s (\widehat{\thetab}^{(j+1)}_{s})$, as follows:
    \begin{align*}
         \ell^{(j)}_s (\widehat{\thetab}^{(j+1)}_{s})
        &= -  \log \left( 
                \frac{\exp \left( \phi(x_s, \sigma_{sj})^\top \widehat{\thetab}^{(j+1)}_{s}   \right)}{
                \sum_{k=j}^{|S_s|} \exp \left( 
                     \phi(x_s, \sigma_{sk})^\top \widehat{\thetab}^{(j+1)}_{s} )
                \right)
                }
            \right)
        = - \log \left( 
                \frac{\exp \left( a_{sj}   \right)}{
                \sum_{k=j}^{|S_s|} \exp \left( 
                     a_{sk}
                \right)
                }
            \right)
            \\
        &=: \bar{\ell}^{(j)}_s (\ab_s^{(j)}),
        \numberthis \label{eq:MNL_equi_form}
    \end{align*}
    where $a_{sj} = \phi(x_s, \sigma_{sj})^\top \widehat{\thetab}^{(j+1)}_{s}$, 
    $\ab^{(j)}_s = (a_{sk})_{k =j}^{|S_s|} \in \RR^{|S_s| - j +1}$.
    Define the matrix 
    \begin{align*}
        \bm{\Phi}_s^{(j)} =
         \begin{pmatrix} 
         \phi(x_s, \sigma_{sj})^\top \\ \vdots \\ \phi(x_s, \sigma_{s|S_s|})^\top 
         \end{pmatrix} \in \mathbb{R}^{(|S_s| - j + 1) \times d},
    \end{align*}
    where each row corresponds to the feature vector of an action ranked from position $j$ to $|S_s|$ in the ranking $\sigma_s$.
    Moreover, we define 
    $\widehat{a}_{sj} = \phi(x_s, \sigma_{sj})^\top \widehat{\thetab}_s$
    and
    $\widehat{\ab}^{(j)}_s = (\widehat{a}_{sk})_{k =j}^{|S_s|} \in \RR^{|S_s| - j +1}$
    
    Then, using the $\ell_\infty$-norm self-concordant property of the MNL loss~\citep{lee2025improved},
    for any $s \in [t-1] \setminus \WarmupRounds$,
    we obtain
    \begin{align*}
        \nabla^2  \ell^{(j)}_s (\widehat{\thetab}^{(j+1)}_{s})
        &= \left( \bm{\Phi}_s^{(j)} \right)^\top 
        \nabla_{\ab}^2 \, \bar{\ell}^{(j)}_s (\ab_s^{(j)}) 
        \,
        \bm{\Phi}_s^{(j)}
        \tag{Eqn.~\eqref{eq:MNL_equi_form}}
        \\
        &\succeq 
        e^{ -3 \sqrt{2} \|  \ab_s^{(j)} - \widehat{\ab}_s^{(j)}    \|_{\infty} }
        \left( \bm{\Phi}_s^{(j)} \right)^\top 
        \nabla_{\ab}^2 \, \bar{\ell}^{(j)}_s (\widehat{\ab}_s^{(j)}) 
        \,
        \bm{\Phi}_s^{(j)}
        \tag{Lemma~\ref{lemma:self_MNL_hessian_norm}}
        \\
        &\succeq
        \frac{1}{e^2}  \left( \bm{\Phi}_s^{(j)} \right)^\top 
        \nabla_{\ab}^2 \, \bar{\ell}^{(j)}_s (\widehat{\ab}_s^{(j)}) 
        \,
        \bm{\Phi}_s^{(j)}
        \tag{$ \|  \ab_s^{(j)} - \widehat{\ab}_s^{(j)}    \|_{\infty} \leq \frac{2}{3 \sqrt{2}}$}
        \\
        &= \frac{1}{e^2} \nabla^2  \ell^{(j)}_s (\widehat{\thetab}_s)
        \tag{Eqn.~\eqref{eq:MNL_equi_form}}
        ,
    \end{align*}
    where the last inequality holds because, for any  $s \in [t-1] \setminus \WarmupRounds$ and $j \leq |S_s|$, the following holds:
    \begin{align*}
        \|  \ab_s^{(j)} - \widehat{\ab}_s^{(j)}    \|_{\infty}
        &\leq \max_{k=j, \dots, |S_s|} 
        \left|
             \phi(x_k, \sigma_{sk})^\top 
             \left(
                \widehat{\thetab}_s^{(k+1)}
                - \widehat{\thetab}_s
             \right)
        \right|
        \\
        &\leq 
        \max_{k=j, \dots, |S_s|} 
        \left\|
             \phi(x_k, \sigma_{sk})
        \right\|_{H_s^{-1}}
        \left(
            \left\|
                \widehat{\thetab}_s^{(k+1)}
                - \thetab^\star
            \right\|_{H_s}
            +
             \left\|
                 \widehat{\thetab}_s
                - \thetab^\star
            \right\|_{H_s}
        \right)
        \tag{Hölder's inequality}
        \\
        &\leq \frac{1}{3 \sqrt{2} \beta_{T+1}(\delta)}
        \max_{k=j, \dots, |S_s|} 
        \left(\left\|
            \widehat{\thetab}_s^{(k+1)}
            - \thetab^\star
        \right\|_{H_s^{(k+1)}}
        + \left\|
                \widehat{\thetab}_s
                - \thetab^\star
            \right\|_{H_s}
        \right)
        \tag{$ s \notin \WarmupRounds$, $H_s \preceq H_s^{(k+1)}$}
        \\
        &\leq \frac{2\beta_{T+1}(\delta)}{3 \sqrt{2} \beta_{T+1}(\delta)}
        \tag{Corollary~\ref{cor:online_CB_PL}, $\beta_t(\delta)$ is non-decreasing}
        \\
        &= \frac{2}{3\sqrt{2}}.
    \end{align*}
    Therefore, we get
    \begin{align*}
        H_t \succeq
        \!\sum_{s \in [t-1] \setminus \Tcal^w} \sum_{j=1}^{|S_s|}
        \nabla^2 \ell^{(j)}_s (\widehat{\thetab}^{(j+1)}_{s})
        + \lambda \Ib_d
        \succeq
        \frac{1}{e^2}
        \sum_{s \in [t-1] \setminus \Tcal^w} \sum_{j=1}^{|S_s|}
        \nabla^2 \ell^{(j)}_s (\widehat{\thetab}_s)
        + \lambda \Ib_d
        .
    \end{align*}
    Recall the definition of the MNL choice probability~\eqref{eq:MNL}:
    \begin{align*}
        P_s( a | S ; \thetab) := \frac{\exp \left( \phi(x_s, a)^\top \thetab \right)}{
        \sum_{a' \in S} \exp \left(
            \phi(x_s, a')^\top \thetab
        \right)
        }
        ,
        \quad 
        \forall
        a \in S.
    \end{align*}
    Given $S_s = \{  \sigma_{s1}, \dots, \sigma_{s|S_s|}  \} $ and the ranking feedback $\sigma_s$,
    we define  $S^{(j)}_{\sigma_s} :=  \{ \sigma_{sj}, \dots, \sigma_{s|S_s|} \} $ as the subset of remaining actions in $S_s$ after removing the first $j-1$ actions.
    For simplicity, let $\widehat{P}_{s}^{(j)} =  P_{s} (\cdot | S^{(j)}_{\sigma_s}  ; \widehat{\thetab}_t )$.
    Then, we can express the Hessian of the loss as follows:
    \begin{align*}
        \sum_{j=1}^{|S_s|}
        \nabla^2 \ell^{(j)}_s (\widehat{\thetab}_s)
        &= \sum_{j=1}^{|S_s|}
        \sum_{k=j}^{|S_s|}
        \sum_{k'=j}^{|S_s|}
        \frac{\exp\left( 
            \left(
                \phi(x_s, \sigma_{sk})
                + \phi(x_s, \sigma_{sk'})
            \right)^\top \widehat{\thetab}_s 
        \right)}{
        2 \left( 
                \sum_{k'=j}^{|S_s|} \exp\left(
                \phi(x_s, \sigma_{s k'})^\top \widehat{\thetab}_s         
            \right)
          \right)^2
        }
        \cdot
        z_{s k k'} z_{s k k'}^\top
        \\
        &= 
        \frac{1}{2} 
        \sum_{j=1}^{|S_s|}
        \sum_{k=j}^{|S_s|}
        \sum_{k'=j}^{|S_s|}
        P_s (\sigma_{sk} | S_s^{(j)} ; \widehat{\thetab}_s )
        P_s (\sigma_{sk'} | S_s^{(j)}  ; \widehat{\thetab}_s)
        z_{s k k'} z_{s k k'}^\top
        \\
        &= 
        \frac{1}{2} 
        \sum_{j=1}^{|S_s|}
        \sum_{k=j}^{|S_s|}
        \sum_{k'=j}^{|S_s|}
        \EE_{(a, a') \sim \widehat{P}_{s}^{(j)} \times \widehat{P}_{s}^{(j)} }
        \left[ 
            \big(\phi(x_s, a) -  \phi(x_s, a')  \big)
            \big(\phi(x_s, a) -  \phi(x_s, a')  \big)^\top
        \right]
        \\
        &=  \sum_{j=1}^{|S_s|}
            \EE_{\widehat{P}_{s}^{(j)} }
            \left[ 
                \big(\phi(x_s, a) - \EE_{\widehat{P}_{s}^{(j)} }[ \phi(x_s, a') ] \big)
                \big(\phi(x_s, a) - \EE_{\widehat{P}_{s}^{(j)} }[ \phi(x_s, a') ] \big)^\top
            \right]
        \succeq 0 
        \numberthis \label{eq:hessian_to_var}
        ,
    \end{align*}
    where $z_{s k k'} = \phi(x_s, \sigma_{sk}) - \phi(x_s, \sigma_{sk'})$.
    Therefore, $\sum_{j=1}^{|S_s|} \nabla^2 \ell^{(j)}_s (\widehat{\thetab}_s)$ is a random PSD matrix conditional on the assortment $S_s$, where the randomness arises from the realized ranking $\sigma_s$. 
    Note that the true PL distribution $\PP_s (\cdot | S_s ; \thetab^\star)$ is measurable with respect to the filtration $\Fcal_{s-1} = \sigmab( x_1, S_1, \sigmab_1, \dots, x_s, S_s )$.
    Moreover, the maximum eigenvalue of the Hessian satisfies $\lambda_{\max}\left(\sum_{j=1}^{|S_s|} \nabla^2 \ell^{(j)}_s(\widehat{\thetab}_s)\right) \leq K$.
    Then, by the PSD matrices concentration lemma (Lemma~\ref{lemma:concentration_PSD}), with probability $1-\delta$, we have
    \begin{align*}
        H_t &\succeq 
        \frac{1}{e^2}
        \sum_{s \in [t-1] \setminus \Tcal^w} \sum_{j=1}^{|S_s|}
        \nabla^2 \ell^{(j)}_s (\widehat{\thetab}_s)
        + \lambda \Ib_d
        \\
        &= \frac{K}{e^2} 
        \sum_{s \in [t-1] \setminus \Tcal^w} \frac{1}{K} \sum_{j=1}^{|S_s|}
        \nabla^2 \ell^{(j)}_s (\widehat{\thetab}_s)
        + \lambda \Ib_d
        \\
        &\succeq 
        \frac{K}{3 e^2}
        \left(
            \sum_{s \in [t-1] \setminus \Tcal^w}
            \EE_{\sigma \sim \PP_s(\cdot | S_s, \thetab^\star) } \left[ \frac{1}{K}  \sum_{j=1}^{|S_s|}
            \nabla^2 \ell^{(j)}_{s, \sigma} (\widehat{\thetab}_s) \right]
            + \lambda \Ib_d
        \right)
        \tag{Lemma~\ref{lemma:concentration_PSD}, with prob. $1-\delta$}
        \\
        &= 
        \frac{1}{3 e^2}
        \left(
            \sum_{s \in [t-1] \setminus \Tcal^w}
            \sum_{j=1}^{|S_s|}
            \EE_{\sigma \sim \PP_s(\cdot | S_s, \thetab^\star) } \left[\nabla^2 \ell^{(j)}_{s, \sigma} (\widehat{\thetab}_s) \right]
            + \lambda \Ib_d
        \right)
        = \frac{1}{3 e^2}
            M_t
        .
    \end{align*}
    This concludes the proof of Lemma~\ref{lemma:H_t_Lambda_PL}.
\end{proof}
\subsubsection{Proof of Lemma~\ref{lemma:EPL_assortment}}
\label{app_subsubsec:proof_of_lemma:EPL_assortment}
\begin{proof}[Proof of Lemma~\ref{lemma:EPL_assortment}]
    By the definition of $M_t$~\eqref{eq:M_t_def} and Equation~\eqref{eq:hessian_to_var}, for $t \notin \WarmupRounds$, we have
    \begin{align*}
         \operatorname{det} \left( M_{t+1} \right)
        &=  
          \operatorname{det} \left( M_t + \sum_{j=1}^{|S_t|}
            \EE_{\sigma \sim \PP_t(\cdot | S_t, \thetab^\star) } \left[ \nabla^2 \ell^{(j)}_{t, \sigma} (\widehat{\thetab}_t) \right]  \right)
        \\
        &= 
        \operatorname{det} \left( M_t + \sum_{j=1}^{|S_t|}
            \EE_{
                \substack{\sigma \sim \PP_t(\cdot | S_t, \thetab^\star) 
                \\
                a \sim P_t(\cdot | S^{(j)}_{t, \sigma}; \widehat{\thetab}_t)
                }
            } \left[ 
                    \bar{\phi}^{(j)}_{t}(a)
                    \bar{\phi}^{(j)}_{t}(a)^\top
            \right]  \right)
        \tag{Eqn.~\eqref{eq:hessian_to_var}
        }
        \\
        &=
        \operatorname{det} \left( M_{t} \right)
        \left(
            1 + \sum_{j=1}^{|S_t|}
            \EE_{
                \substack{\sigma \sim \PP_t(\cdot | S_t, \thetab^\star) 
                \\
                a \sim P_t(\cdot | S^{(j)}_{t, \sigma}; \widehat{\thetab}_t)
                }
            } \left[ 
                    \| \bar{\phi}^{(j)}_{t}(a) \|_{M_t^{-1}}^2
            \right] 
        \right)
        \\
        &\geq  \operatorname{det} \left( \lambda \Ib_d \right)
        \prod_{s \in [t] \setminus \WarmupRounds}
        \left( 
            1 +  \min \left\{ 1,  \sum_{j=1}^{|S_s|} \EE_{
                \substack{\sigma \sim \PP_s(\cdot | S_s, \thetab^\star) 
                \\
                a \sim P_s(\cdot | S^{(j)}_{s, \sigma}; \widehat{\thetab}_s)
                }
            } \left[ 
                    \| \bar{\phi}^{(j)}_{s}(a) \|_{M_s^{-1}}^2
            \right] \right\}
        \right)
            ,
    \end{align*}
    where, for simplicity, we denote $\bar{\phi}^{(j)}_{t}(a) = \phi(x_t, a) - \EE_{a' \sim P_t(\cdot | S^{(j)}_{t, \sigma}; \widehat{\thetab}_t) }[ \phi(x_t, a')] $.

    For $t \in \WarmupRounds$, it is clear that
    $ \operatorname{det} \left( M_{t+1} \right) =  \operatorname{det} \left( M_{t} \right)$.

    Hence, using the fact that $a \leq 2 \log (1 + a)$ for any $a \in [0,1]$, we get
    \begin{align*}
        \sum_{t \notin \WarmupRounds}
            &\min \left\{ 1,  \sum_{j=1}^{|S_t|} \EE_{
                \substack{\sigma \sim \PP_t(\cdot | S_t, \thetab^\star) 
                \\
                a \sim P_t(\cdot | S^{(j)}_{t, \sigma}; \widehat{\thetab}_t)
                }
            } \left[ 
                    \| \bar{\phi}^{(j)}_{t}(a) \|_{M_t^{-1}}^2
            \right] \right\}
        \\
        &\leq 
        2 \sum_{t \notin \WarmupRounds}
        \log \left(
            1 +  
            \min \left\{ 1,  \sum_{j=1}^{|S_t|} \EE_{
                \substack{\sigma \sim \PP_t(\cdot | S_t, \thetab^\star) 
                \\
                a \sim P_t(\cdot | S^{(j)}_{t, \sigma}; \widehat{\thetab}_t)
                }
            } \left[ 
                    \| \bar{\phi}^{(j)}_{t}(a) \|_{M_t^{-1}}^2
            \right] \right\}
        \right)
        \\
        &\leq 
        2 
        \log \left(
            \frac{\operatorname{det} \left( M_{T+1} \right)}{\operatorname{det} \left( \lambda \Ib_d  \right)}
        \right)
        \\
        &\leq 
        2 d
        \log \left(
            1 + \frac{4 KT}{d \lambda}
        \right),
    \end{align*}
    where the last inequality holds because
    \begin{align*}
        \operatorname{det} \left( M_{T+1} \right)
        &\leq 
        \left( \frac{\lambda_1 + \cdots + \lambda_d}{d} \right)^d 
        \tag{$\lambda_1, \cdots, \lambda_d$ are eigenvalues of $M_{T+1}$, AM-GM ineq.}
        \\
        &\leq\left( \frac{\operatorname{trace}(M_{T+1})}{d} \right)^d
        \\
        &= \left( \frac{ \lambda d + \sum_{t=1}^T \sum_{j=1}^{|S_t|} 
           \EE \|\bar{\phi}^{(j)}_{t}(a)\|_2^2 }{d} \right)^d
        \leq \left( \lambda + \frac{ KT}{d} \right)^d
        \tag{$\EE \|\bar{\phi}^{(j)}_{t}(a)\|_2^2  \leq 1$}
        .
    \end{align*}
    This concludes the proof of Lemma~\ref{lemma:EPL_assortment}.
\end{proof}
%
\subsubsection{Proof of Lemma~\ref{lemma:bound_Tw_PL}}
\label{app_subsubsec:proof_of_lemma:bound_Tw_PL}
\begin{proof}[Proof of Lemma~\ref{lemma:bound_Tw_PL}]
    Let
    \begin{align*}
        (\tilde{a}_t, \bar{a}_t) =  \argmax_{a,a' \in \Acal} \dot\mu\left(\big(\phi(x_t, a) - \phi(x_t, a')\big)^\top \widehat{\thetab}_t  \right) \| \phi(x_t, a) - \phi(x_t, a') \|_{H_t^{-1} }^2,
    \end{align*}
    where $\mu(w) = \frac{1}{1 + e^{-w}}$ denote the sigmoid function.
    Note that, by our assortment selection rule in Equation~\eqref{eq:S_t_selection_greedy}, the two actions $a, a' \in \Acal$ that maximize
    $\dot\mu\left(\big(\phi(x_t, a) - \phi(x_t, a')\big)^\top \widehat{\thetab}_t  \right)
        \| \phi(x_t, a) - \phi(x_t, a') \|^2_{H_t^{-1} }$
    are always included in $S_t$.
    It is because
    \begin{align*}
        \dot\mu&\left(\big(\phi(x_t, a) - \phi(x_t, a')\big)^\top \widehat{\thetab}_t  \right)
        \| \phi(x_t, a) - \phi(x_t, a') \|^2_{H_t^{-1} }
        \\
        &= 
       2  
        \EE_{a \sim P_t(\cdot | \{ a, a' \} ; \widehat{\thetab}_t )}
        \left[ \left\|
            \phi(x_t, a) - 
            \EE_{a' \sim P_t(\cdot | \{ a, a' \} ; \widehat{\thetab}_t )}
            [\phi(x_t, a')]
        \right\|_{H_t^{-1}}^2 \right]
        \\
        &= 
        2 \EE_{
        \substack{
        \sigma \sim \PP_t(\cdot | \{a, a' \} ; \widehat{\thetab}_t  )
        \\
        a \sim P_t(\cdot | \{ a, a' \} ; \widehat{\thetab}_t )}}
        \left[ \left\|
            \phi(x_t, a) - 
            \EE_{a' \sim P_t(\cdot | \{ a, a' \} ; \widehat{\thetab}_t )}
            [\phi(x_t, a')]
        \right\|_{H_t^{-1}}^2 \right]
        = 2 f_t(\{a, a'\} )
        ,
    \end{align*}
    which directly implies that $\tilde{a}_t, \ReferenceAction_t \in S_t$.

    For simplicity, we denote $z_{s k k'} = \phi(x_s, \sigma_{sk}) - \phi(x_s, \sigma_{sk'})$.
    Then, by the definition of $H_t $, we have
    \begin{align*}
        H_{t}
        &= \sum_{s =1}^{t-1} \sum_{j=1}^{|S_s|} \nabla^2 \ell_s^{(j)} (\widehat{\thetab}_{s}^{(j+1)}) + \lambda \Ib_d
        \\
        &= \sum_{s =1}^{t-1} 
        \sum_{j=1}^{|S_s|}
        \sum_{k=j}^{|S_s|}
        \sum_{k'=j}^{|S_s|}
        \frac{\exp\left( 
            \left(
                \phi(x_s, \sigma_{sk})
                + \phi(x_s, \sigma_{sk'})
            \right)^\top \widehat{\thetab}_{s}^{(j+1)} 
        \right)}{
        2 \left( 
                \sum_{k'=j}^{|S_s|} \exp\left(
                \phi(x_s, \sigma_{s k'})^\top \widehat{\thetab}_{s}^{(j+1)}          
            \right)
          \right)^2
        }
        \cdot
        z_{s k k'} z_{s k k'}^\top
        + \lambda \Ib_d 
        \\
        &\succeq
        \frac{e^{-4B}}{2 K^2}
        \sum_{s =1}^{t-1}
        \sum_{j=1}^{|S_s|}
        \sum_{k=j}^{|S_s|}
        \sum_{k'=j}^{|S_s|}
        z_{s k k'} z_{s k k'}^\top
        + \lambda \Ib_d 
        \\
        &\succeq
        \frac{e^{-4B}}{2 K^2}
        \sum_{s=1}^{t-1}
        \big(
            \phi(x_s, \tilde{a}_s) - \phi(x_s, \ReferenceAction_s)
        \big)
        \big(
            \phi(x_s, \tilde{a}_s) - \phi(x_s, \ReferenceAction_s)
        \big)^\top
        +  \lambda \Ib_d
        \tag{ $\tilde{a}_s, \bar{a}_s\in S_s$}
        \\
        &\succeq \frac{e^{-4B}}{2 K^2} 
        \Bigg(\sum_{s \in \WarmupRounds_t}
        \dot\mu\!\left(\big(\phi(x_s, \tilde{a}_s) - \phi(x_s, \ReferenceAction_s)\big)^\top \widehat{\thetab}_s  \right)
        \big(
            \phi(x_s, \tilde{a}_s) - \phi(x_s, \ReferenceAction_s)
        \big)
        \big(
            \phi(x_s, \tilde{a}_s) - \phi(x_s, \ReferenceAction_s)
        \big)^\top
        \\
        &\quad\quad\quad\quad+  \lambda \Ib_d \Bigg)
        ,
        \numberthis \label{eq:H_t_lower_Lambda_w_PL}
    \end{align*}
    where
    the last inequality holds since $\dot\mu\!\left(\big(\phi(x_s, a) - \phi(x_s, a')\big)^\top \widehat{\thetab}_s  \right) \leq \frac{1}{4}$.
    In addition, we denote $ \WarmupRounds_t = \WarmupRounds \setminus \{t, \dots, T\}$ in the last inequality.
    To better presentation, we define
    \begin{align*}
        \Lambda^{w}_t
        :=& \sum_{s \in \WarmupRounds_t}
        \dot\mu\!\left(\big(\phi(x_s, \tilde{a}_s) - \phi(x_s, \ReferenceAction_s)\big)^\top \widehat{\thetab}_s  \right)
        \big(
            \phi(x_s, \tilde{a}_s) - \phi(x_s, \ReferenceAction_s)
        \big)
        \big(
            \phi(x_s, \tilde{a}_s) - \phi(x_s, \ReferenceAction_s)
        \big)^\top
        \\
        &+ \lambda \Ib_d.
    \end{align*}
    On the other hand, by setting $\lambda \geq 1$, for any $t$, we can ensure that  
    \begin{align*}
         \dot\mu\left(\big(\phi(x_t, \tilde{a}_t) - \phi(x_t, \ReferenceAction_t)\big)^\top \widehat{\thetab}_t  \right) \| \phi(x_t, \tilde{a}_t) - \phi(x_t, \ReferenceAction_t) \|^2_{\big(\Lambda^w_t\big)^{-1}}
        &\leq \frac{1}{4} \cdot \frac{2}{\lambda}
        \tag{$\dot{\mu}(\cdot) \leq \frac{1}{4}$}
        \\
        &\leq 1 
        .
        \numberthis \label{eq:ep_max}
    \end{align*}
    Then, we can derive that
    \begin{align*}
        \sum_{t \in \WarmupRounds} &\max_{a, a' \in \Acal} \| \phi(x_t, a) - \phi(x_t, a') \|^2_{H_t^{-1} }
        \\
        &= \sum_{t \in \WarmupRounds} \max_{a, a' \in \Acal} \frac{\dot\mu\left(\big(\phi(x_t, a) - \phi(x_t, a')\big)^\top \widehat{\thetab}_t \right)}{\dot\mu\left(\big(\phi(x_t, a) - \phi(x_t, a')\big)^\top \widehat{\thetab}_t \right)}
        \| \phi(x_t, a) - \phi(x_t, a') \|^2_{H_t^{-1} }
        \\
        &\leq 4 e^{2B} 
        \sum_{t \in \WarmupRounds} \max_{a, a' \in \Acal} 
        \dot\mu\left(\big(\phi(x_t, a) - \phi(x_t, a')\big)^\top \widehat{\thetab}_t  \right)
        \| \phi(x_t, a) - \phi(x_t, a') \|^2_{H_t^{-1} }
        \tag{Assumption~\ref{assum:linear_reward}}
        \\
        &=
           4 e^{2B} 
            \sum_{t \in \WarmupRounds}
            \dot\mu\left(\big(\phi(x_t, \tilde{a}_t) - \phi(x_t, \ReferenceAction_t)\big)^\top \widehat{\thetab}_t  \right)
            \| \phi(x_t, \tilde{a}_t) - \phi(x_t, \ReferenceAction_t) \|^2_{H_t^{-1} }
        \\
        &\leq 8 K^2 e^{6B}
         \sum_{t \in \WarmupRounds}
         \dot\mu\left(\big(\phi(x_t, \tilde{a}_t) - \phi(x_t, \ReferenceAction_t)\big)^\top \widehat{\thetab}_t  \right)
         \| \phi(x_t, \tilde{a}_t ) - \phi(x_t, \ReferenceAction_t) \|^2_{\big(\Lambda^w_t\big)^{-1}}
        \tag{Eqn.~\eqref{eq:H_t_lower_Lambda_w_PL}}
        \\
        &= 
         8 K^2 e^{6B}
         \sum_{t \in \WarmupRounds}
         \min \left\{ 1,  \dot\mu\left(\big(\phi(x_t, \tilde{a}_t) - \phi(x_t, \ReferenceAction_t)\big)^\top \widehat{\thetab}_t  \right) \| \phi(x_t, \tilde{a}_t ) - \phi(x_t, \ReferenceAction_t) \|^2_{\big(\Lambda^w_t\big)^{-1}} \right\}
         \tag{Eqn.~\eqref{eq:ep_max} }
         \\
         &\leq 16 K^2 e^{6B}
         d \log \left(
            1 + \frac{2 T}{d \lambda}
         \right).
         \tag{Lemma~\ref{lemma:EPL}}
    \end{align*}
    On the other hand, for $t \in \WarmupRounds $, we know that
    \begin{align*}
         \sum_{t \in \WarmupRounds} \max_{a, a' \in \Acal} \| \phi(x_t, a) - \phi(x_t, a') \|^2_{H_t^{-1} }
         \geq  \frac{|\WarmupRounds|}{18 K^2 \beta_{T+1}(\delta)^2}.
    \end{align*}
    By combining the two results above, and setting $\kappa = e^{-6}$, we obtain
    \begin{align*}
        | \WarmupRounds |
        \leq \frac{288 K^4}{\kappa} \beta_{T+1}(\delta)^2
         d \log \left(
            1 + \frac{2 KT}{d \lambda}
         \right),
    \end{align*}
    which concludes the proof.
\end{proof}

\section{Proof of Theorem~\ref{thm:main_RB}}
\label{app_sec:proof_upper_RB}
\subsection{Main Proof of Theorem~\ref{thm:main_RB}}
\label{app_subsec:proof_main_thm}
In this section, we present the proof of Theorem~\ref{thm:main_RB}, which is obtained by using the RB loss~\eqref{eq:RB_loss} instead of the PL loss~\eqref{eq:PL_loss}.
Note that this approach is based on the concept of \textit{rank-breaking} (RB), which decomposes (partial) ranking data into individual pairwise comparisons, treats each comparison as independent, and has been extensively studied in previous works~\citep{azari2013generalized, khetan2016data, jang2017optimal, saha2024stop}.
Moreover, this RB approach is applied in the current RLHF for
LLM (e.g.,~\citet{ouyang2022training}) and is also studied theoretically in~\citet{zhu2023principled} under the offline setting.

\textbf{RB loss and OMD.}
We begin by recalling the loss function and the parameter update rule.
Specifically, we use the PL loss defined in Equation~\eqref{eq:RB_loss} and update the parameter according to Equation~\eqref{eq:online_update_RB}.
\begin{align*}
    \ell_t(\thetab) 
    &:= 
    \sum_{j=1}^{|S_t|-1} \!\! \sum_{k=j+1}^{|S_t|}
    \underbrace{-
    \log \left( 
        \frac{\exp \left( \phi(x_t, \sigma_{tj})^\top \thetab \right)}{
        \exp \left( \phi(x_t, \sigma_{tj})^\top \thetab \right) 
        + \exp \left( \phi(x_t, \sigma_{tk})^\top \thetab \right)
        }
    \right)}_{=: \ell^{(j,k)}_t (\thetab)}
    =  \sum_{j=1}^{|S_t|-1} \!\! \sum_{k=j+1}^{|S_t|} 
    \ell^{(j,k)}_t (\thetab).
\end{align*}
and
\begin{align*}
    \widehat{\thetab}_{t}^{(j, k+1)} 
    &= \argmin_{\thetab \in \Theta }   \, \langle \nabla \ell_t^{(j,k)} (\widehat{\thetab}_{ t }^{(j,k)} ), \thetab \rangle
    + \frac{1}{2 \eta} \| \thetab - \widehat{\thetab}_t^{(j,k)} \|_{\tilde{H}_{t}^{(j,k)}}^2,
    \quad \, 1 \leq j < k \leq |S_t|,
\end{align*} 
where if $k=|S_t|$, we set $\widehat{\thetab}_{t}^{(j, k+1)} = \widehat{\thetab}_{t}^{(j+1, j+2)}$, and for the final pair, let
$\widehat{\thetab}_{t}^{(|S_t|-1, |S_t|+1)}= \widehat{\thetab}_{t+1}^{(1,2)} $.
Also, the matrix $\tilde{H}_{t}^{(j,k)}$ is defined as 
$\tilde{H}_{t}^{(j,k)} := H_{t} + \eta \sum_{(j',k') \leq (j,k)}  \!\ \nabla^2 \ell_{t}^{(j',k')}(\widehat{\thetab}^{(j',k')}_t)$
\footnote{We write $(j',k') \le (j,k)$ to indicate lexicographic order, i.e., $j' < j$ or $j' = j$ and $k' \le k$.}
, where
\begin{align*}
    H_{t}&:= \sum_{s =1}^{t-1} \sum_{j=1}^{|S_s|-1} 
    \!\! \sum_{k=j+1}^{|S_s|} 
    \nabla^2 \ell_s^{(j,k)} (\widehat{\thetab}_{s}^{(j,k+1)}) + \lambda \Ib_d
    \\
    &= \sum_{s =1}^{t-1} \sum_{j=1}^{|S_s|-1} 
    \!\! \sum_{k=j+1}^{|S_s|} 
    \dot{\mu}
    \left(
        \big(\phi(x_t, \sigma_j) - \phi(x_t, \sigma_k)\big)^\top
        \widehat{\thetab}_{s}^{(j,k+1)}
    \right)
    \\
    &\quad\quad\quad\quad\quad\quad\quad
    \cdot \big(\phi(x_t, \sigma_j) - \phi(x_t, \sigma_k)\big)
    \big(\phi(x_t, \sigma_j) - \phi(x_t, \sigma_k)\big)^\top
    + \lambda \Ib_d
    .
\end{align*}

\textbf{Online confidence bound for RB loss.}
Now, we introduce the online confidence bound for RB loss.
Since the total number of updates up to round $t$ is $\sum_{s=1}^{t} \binom{|S_s|}{2}$, a modification of Lemma~\ref{lemma:online_CB} yields the following result:
\begin{proposition} [Online confidence bound for RB loss]
    \label{prop:online_CB_RB}
    Let $\delta \in (0, 1]$.
    We set $\eta = (1+ 3\sqrt{2} B)/2$ and $\lambda = \max \{ 12 \sqrt{2} B \eta, 144 \eta d, 2 \} $.
    If $ K^2 \leq (2 \sqrt{6} + 12 \sqrt{3}B)d$,
    then under Assumption~\ref{assum:linear_reward}, with probability at least $1 - \delta$, we have
    \begin{align*}
        \| \widehat{\thetab}_t^{(j,k)} -\thetab^\star \|_{H_t^{(j, k)}} 
        \leq \beta_t (\delta)
        = \BigO\left(
        B
        \sqrt{
            d \log (t K/\delta)
            }
            + B \sqrt{\lambda}
        \right),
        \quad \forall t \geq 1,\, 1 \leq  j < k \leq |S_t|
        ,
    \end{align*}
    where $H_t^{(j,k)} := H_t + 
    \sum_{(j',k') < (j,k)}
    \nabla^2 \ell_t^{(j', k')} (\widehat{\thetab}_{t}^{(j', k'+1)}) + \lambda \Ib_d$ and
    $\widehat{\thetab}^{(1,2)}_t = \widehat{\thetab}_t$.
\end{proposition}
The proof is deferred to Appendix~\ref{app_subsubsec:proof_of_prop:online_CB_RB}.

\textbf{Useful definitions.}
We define the set of \textit{warm-up rounds} $\WarmupRounds$ in a similar manner to the analysis of Theorem \ref{thm:main_PL}:
\begin{align*}
    \WarmupRounds
    &:= \left\{
        t \in [T]:
        \max_{a, a' \in \Acal} \| \phi(x_t, a) - \phi(x_t, a') \|_{H_t^{-1} } \geq 
        \frac{1}{ \beta_{T+1}(\delta)}
    \right\} \tag{warm-up rounds}
    .
\end{align*} 
Moreover, we define the matrix
\begin{align*}
    \Lambda_t =  \!\!&\sum_{s \in [t-1] \setminus \Tcal^w} 
        \!\sum_{a, a'\in S_s}            
            \dot{\mu}
            \left(
                \big(\phi(x_t, a) - \phi(x_t, a')\big)^\top
                \widehat{\thetab}_{s}
            \right)
            \big(\phi(x_t, a) - \phi(x_t, a')\big)
            \big(\phi(x_t, a) - \phi(x_t, a')\big)^\top
            \\
            &+ \lambda \Ib_d
    \numberthis \label{eq:lambda_t_def}
\end{align*}
%
%
\textbf{Key lemmas.}
The relationship between $H_t$ and $\Lambda_t$ is as follows:
\begin{lemma} 
\label{lemma:H_t_Lambda}
    Let $\Lambda_t$ be defined as in Equation~\eqref{eq:lambda_t_def}.
    Then, for all $t \in [T]$, 
    we have 
    \begin{align*}
        H_t \succeq  \frac{1}{2 e^2}
            \Lambda_t.
    \end{align*}
\end{lemma}
The proof is deferred to Appendix~\ref{app_subsubsec:proof_of_lemma:H_t_Lambda}.

The size of the set $\WarmupRounds$ is bounded as described in the following lemma:
\begin{lemma} 
\label{lemma:bound_Tw} 
    Let
        $\WarmupRounds
        = \big\{
            t \in [T]:
            \max_{a, a' \in \Acal} \| \phi(x_t, a) - \phi(x_t, a') \|_{H_t^{-1} } \geq 
            \frac{1}{\beta_{T+1}(\delta)}
        \big\}$.
    Define $\kappa := e^{-6B}$.
    Set $\lambda \geq 1$.
    Then, the size of the set $\WarmupRounds $ is bounded as follows:
    \begin{align*}
         \left| \WarmupRounds \right|
         \leq  \frac{32}{\kappa} \beta_{T+1}(\delta)^2
         d \log \left(
            1 + \frac{2 T}{d \lambda}
         \right)
         .
    \end{align*}
\end{lemma}
The proof is deferred to Appendix~\ref{app_subsubsec:proof_of_lemma:bound_Tw}.
Note that the bound on the cardinality of 
$\WarmupRounds$ is tighter than in the PL loss case (Lemma~\ref{lemma:bound_Tw_PL}), by up to a factor of $K^4$ (and up to logarithmic factors in $K$).

We are now ready to provide the proof of Theorem~\ref{thm:main_RB}.
%
%
\begin{proof}[Proof of Theorem~\ref{thm:main_RB}]
    The overall proof structure is similar to that of Theorem~\ref{thm:main_PL}.
    We start from Equation~\eqref{eq:main_proof_major_term_low_EP}, but apply Lemma~\ref{lemma:bound_Tw} instead of Lemma~\ref{lemma:bound_Tw_PL}.
    Then, with probability at least $1-\delta$, we have
    \begin{align*}
        \SubOpt(T) 
        &= \EE_{x \sim \rho} \left[
            \big(
                \phi \left( x, \pi^\star(x) \right)
                - \phi \left( x, \widehat{\pi}_T(x) \right)
            \big)^\top 
            \thetab^\star
        \right]
        \\
        &\leq  
        \BigOTilde \left( \frac{1}{\sqrt{T}} \right) 
        + \BigO \left(\frac{ B}{ \kappa T} \beta_{T+1}(\delta)^2
         d \log \left(
            1 + \frac{ T}{d \lambda}
         \right)
         \right)
         \tag{Lemma~\ref{lemma:bound_Tw}}
        \\
        &+
        \frac{1}{T} \sum_{t \notin \WarmupRounds}
        \big(
            \phi \left( x_t, \pi^\star(x_t) \right)
            - \phi \left( x_t, \widehat{\pi}_T(x_t) \right)
        \big)^\top 
        \left(
            \thetab^\star
            - \widehat{\thetab}_{T+1}
        \right).
        \numberthis \label{eq:main_proof_major_term_low_EP_RB}
    \end{align*}
    To further bound the last term of Equation~\eqref{eq:main_proof_major_term_low_EP_RB}, 
    by following the similar reasoning from Equation~\eqref{eq:main_proof_major_term_low_EP} to Equation~\eqref{eq:main_proof_major_term_before_S_selection}, with probability at least $1-\delta$, 
    we obtain
    \begin{align*}
        \sum_{t \notin   \WarmupRounds}
        &\left\| 
            \phi \left( x_t, \pi^\star(x_t) \right)
            - \phi \left( x_t, \widehat{\pi}_T(x_t) \right)
        \right\|_{H_t^{-1} }
        \\
         &\leq
         e(1+e)^2
            \sum_{t \notin   \WarmupRounds}
            \frac{1}{|S^\star_t |} 
             \sum_{a, a' \in S^\star_t }
             \dot{\mu}\big( \psi_t(a,a')^\top \widehat{\thetab}_t \big)
             \left\| 
                        \psi_t(a,a')
                    \right\|_{H_t^{-1}}
        \tag{$S^\star_t := \{ \pi^\star(x_t), \widehat{\pi}_T(x_t) \}$, and let $\psi_t(a,a') = \phi(x_t, a) - \phi(x_t, a')$}
        \\
        &\leq
        e(1+e)^2
         \sqrt{\sum_{t = 1}^T
            \frac{1}{|S_t|}
        }
        \sqrt{
        \sum_{t \notin   \WarmupRounds}
        \frac{|S_t|}{|S^\star_t|}
            \sum_{a, a' \in S^\star_t }
             \dot{\mu}\big( \psi_t(a,a')^\top \widehat{\thetab}_t \big)
             \left\| 
                        \psi_t(a,a')
                    \right\|_{H_t^{-1}}^2
        }
        \tag{Cauchy-Schwartz ineq., and $\WarmupRounds \subseteq [T]$}
        \\
        &\leq
         e(1+e)^2 
         \sqrt{\sum_{t = 1}^T
            \frac{1}{|S_t|}
        }
        \sqrt{
        \sum_{t \notin   \WarmupRounds}
        \frac{\cancel{|S_t|}}{\cancel{|\textcolor{blue}{S_t}|}}
            \sum_{a, a' \in \textcolor{blue}{S_t}  }
             \dot{\mu}\big( \psi_t(a,a')^\top \widehat{\thetab}_t \big)
             \left\| 
                        \psi_t(a,a')
                    \right\|_{H_t^{-1}}^2
        }
        \\
        &\leq
         \sqrt{2}e^2(1+e)^2 
         \sqrt{\sum_{t = 1}^T
            \frac{1}{|S_t|}
        }
        \sqrt{
            \sum_{t \notin   \WarmupRounds}
            \sum_{a, a' \in S_t  }
             \dot{\mu}\big( \psi_t(a,a')^\top \widehat{\thetab}_t \big)
             \left\| 
                        \psi_t(a,a')
                    \right\|_{\textcolor{red}{\Lambda_t^{-1}}}^2
        }
        \tag{Lemma~\ref{lemma:H_t_Lambda}}
        .
    \end{align*}
    Moreover, by setting $\lambda = d \log T$ and $ T \geq e^{K^2/d}$, for any $t$, we have
    \begin{align*}
        \sum_{a, a' \in S_t  }
             \dot{\mu}\big( \psi_t(a,a')^\top \widehat{\thetab}_t \big)
             \left\| 
                        \psi_t(a,a')
                    \right\|_{\Lambda_t^{-1}}^2
        \leq \frac{2K^2}{\lambda} \leq 2.
        \numberthis \label{eq:samll_EP_RB}
    \end{align*}
    Hence, by the elliptical potential lemma (Lemma~\ref{lemma:EPL}), we derive that
    \begin{align*}
        \frac{1}{T} \sum_{t \notin   \WarmupRounds}
        &\big(
            \phi \left( x_t, \pi^\star(x_t) \right)
            - \phi \left( x_t, \widehat{\pi}_T(x_t) \right)
        \big)^\top 
        \left( \thetab^\star
            - \widehat{\thetab}_{T+1}
        \right)
        \\
        &\leq 
        \frac{  \beta_{T+1} (\delta) }{T}   \sum_{t \notin   \WarmupRounds}
        \left\| 
            \phi \left( x_t, \pi^\star(x_t) \right)
            - \phi \left( x_t, \widehat{\pi}_T(x_t) \right)
        \right\|_{H_t^{-1} }
        \\
        &= 
        \BigO \!\left(
            \frac{\beta_{T+1} (\delta) }{T} 
             \sqrt{\sum_{t = 1}^T
                \frac{1}{|S_t|}
            }
            \cdot
                \sqrt{d \log \left( KT \right)}
        \right).
        \numberthis \label{eq:main_proof_major_term_final}
    \end{align*}
    By plugging Equation~\eqref{eq:main_proof_major_term_final} into Equation~\eqref{eq:main_proof_major_term_low_EP_RB} and setting $\beta_{T+1}(\delta) = \mathcal{O}\big( B \sqrt{d \log (KT)} + B\sqrt{\lambda} \big)$,
    then, with probability at least $1 - 2\delta$, we derive that
    \begin{align*}
        \SubOpt(T) 
        &= 
        \BigOTilde \left(
            \frac{d}{T}
            \sqrt{
                \sum_{t=1}^T \frac{1}{|S_t|}
            }
            + \frac{d^2}{\kappa T}
        \right).
    \end{align*}
    Substituting $\delta \leftarrow \frac{\delta}{2} $, we conclude the proof of Theorem~\ref{thm:main_RB}.
\end{proof}

\subsection{Proofs of Propositions and Lemmas for Theorem~\ref{thm:main_RB}} 
\label{app_subsec:useful_lemmas_thm:main_RB}
\subsubsection{Proof of Proposition~\ref{prop:online_CB_RB}}
\label{app_subsubsec:proof_of_prop:online_CB_RB}
\begin{proof} [Proof of Proposition~\ref{prop:online_CB_RB}]
    Since the rank-breaking pairs generated from the same set $S_t$ are not mutually independent, Lemma~\ref{lemma:online_CB} cannot be applied directly.
    Therefore, instead of Lemma C.2 of~\citet{lee2025improved}, we use the following lemma.
    \begin{lemma} \label{lemma:online_regret_intermediate_(a)}
        Let $\delta \in (0,1]$.
        We define a pseudo-inverse function of $\mu_t^{(j,k)}(\cdot)$ as $\mu_t^{(j,k),+} : \RR \rightarrow \RR$, where $[\mu_t^{(j,k),+}(\qb)]_i = \log \left( q_i / (1- \| \qb \|_1)\right)$ for any $\qb \in \{ \pb \in [0,1] \mid \| \pb \|_1 < 1 \}$.
        Let $\psi_t(a,a') = \phi(x_t, a) - \phi(x_t, a')$,
        and the intermediary parameter denote 
        $ \tilde{z}^{(j,k)}_t := \mu_t^{(j,k),+} \left( \EE_{\thetab \sim P_t^{(j,k)}} \left[\mu_t^{(j,k)}\left( \psi_t(\sigma_j, \sigma_k)^\top \thetab   \right) \right] \right)$,
        where $P_t^{(j,k)} := 
        \mathcal{N} \left( \widehat{\thetab}_t^{(j,k)}, c H_t^{-1} \right) $ is a multivariate normal distribution with mean $\widehat{\thetab}_t^{(j,k)}$ and covariance $cH_t^{-1}$ ($c > 0$).
        Then,
        for any $t \in [T]$, with probability at least $1-\delta$, we have
        \begin{align*}
            \sum_{s=1}^t 
            \sum_{j< k \leq |S_s|}
            \left(
                \ell_{s}^{(j,k)}(\thetab^\star)
                - 
                \bar{\ell}_{s}^{(j,k)}(\tilde{z}_s^{(j,k)}) 
            \right)
            \leq \frac{K(K-1)}{2} \log \frac{1}{\delta},
        \end{align*}
        where $ \bar{\ell}_t(z) = - \log \left( \frac{\exp( z  )}{ 1 +\exp( z )}\right)$.
    \end{lemma}
    \begin{proof} [Proof of Lemma~\ref{lemma:online_regret_intermediate_(a)}]
        The proof of Lemma~\ref{lemma:online_regret_intermediate_(a)} mostly follows the analysis of Lemma~C.2 in~\citet{lee2025improved}.
        However, unlike their setting, the rank-breaking pairs generated from the same assortment 
        $S_t$ are generally dependent.
        Therefore, instead of applying Ville’s inequality~\citep{ville1939etude} to individual pairs, we apply it at the \textit{block level}, where each block corresponds to one round $t$, by aggregating pairwise likelihood ratios through a mixture construction. 
        This preserves the supermartingale property and allows us to obtain the desired bound.
        
        For simplicity, we write
        \[p^{(j,k)}_t := \mu_t^{(j,k)}\left( \psi_t(\sigma_j, \sigma_k)^\top \thetab^\star   \right),
        \quad
        \text{and}
        \quad
        \tilde{p}^{(j,k)}_t := \EE_{\thetab \sim P_t^{(j,k)}} \left[\mu_t^{(j,k)}\left( \psi_t(\sigma_j, \sigma_k)^\top \thetab   \right) \right].
        \]
        We define 
        \[
            M_t := \sum_{j< k \leq |S_t|} \frac{1}{\binom{|S_t|}{2}}
            \exp\left(
                 \ell_{t}^{(j,k)}(\thetab^\star)
                - 
                \ell_{t}^{(j,k)}(\tilde{z}_s^{(j,k)}) 
            \right)
            = \sum_{j< k \leq |S_t|} \frac{1}{\binom{|S_t|}{2}} \cdot
            \frac{\tilde{p}^{(j,k)}_t (Y_{t}^{(j,k)})  }{p^{(j,k)}_t (Y_{t}^{(j,k)}  )},
        \]
        where 
        $p^{(j,k)}_t (Y_{t}^{(j,k)} ) = (p^{(j,k)}_t)^{Y_{t}^{(j,k)}} (1- p^{(j,k)}_t)^{1-Y_{t}^{(j,k)}}\!$.
        Then, 
        we have
        \[
            \EE [M_t \mid \Fcal_{t-1}]
            =   \frac{1}{\binom{|S_t|}{2}} \sum_{j < k \leq |S_t|}
            \EE \left[  \frac{\tilde{p}^{(j,k)}_t (Y_{t}^{(j,k)})  }{p^{(j,k)}_t (Y_{t}^{(j,k)}  )} 
            \mid \Fcal_{t-1} \right]
            =   \frac{1}{\binom{|S_t|}{2}} \sum_{j < k \leq |S_t|} 
            \sum_{t \in \{0,1 \} } \tilde{p}^{(j,k)}_t (y) 
            = 1.
        \]
        Let $A_t := \Pi_{s=1}^t M_s$ and $A_0 := 1$.
        Then, it follows that
        $(A_t)_{t \geq 1}$ is a  nonnegative supermartingale with respect to the filtration $\Fcal_{t-1} = \sigmab( x_1, S_1, \sigmab_1, \dots, x_t, S_t)$.

        Hence, following the analysis of Lemma~C.2 in~\citet{lee2025improved} and applying Ville’s inequality, we obtain that
        \begin{align*}
            \PP
            \left(
                \sup_{t} A_t \geq \frac{1}{\delta}
            \right)
            &\leq 
            \EE [A_0] \delta
            \tag{Ville’s inequality}
            = \delta,
        \end{align*}        
        which is equivalent to
        \begin{align*}
            \PP
             \left(
             \exists t :
                 \sum_{s=1}^t \log M_s
                  \geq \log \frac{1}{\delta}
            \right)
            \leq \delta.
        \end{align*}
        Therefore, with probability at least $1-\delta$, we have
        \begin{align*}
            \sum_{s=1}^t 
            \sum_{j< k \leq |S_s|}
            \left(
                \ell_{s}^{(j,k)}(\thetab^\star)
                - 
                \bar{\ell}_{s}^{(j,k)}(\tilde{z}_s^{(j,k)}) 
            \right)
            &\leq 
            \frac{K(K-1)}{2} \sum_{s=1}^t 
            \frac{1}{\binom{|S_s|}{2}} \sum_{j < k \leq |S_s|} 
            \left(
                \ell_{s}^{(j,k)}(\thetab^\star)
                - 
                \bar{\ell}_{s}^{(j,k)}(\tilde{z}_s^{(j,k)}) 
            \right)
            \\
            &\leq \frac{K(K-1)}{2} \sum_{s=1}^t 
            \log M_s 
            \tag{Jensen's inquality}
            \\
            &\leq  \frac{K(K-1)}{2}  \log \frac{1}{\delta}.
        \end{align*}
        This concludes the proof of Lemma~\ref{lemma:online_regret_intermediate_(a)}.
    \end{proof}
    Now we follow the proof of Theorem 4.2 in~\citet{lee2025improved}, with substituting Lemma C.2 of theirs with our Lemma~\ref{lemma:online_regret_intermediate_(a)},
    then, we obtain that
    \begin{align*}
     \| \widehat{\thetab}_t^{(j,k)} -\thetab^\star \|_{H_t^{(j, k)}}^2 
     &\leq \frac{1 + 3\sqrt{2}B}{2} K(K-1)  \log \frac{1}{\delta}
     + 4 \sqrt{6} \left( \frac{1 + 3\sqrt{2}B}{2} \right)^2
     d \log (tK^2+2)
     + 4 B^2 \lambda
     \\
     &=  \BigO \left(
        B^2 
            d \log (tK/\delta)
            + B^2 \lambda
        \right).
        \tag{Let $K^2 \leq (2 \sqrt{6} + 12 \sqrt{3}B)d $}
    \end{align*}
    This completes the proof of Proposition~\ref{prop:online_CB_RB}.
\end{proof}

\subsubsection{Proof of Lemma~\ref{lemma:H_t_Lambda}}
\label{app_subsubsec:proof_of_lemma:H_t_Lambda}
\begin{proof}[Proof of Lemma~\ref{lemma:H_t_Lambda}]
    Recall that, under the Bradley–Terry-Luce (BTL) model defined in Equation~\eqref{eq:BT_model}, the probability that action $a$ is preferred over action $a'$ is given by:
    \begin{align*} 
        \PP\left( a \succ a' | x_t, ; \thetab \right) 
        &= \frac{\exp\left( \phi(x_t, a)^\top \thetab \right)} {\exp\left( \phi(x_t, a)^\top \thetab \right) + \exp\left( \phi(x_t, a')^\top \thetab \right)}
        = \mu \left(  \big( \phi(x_t, a) - \phi(x_t, a') \big)^\top \thetab \right)
        . 
    \end{align*}
    Then, we can derive a lower bound on the matrix $H_t$ as follows:
    \begin{align*}
        H_t
        &= \sum_{s =1}^{t-1} 
        \sum_{j=1}^{|S_s|-1} 
            \!\! \sum_{k=j+1}^{|S_s|} 
            \nabla^2 \ell_s^{(j,k)} (\widehat{\thetab}_{s}^{(j,k+1)}) + \lambda \Ib_d
        \\
        &\succeq
        \!\sum_{s \in [t-1] \setminus \Tcal^w}\!\!
        \sum_{j=1}^{|S_s|-1} 
            \!\! \sum_{k=j+1}^{|S_s|} 
            \nabla^2 \ell_s^{(j,k)} (\widehat{\thetab}_{s}^{(j,k+1)}) + \lambda \Ib_d
        \\
        &= \!\sum_{s \in [t-1] \setminus \Tcal^w}\!\!
        \sum_{j=1}^{|S_s|-1} \!\! \sum_{k=j+1}^{|S_s|}
        \dot{\mu} \left(
            z_{sjk}^\top \widehat{\thetab}_{s}^{(j,k+1)}
        \right)
        z_{sjk}
        z_{sjk}^\top
        +  \lambda \Ib_d
        \\
        &\succeq  \!\sum_{s \in [t-1] \setminus \Tcal^w}\!\!
        \sum_{j=1}^{|S_s|-1} \!\! \sum_{k=j+1}^{|S_s|}
        \dot{\mu} \left(
             z_{sjk}^\top  \widehat{\thetab}_s
        \right)
        e^{- \left| z_{sjk}^\top \big( \widehat{\thetab}_{s}^{(j,k+1)} - \widehat{\thetab}_s \big) \right|}
        z_{sjk}
        z_{sjk}^\top
        +  \lambda \Ib_d
        \tag{Lemma~\ref{lemma:dot_sigmoid_bound}}
        \\
        & \succeq e^{-2} \!\!\!\sum_{s \in [t-1] \setminus \Tcal^w}\!\!
        \sum_{j=1}^{|S_s|-1} \!\! \sum_{k=j+1}^{|S_s|}
        \dot{\mu} \left(
             z_{sjk}^\top  \widehat{\thetab}_s
        \right)
        z_{sjk}
        z_{sjk}^\top
        +  \lambda \Ib_d
        \\
        &\succeq \frac{ e^{-2}}{2} \Lambda_t
        ,
    \end{align*}
    where the second-to-last inequality holds because, for any $s \notin \WarmupRounds$, the following property is satisfied:
    \begin{align*}
        \left| z_{sjk}^\top  \big( \widehat{\thetab}_{s}^{(j,k+1)} - \widehat{\thetab}_s \big) \right|
        &=\left| \big( \phi(x_s, \sigma_{sj}) - \phi(x_s, \sigma_{sk}) \big)^\top 
         \left( \widehat{\thetab}_{s}^{(j,k+1)} - \widehat{\thetab}_s \right) \right|
        \\
        &\leq \big\| \phi(x_s, \sigma_{sj}) - \phi(x_s, \sigma_{sk}) \big\|_{H_s^{-1} }
        \left(
            \big\| \widehat{\thetab}_{s}^{(j,k+1)} - \thetab^\star \big\|_{H_s}
            + \big\| \widehat{\thetab}_s - \thetab^\star \big\|_{H_s}
        \right)
        \tag{Hölder's inequality}
        \\
        &\leq \frac{1}{\beta_{T+1}(\delta)}  
        \left(
            \big\| \widehat{\thetab}_{s}^{(j,k+1)} - \thetab^\star \big\|_{H^{(j,k+1)}_s}
            + \big\| \widehat{\thetab}_s - \thetab^\star \big\|_{H_s}
        \right)
        \tag{$s \neq \WarmupRounds, H_s \preceq H^{(j,k+1)}_s$}
        \\
        &\leq \frac{2\beta_{T+1}(\delta)}{\beta_{T+1}(\delta)} 
        \tag{Proposition~\ref{prop:online_CB_RB} and $\beta_t(\delta)$ is non-decreasing}
        \\
        &\leq 2.
    \end{align*}
    This concludes the proof of Lemma~\ref{lemma:H_t_Lambda}.
\end{proof}
\subsubsection{Proof of Lemma~\ref{lemma:bound_Tw}}
\label{app_subsubsec:proof_of_lemma:bound_Tw}
\begin{proof}[Proof of Lemma~\ref{lemma:bound_Tw}]
    The proof follows the same structure as the analysis of Lemma~\ref{lemma:bound_Tw_PL}, with the main difference that we use the Hessian corresponding to the RB loss.
    Let
    \begin{align*}
        (\tilde{a}_t, \bar{a}_t) =  \argmax_{a,a' \in \Acal} \dot\mu\left(\big(\phi(x_t, a) - \phi(x_t, a')\big)^\top \widehat{\thetab}_t  \right) \| \phi(x_t, a) - \phi(x_t, a') \|_{H_t^{-1} }^2.
    \end{align*}
    By the assortment selection rule in~\eqref{eq:S_t_selection_greedy}, the two actions 
    $\tilde{a}_t, \bar{a}_t$ 
    are always included in \(S_t\). 
    It is because, for any pair \(\{a,a'\}\), we have
    \begin{align*}
        \dot\mu\left(\big(\phi(x_t, a) - \phi(x_t, a')\big)^\top \widehat{\thetab}_t  \right)
        \| \phi(x_t, a) - \phi(x_t, a') \|^2_{H_t^{-1} }
        = 2 f_t(\{a, a'\} )
        ,
    \end{align*}
    which directly implies that $\tilde{a}_t, \ReferenceAction_t \in S_t$.

    For simplicity, we denote $z_{s k k'} = \phi(x_s, \sigma_{sk}) - \phi(x_s, \sigma_{sk'})$.
    Recall that by the definition of $H_t $,
    \begin{align*}
        H_t
        &= \sum_{s =1}^{t-1} 
        \sum_{j=1}^{|S_s|-1} 
            \!\! \sum_{k=j+1}^{|S_s|} 
            \nabla^2 \ell_s^{(j,k)} (\widehat{\thetab}_{s}^{(j,k+1)}) + \lambda \Ib_d
        \\
        &\succeq 
        \frac{e^{-4B}}{4}
        \sum_{s=1}^{t-1}
        \sum_{j=1}^{|S_s|-1} \!\! \sum_{k=j+1}^{|S_s|}
        z_{sjk}
        z_{sjk}^\top
        +  \lambda \Ib_d
        \\
        &\succeq 
        \frac{e^{-4B}}{4}
        \sum_{s=1}^{t-1}
        \big(
            \phi(x_s, \tilde{a}_s) - \phi(x_s, \ReferenceAction_s)
        \big)
        \big(
            \phi(x_s, \tilde{a}_s) - \phi(x_s, \ReferenceAction_s)
        \big)^\top
        +  \lambda \Ib_d
        \tag{ $\tilde{a}_s, \bar{a}_s\in S_s$}
        \\
        &\succeq 
        \frac{e^{-4B}}{4}
        \Bigg(\sum_{s \in \WarmupRounds_t}
        \dot\mu\!\left(\big(\phi(x_s, \tilde{a}_s) - \phi(x_s, \ReferenceAction_s)\big)^\top \widehat{\thetab}_s  \right)
        \big(
            \phi(x_s, \tilde{a}_s) - \phi(x_s, \ReferenceAction_s)
        \big)
        \big(
            \phi(x_s, \tilde{a}_s) - \phi(x_s, \ReferenceAction_s)
        \big)^\top
        \\
        &\quad\quad\quad\quad+  \lambda \Ib_d \Bigg)
        \numberthis \label{eq:H_t_lower_Lambda_w_RB}
        ,
    \end{align*}
    where in the last inequality, we use the fact that $\dot\mu (\cdot) \leq \frac{1}{4}$ and let $ \WarmupRounds_t = \WarmupRounds \setminus \{t, \dots, T\}$.
    Define
    \begin{align*}
        \Lambda^{w}_t
        :=& \sum_{s \in \WarmupRounds_t}
        \dot\mu\!\left(\big(\phi(x_s, \tilde{a}_s) - \phi(x_s, \ReferenceAction_s)\big)^\top \widehat{\thetab}_s  \right)
        \big(
            \phi(x_s, \tilde{a}_s) - \phi(x_s, \ReferenceAction_s)
        \big)
        \big(
            \phi(x_s, \tilde{a}_s) - \phi(x_s, \ReferenceAction_s)
        \big)^\top
        \\
        &+ \lambda \Ib_d.
    \end{align*}
    Furthermore, by setting $\lambda \geq 1$, for any $t$, the following holds  
    \begin{align*}
         \dot\mu\left(\big(\phi(x_t, \tilde{a}_t) - \phi(x_t, \ReferenceAction_t)\big)^\top \widehat{\thetab}_t  \right) \| \phi(x_t, \tilde{a}_t) - \phi(x_t, \ReferenceAction_t) \|^2_{\big(\Lambda^w_t\big)^{-1}}
        &\leq \frac{1}{4} \cdot \frac{2}{\lambda}
        \tag{$\dot{\mu}(\cdot) \leq \frac{1}{4}$}
        \\
        &\leq 1 
        .
        \numberthis \label{eq:ep_max_RB}
    \end{align*}
    Then, we obtain
    \begin{align*}
        \sum_{t \in \WarmupRounds} &\max_{a, a' \in \Acal} \| \phi(x_t, a) - \phi(x_t, a') \|^2_{H_t^{-1} }
        \\
        &= \sum_{t \in \WarmupRounds} \max_{a, a' \in \Acal} \frac{\dot\mu\left(\big(\phi(x_t, a) - \phi(x_t, a')\big)^\top \widehat{\thetab}_t \right)}{\dot\mu\left(\big(\phi(x_t, a) - \phi(x_t, a')\big)^\top \widehat{\thetab}_t \right)}
        \| \phi(x_t, a) - \phi(x_t, a') \|^2_{H_t^{-1} }
        \\
        &\leq 4 e^{2B} 
        \sum_{t \in \WarmupRounds} \max_{a, a' \in \Acal} 
        \dot\mu\left(\big(\phi(x_t, a) - \phi(x_t, a')\big)^\top \widehat{\thetab}_t  \right)
        \| \phi(x_t, a) - \phi(x_t, a') \|^2_{H_t^{-1} }
        \tag{Assumption~\ref{assum:linear_reward}}
        \\
        &=
           4 e^{2B} 
            \sum_{t \in \WarmupRounds}
            \dot\mu\left(\big(\phi(x_t, \tilde{a}_t) - \phi(x_t, \ReferenceAction_t)\big)^\top \widehat{\thetab}_t  \right)
            \| \phi(x_t, \tilde{a}_t) - \phi(x_t, \ReferenceAction_t) \|^2_{H_t^{-1} }
        \\
        &\leq 16 e^{6B}
         \sum_{t \in \WarmupRounds}
         \dot\mu\left(\big(\phi(x_t, \tilde{a}_t) - \phi(x_t, \ReferenceAction_t)\big)^\top \widehat{\thetab}_t  \right)
         \| \phi(x_t, \tilde{a}_t ) - \phi(x_t, \ReferenceAction_t) \|^2_{\big(\Lambda^w_t\big)^{-1}}
        \tag{Eqn.~\eqref{eq:H_t_lower_Lambda_w_RB}}
        \\
        &= 
         16 e^{6B}
         \sum_{t \in \WarmupRounds}
         \min \left\{ 1,  \dot\mu\left(\big(\phi(x_t, \tilde{a}_t) - \phi(x_t, \ReferenceAction_t)\big)^\top \widehat{\thetab}_t  \right) \| \phi(x_t, \tilde{a}_t ) - \phi(x_t, \ReferenceAction_t) \|^2_{\big(\Lambda^w_t\big)^{-1}} \right\}
         \tag{Eqn.~\eqref{eq:ep_max_RB} }
         \\
         &\leq 32 e^{6B}
         d \log \left(
            1 + \frac{2 T}{d \lambda}
         \right).
         \tag{Lemma~\ref{lemma:EPL}}
    \end{align*}
    On the other hand, for $t \in \WarmupRounds $, we know that
    \begin{align*}
         \sum_{t \in \WarmupRounds} \max_{a, a' \in \Acal} \| \phi(x_t, a) - \phi(x_t, a') \|^2_{H_t^{-1} }
         \geq  \frac{|\WarmupRounds|}{ \beta_{T+1}(\delta)^2}.
    \end{align*}
    By combining the two results above, and setting $\kappa = e^{-6}$, we derive that 
    \begin{align*}
        | \WarmupRounds |
        \leq \frac{32}{\kappa} \beta_{T+1}(\delta)^2
         d \log \left(
            1 + \frac{2 T}{d \lambda}
         \right),
    \end{align*}
    which concludes the proof.
\end{proof}

\section{Technical Lemmas}
\label{app_sec:technical_lemmas_for_PL}
%
%
\begin{lemma} [Proposition B.5 of~\citealt{lee2025improved}]
\label{lemma:self_MNL_hessian_norm}
The Hessian of the multinomial logistic loss $\bar{\ell}: \RR^{M} \rightarrow \RR$ satisfies that, for any $\ab_1, \ab_2 \in \RR^{M}$, we have:
\begin{align*}
    e^{-3\sqrt{2} \| \ab_1 - \ab_2 \|_{\infty} }  \nabla^2 \bar{\ell}  (\ab_1)
    \preceq \nabla^2  \bar{\ell} (\ab_2)
    \preceq e^{3\sqrt{2} \| \ab_1 - \ab_2 \|_{\infty} } \nabla^2 \bar{\ell} (\ab_1).
\end{align*}
\end{lemma}
\begin{lemma} [Lemma 9 of~\citealt{abeille2021instance}] \label{lemma:dot_sigmoid_bound}
    Let $f$ be a strictly increasing function such that $|\ddot{f}| \leq \dot{f}$, and let $\Zcal$ be any bounded interval of $\RR$.
    Then, for all $z_1, z_2 \in \Zcal$, we have
    \begin{align*}
        \dot{f}(z_2) \exp \left( - |z_2 - z_1| \right)
        \leq \dot{f}(z_1)
        \leq \dot{f}(z_2) \exp \left( |z_2 - z_1| \right).
    \end{align*}
\end{lemma}
We also provide a concentration lemma for positive semi-definite (PSD) random matrices.
\begin{lemma} [Concentration of PSD matrices]
\label{lemma:concentration_PSD}
    Let $\mu_i$ denote the conditional distribution of a positive semi-definite $M \in \RR^{d \times d}$ conditioned on the filtration 
    $\Fcal_{i-1}$.
    Assume $\lambda_{\max} (M) \leq 1$.
    Define $\widebar{M} := \frac{1}{n} \sum_{i=1}^n \EE_{M \sim \mu_i} M$. 
    If $\lambda = \Omega(d \log (n/\delta))$, then with probability at least $1-\delta$, for any $n \geq 1$, 
    \begin{align*}
        \frac{1}{3} \left( n \widebar{M} + \lambda \Ib_d \right)
        \preceq  \sum_{i=1}^n M_i +  \lambda \Ib_d 
        \preceq \frac{5}{3}  \left( n \widebar{M} + \lambda \Ib_d \right).
    \end{align*}
\end{lemma}
\begin{proof} [Proof of Lemma~\ref{lemma:concentration_PSD}]
    The overall structure of the proof closely follows that of Lemma 39 in~\citet{zanette2021cautiously}. 
    For completeness, we provide the full proof below.

    Fix $x \in \RR^d$ such that $\|x\|_2 = 1$. 
    Let $\widebar{M}_i = \EE_{M \sim \mu_i} M$ and $\widebar{M} = \frac{1}{n} \sum_{i=1}^n \widebar{M}_i$.
    Then, we have
    \begin{align*}
        \EE_{M \sim \mu_i} x^\top M x 
        = x^\top \EE_{M \sim \mu_i} M x
        = x^\top \widebar{M}_i x.
    \end{align*}
    Since $M$ is a positive semi-definite matrix, the random variable $x^\top M x$ is non-negative, and it satisfies
    $x^\top M x \leq \lambda_{\max} (M) \|x\|_2^2 \leq 1$. 
    Thus, the conditional variance is at most $x^\top \widebar{M}_i x$ because
    \begin{align*}
        \operatorname{Var}_{M \sim \mu_i} (x^\top M x)
        \leq \EE_{M \sim \mu_i}(x^\top M x)^2
        \leq \EE_{M \sim \mu_i} x^\top M x
        = x^\top \widebar{M}_i x.
    \end{align*}
    Applying Lemma~\ref{lemma:Berstein_for_Martingales} with the filtration $\Fcal_i$, we obtain that, with probability at least $1-\delta$, there exists a universal constant $c$ such that
    \begin{align*}
        \left|
            \frac{1}{n} 
            \sum_{i=1}^n
            \left(
                x^\top M_i x
                - x^\top \widebar{M}_i x
            \right)
        \right|
        =
        \left|
            \frac{1}{n} 
            \sum_{i=1}^n
                x^\top M_i x
            - x^\top \widebar{M} x
        \right|
        \leq 
        c \left(
            \sqrt{\frac{2 x^\top \widebar{M} x}{n}\log (2/\delta)}
            + \frac{\log (2/\delta)}{3n}
        \right).
    \end{align*}        
    Now, we will show that if $\lambda = \Omega(\log (1/\delta))$, we can derive
    \begin{align*}
        c \left(
            \sqrt{ \frac{2 x^\top \widebar{M} x}{n}\log (2/\delta)}
            + \frac{\log (2/\delta)}{3n}
        \right)
        \leq \frac{1}{2}
        \left( x^\top  \widebar{M} x + \frac{\lambda}{n} \right).
        \numberthis \label{eq:concentration_PSD_condition}
    \end{align*}
    \textbf{Case 1.} $x^\top \widebar{M} x \leq \frac{\lambda}{n}$.
    \\
    In this case, it is sufficient to satisfy for some constants  $c', c''$
    \begin{align*}
        \sqrt{\frac{2 \log (2/\delta)}{n}} \leq c' \sqrt{\frac{\lambda}{n}}
        \quad &\longleftrightarrow \quad
        \Omega (\log (1/ \delta)) \leq \lambda
        \\
        \frac{\log (2/\delta)}{3n} \leq c'' \left(\frac{\lambda}{n}\right)
        \quad &\longleftrightarrow \quad 
        \Omega (\log (1/ \delta)) \leq \lambda.
    \end{align*}

    \textbf{Case 2.} $x^\top \widebar{M} x > \frac{\lambda}{n}$.
    \\
    In this case, it is sufficient to satisfy for some constants  $c''', c'''$
    \begin{align*}
        \sqrt{\frac{ 2 x^\top \widebar{M} x}{n}\log (2/\delta)}
        \leq c''' \left(\frac{\lambda}{n}\right)
        \quad &\longleftrightarrow \quad
        \Omega (\log (1/ \delta)) \leq \lambda
        \\
        \frac{\log (2/\delta)}{3n} \leq c'''' \left(\frac{\lambda}{n}\right)
        \quad &\longleftrightarrow \quad 
        \Omega (\log (1/ \delta)) \leq \lambda. 
    \end{align*}
    Therefore, Equation~\eqref{eq:concentration_PSD_condition} is satisfied.
    Since $\|x\|_2 \leq 1$, this implies
    \begin{align*}
        \left|
        x^\top
        \left(
            \frac{1}{n} 
            \sum_{i=1}^n
                 M_i 
                -  \widebar{M} 
        \right)
        x
        \right|
        \leq 
        \frac{1}{2}
        x^\top \left(
            \widebar{M} + \frac{\lambda}{n} \Ib_d
        \right)
        x.
        \numberthis \label{eq:concentration_PSD_condition_2}
    \end{align*}
    We denote the boundary of the unit ball by $\partial \mathcal{B} = \left\{ \|x\|_2 = 1 \right\}$.
    Then, for any $x \in \partial \mathcal{B}$, we know there exists a $x'$ in the $\epsilon$-covering such that $\| x - x' \|_2 \leq \epsilon$. 
    Let $\mathcal{N}_{\epsilon}$ be the $\epsilon$-covering number of $\partial \mathcal{B}$.
    Then, by the covering number of Euclidean ball lemma (Lemma~\ref{lemma:covering_number}), we get 
    \begin{align*}
        \mathcal{N}_{\epsilon} \leq \left(\frac{3}{\epsilon} \right)^d.
        \numberthis \label{eq:covering_number}
    \end{align*}
    Taking a union bound over $x'$ and the number of samples $n$, with probability at least $1- n \mathcal{N}_{\epsilon} \delta$, we obtain
    \begin{align*}
        \left|
        x^\top
        \left(
            \frac{1}{n} 
            \sum_{i=1}^n
                 M_i 
                -  \widebar{M} 
        \right)
        x
        \right|
        &\leq 
        \left|
        (x')^\top
        \left(
            \frac{1}{n} 
            \sum_{i=1}^n
                 M_i 
                -  \widebar{M} 
        \right)
        x'
        \right|
        + 
        \left|
        ( x- x' )^\top
        \left(
            \frac{1}{n} 
            \sum_{i=1}^n
                 M_i 
                -  \widebar{M} 
        \right)
        x'
        \right|
        \\
        &+ \left|
        (x')^\top
        \left(
            \frac{1}{n} 
            \sum_{i=1}^n
                 M_i 
                -  \widebar{M} 
        \right)
        (x -x')
        \right|
        \\
        &\leq 
        \left|
        (x')^\top
        \left(
            \frac{1}{n} 
            \sum_{i=1}^n
                 M_i 
                -  \widebar{M} 
        \right)
        x'
        \right|
        + 4 \epsilon.
        \tag{$\| x - x'\|_2 \leq \epsilon$ and $M_i, \| \widebar{M} \|_2 \leq 1$ }
        \\
        &\leq 
        \frac{1}{2}
         (x')^\top  \left( \widebar{M} + \frac{\lambda}{n} \Ib_d \right) x'  
        + 4 \epsilon
        \tag{Eqn.~\eqref{eq:concentration_PSD_condition_2}}
        \\
        &\leq \frac{1}{2}
        x^\top \left( \widebar{M} + \frac{\lambda}{n} \Ib_d \right) x 
        + \frac{9}{2} \epsilon
        \tag{$\| x - x'\|_2 \leq \epsilon$ and $\| \widebar{M} \|_2 \leq 1$ }
        \\
        &\leq 
        \frac{2}{3}
         x^\top \left( \widebar{M} + \frac{\lambda}{n} \Ib_d \right) x 
        \tag{set $\epsilon = \BigO(\frac{1}{n})$},
    \end{align*}
    where $\lambda = \Omega \left( \log \left( \frac{2 n \mathcal{N}_{\epsilon}}{\delta} \right) \right)$.
    By substituting $\delta \leftarrow \delta / (n \mathcal{N}_{\epsilon} + 1)$ and combining this with Equation~\eqref{eq:covering_number}, we obtain:
    \begin{align*}
        \frac{1}{3}\left( \widebar{M} + \frac{\lambda}{n} \Ib_d \right) 
        \preceq 
        \frac{1}{n} \sum_{i=1}^n M_i + \frac{\lambda}{n} \Ib_d 
        \preceq
        \frac{5}{3} \left( \widebar{M} + \frac{\lambda}{n} \Ib_d \right),
    \end{align*}
    which concludes the proof.
\end{proof}

\begin{lemma} [Bernstein for martingales, Theorem 1 of~\citealt{beygelzimer2011contextual} and Lemma 45 of~\citealt{zanette2021cautiously}]
\label{lemma:Berstein_for_Martingales}
    Consider the stochastic process $\{ X_n\}$ adapted to the filtration $\{ \mathcal{F}_n \}$.
    Assume $\EE X_n = 0$ and $c X_n \leq 1$ for every $n$; then for every constant $z \neq 0$ it holds that
    \begin{align*}
        \operatorname{Pr} \left(
            \sum_{n=1}^N X_n \leq
            z \sum_{n=1}^N \EE (X_n^2 \mid \mathcal{F}_n)
            + \frac{1}{z} \log \frac{1}{\delta}
        \right)
        \geq 1-\delta.
    \end{align*}
    By optimizing the bound as a function of $z$, we also have
    \begin{align*}
        \operatorname{Pr} \left(
            \sum_{n=1}^N X_n \leq
            c 
            \sqrt{
                \sum_{n=1}^N \EE (X_n^2 \mid \mathcal{F}_n)
                \log \frac{1}{\delta}
            }
            +  \log \frac{1}{\delta}
        \right)
        \geq 1-\delta.
    \end{align*}
\end{lemma}
\begin{lemma} [Covering number of Euclidean ball] \label{lemma:covering_number}
    For any $\epsilon > 0$, the $\epsilon$-covering number of the Euclidean ball in $\RR^d$ with radius $R >0$ is upper bounded by $(1 + 2R/\epsilon)^2$.
\end{lemma}
\begin{lemma} [Elliptical potential lemma~\citealt{abbasi2011improved}]
\label{lemma:EPL}
    Let $\{ z_{t} \}_{t \geq 1}$ be a bounded sequence in $\RR^d$ satisfying $\max_{t \geq 1} \| z_{t} \|_2 \leq X$.
    For any $t \geq 1$, we define $\Lambda_t := \sum_{s=1}^{t-1} z_{s} z_{s}^\top + \lambda \Ib_d $ with $\lambda > 0$.
    Then, we have 
    \begin{align*}
        \sum_{t=1}^T
        \min \left\{
            1, \| z_{t} \|_{\Lambda_t^{-1}}^2
        \right\}
        \leq 2 d \log \left(
            1 + \frac{X^2 T}{d \lambda}
        \right).
    \end{align*}        
\end{lemma}
%
\section{Proof of Theorem~\ref{thm:lower_bound}}
\label{app_sec:proof_lower}
\subsection{Main Proof of Theorem~\ref{thm:lower_bound}} \label{app_subsec:main_proof_lower_bound}
Throughout the proof, we consider the setting where the context space is a singleton, i.e., $\Xcal = \{x\}$.
As a result, the problem reduces to a context-free setting, and we focus solely on the action space $\Acal$.
Note that this is equivalent to assuming that $\rho$ is a Dirac distribution.

We first present the following theorem, which serves as the foundation for our analysis.
\begin{theorem} [Lower bound on adaptive PL model parameter estimation]
\label{thm:lower_bound_adap_PL_PLE}
    Let $\Phi = \Scal^{d-1}$ be the unit sphere in $\mathbb{R}^d$, and let $\Theta = \{ -\mu, \mu \}^d$ for some $\mu \in (0, 1/ \sqrt{d} ]$.
    We consider a query model where, at each round $t = 1, \dots, T$, the learner selects a subset $S_t \subseteq \Phi$ of feature vectors, with cardinality satisfying $2 \leq |S_t| \leq K$, and then receives a ranking feedback  $\sigma_t$ drawn from the Plackett–Luce (PL) model defined as: 
    \begin{align*}
        \PP( \sigma_t | S_t ; \thetab)
        =
        \prod_{j=1}^{|S_t|}
        \frac{\exp \left( \phi_{\sigma_{tj}}^\top \thetab \right)}
        {
        \sum_{k=j}^{|S_t|} \exp \left( \phi_{\sigma_{tk}}^\top \thetab \right)
        }
        ,
    \end{align*}
    where $\sigma_t = ( \sigma_{t1}, \dots, \sigma_{t|S_t|})$ is a permutation of the actions in $S_t$,  $\phi_{a} \in \Phi$ denotes the feature vector associated with action $a \in \Acal$ in the selected subset at round $t$,
    and $\thetab \in \Theta$.
    Then, we have
    \begin{align*}
        \inf_{\widehat{\thetab}, \pi}
        \max_{\thetab \in \Theta}
        \EE_{\thetab }
        \left[
            \big\|
                \thetab - \widehat{\thetab}
            \big\|^2_2
        \right]
        \geq \frac{d \mu^2}{2} 
        \left(
            1 - 
            \sqrt{\frac{2 K^2 T \mu^2  }{d}
            }
        \right),
    \end{align*}
    where the infimum is over all measurable estimators $\widehat{\thetab}$ and measurable (but possibly adaptive) query rules $\pi$, 
    and $\EE_{\thetab} [\cdot]$ denotes the expectation over the randomness in the observations and decision rules if $\thetab$ is the true instance.
    In particular, if $T \geq \frac{d^2}{8 K^2 }$, by choosing $\mu = \sqrt{d / (8 K^2 T )}$, we obtain
    \begin{align*}
        \inf_{\widehat{\thetab}, \pi}
        \max_{\thetab \in \Theta}
        \EE_{\thetab }
        \left[
            \big\|
                \thetab - \widehat{\thetab}
            \big\|^2_2
        \right]
        \geq \frac{d^2}{32 K^2 T}.
    \end{align*}
\end{theorem}
\begin{proof} [Proof of Theorem~\ref{thm:lower_bound_adap_PL_PLE}]
    The analysis of this result closely follows the proof of Theorem 3 in~\citet{shamir2013complexity}.
    The key distinction lies in the input structure: our setting involves a set of feature vectors, while theirs is restricted to a single feature vector.

    To begin with, since the worst-case expected regret with respect to $\thetab$ can be lower bounded by the average regret under the uniform prior over $\Theta$, we have:
    \begin{align*}
        \max_{\thetab \in \Theta}
        \EE_{\thetab }
        \left[
            \big\|
                \thetab - \widehat{\thetab}
            \big\|^2_2
        \right]
        &\geq \EE_{\thetab \sim \operatorname{Unif}(\Theta)}  \EE_{\thetab }
        \left[
            \big\|
                \thetab - \widehat{\thetab}
            \big\|^2_2
        \right]
        \\
        &= \EE_{\thetab \sim \operatorname{Unif}(\Theta)}  \EE_{\thetab }
        \left[
            \sum_{i=1}^d
                \left( \thetab - \widehat{\thetab} \right)^2
        \right]
        \\
        &\geq \mu^2 \cdot
        \EE_{\thetab \sim \operatorname{Unif}(\Theta)}
            \EE_{\thetab}
            \left[
                \sum_{i=1}^d 
                \II\left\{ \thetab_i \widehat{\thetab}_i < 0 \right\}
            \right]
        \numberthis \label{eq:proof_thm:lower_bound_adap_PL_PLE_1}
        .
    \end{align*}
    As in~\citet{shamir2013complexity}, we assume that the query strategy is deterministic conditioned on the past: that is, 
    $S_t$ is a deterministic function of the previous queries and observations, i.e., $S_1, \sigma_1, \dots, S_{t-1}, \sigma_{t-1}$.
    This assumption is made without loss of generality, since any randomized querying strategy can be viewed as a distribution over deterministic strategies.
    Therefore, a lower bound that holds uniformly for all deterministic strategies also applies to any randomized strategy.
    Then, we use the following lemma.
    \begin{lemma} [Lemma 4 of~\citealt{shamir2013complexity}] 
        \label{lemma:lemma4_shamir2013}
        Let $\thetab$ be a random vector, none of whose coordinates is supported on $0$, and let $y_1, y_2, \dots, y_T$ be a sequence of queries obtained by a deterministic strategy returning a point $\widehat{\thetab}$ (that is, $\bm{\psi}_t$ is a deterministic function of $\bm{\psi}_1,   y_1, \dots, \bm{\psi}_{t-1},  y_{t-1} $, and $\widehat{\thetab}$ is a deterministic function of $y_1, \dots, y_{T}$). Then, we have
        \begin{align*}
            \EE_{\thetab \sim \operatorname{Unif}(\Theta)}
            \EE_{\thetab}
            \left[
                \sum_{i=1}^d 
                \II\left\{ \thetab_i \widehat{\thetab}_i < 0 \right\}
            \right]
            \geq \frac{d}{2}
            \left(
                1 - 
                \sqrt{\frac{1}{d}
                    \sum_{i=1}^d \sum_{t=1}^T
                    U_{ti}
                }
            \right),
        \end{align*}
        where 
        \begin{align*}
            U_{ti} := 
            \sup_{\thetab_j, j \neq i}
            D_{\text{KL}}\!
            \left(
                P \!\left(y_t | \thetab_i > 0, 
                    \{ \thetab_j \}_{j \neq i},
                    \{ y_s \}_{s=1}^{t-1}
                \right)
                 \|\, 
                 P \!\left(y_t | \thetab_i < 0, 
                    \{ \thetab_j \}_{j \neq i},
                    \{ y_s \}_{s=1}^{t-1}
                \right)
            \right).
        \end{align*}
    \end{lemma}
    In our setting, we interpret $y_{t} = \sigma_t$, and $\bm{\psi}_t = \{ \phi_{a} \}_{a \in S_t} \subseteq \Phi$.
    Then, we can write $U_{ti}$ as follows:
    \begin{align*}
        U_{ti} = 
        \sup_{\thetab_j, j \neq i}
            D_{\text{KL}}\!
            \left(
                \PP\left(\sigma_t | S_t
                    ;\thetab_i > 0, 
                    \{ \thetab_j \}_{j \neq i},
                \right)
                 \|\, 
                 \PP\left(\sigma_t | S_t
                    ;\thetab_i < 0, 
                    \{ \thetab_j \}_{j \neq i},
                \right)
            \right).
    \end{align*}
    For simplicity, let $\PP_{\thetab}(\sigma | S) = \PP\left(\sigma | S
                    ;\thetab
                \right)$.
    Then, we can upper bound $U_{ti}$ using the following lemma.
    \begin{lemma} \label{lemma:KL_PL_upperbound}
        For any $\thetab, \thetab' \in \mathbb{R}^d$,
        let $\mathbb{P}_{\thetab}(\cdot \mid S)$ denote the PL distribution over rankings induced by the action set $S$ and parameter vector $\thetab$.
        Then, we have
        \begin{align*}
            D_{\text{KL}}
            \big(\PP_{\thetab}  (\cdot | S) \|\PP_{\thetab'}  (\cdot | S)   \big)
            \leq \frac{K}{2} 
            \sum_{a \in S }\left( \phi_{a}^\top
                ( \thetab' - \thetab ) \right)^2.
        \end{align*}
    \end{lemma}
    The proof is deferred to Appendix~\ref{subsubsec:proof_lemma:KL_PL_upperbound}.

    By applying Lemma~\ref{lemma:KL_PL_upperbound}, we have
    \begin{align*}
       \sum_{i=1}^d U_{ti} 
            &\leq  \frac{K}{2}
            \sum_{i=1}^d \sum_{a \in S_t }
            \left( 2 \mu \cdot [\phi_{a}]_i \right )^2
            = 2 K\mu^2
            \sum_{a \in S_t }
            \underbrace{\sum_{i=1}^d
            ([\phi_{a}]_i)^2}_{=1}
            \\
            &= 2  K \mu^2 \cdot |S_t|
            \tag{$\phi_{a} \in \Scal^{d-1}$}
            \\
            &\leq  2 K^2 \mu^2.
            \tag{$|S_t| \leq K$}
    \end{align*}
    Hence, by Lemma~\ref{lemma:lemma4_shamir2013}, we get
    \begin{align*}
        \EE_{\thetab \sim \operatorname{Unif}(\Theta)}
            \EE_{\thetab}
            \left[
                \sum_{i=1}^d 
                \II\left\{ \thetab_i \widehat{\thetab}_i < 0 \right\}
            \right]
        &\geq \frac{d}{2}
        \left(
            1 - 
            \sqrt{\frac{1}{d}
                \sum_{i=1}^d \sum_{t=1}^T
                U_{ti}
            }
        \right)
        \\
        &\geq \frac{d}{2}
        \left(
            1 - 
            \sqrt{\frac{2 K^2 T \mu^2  }{d}
            }
        \right).
        \numberthis \label{eq:proof_thm:lower_bound_adap_PL_PLE_2}
    \end{align*}
    Combining Equation~\eqref{eq:proof_thm:lower_bound_adap_PL_PLE_1} and~\eqref{eq:proof_thm:lower_bound_adap_PL_PLE_2},
    we prove the first inequality of Theorem~\ref{thm:lower_bound_adap_PL_PLE}.
    The second inequality directly follows by choosing $\mu = \sqrt{d / (8 K^2 T )}$.
\end{proof}
We are now ready to present the proof of Theorem~\ref{thm:lower_bound}.
\begin{proof} [Proof of Theorem~\ref{thm:lower_bound}]
    The structure of our proof is similar to that of Theorem 2 in~\citet{wagenmaker2022reward}.
    However, while they consider the linear bandit setting, we focus on the Plackett–Luce (PL) bandit setting.
    
    We adopt the same instance construction as in Theorem~\ref{thm:lower_bound_adap_PL_PLE}, where $\Phi = \Scal^{d-1}$ and $\Theta = \{ -\mu, \mu \}^d$.
    Define $\phi^\star(\thetab) = \argmax_{a \in \Acal} \phi_a^\top \thetab $.
    Then, since $\phi^\star(\thetab) \in \Phi$ and $\thetab \in \Phi$, it is clear that 
    \begin{align*}
        \phi^\star(\thetab) = \thetab / \| \thetab \|_2 = \thetab / (\sqrt{d} \mu),
        \quad \text{and }\,
        \phi^\star(\thetab)^\top \thetab = \sqrt{d} \mu.
        \numberthis \label{eq:proof_lower_phi_star}
    \end{align*}
    Fix the suboptimality gap $\epsilon >0$.
    By definition, a policy $\pi \in \triangle_{\Phi}$ is said to be $\epsilon$-optimal if it satisfies
    \begin{align*}
        \EE_{\phi \sim \pi} \left[ \phi^\top \thetab \right]
        = \big(\underbrace{\EE_{\phi \sim \pi} \left[ \phi \right]}_{=: \phi_\pi} \big)^\top \thetab
        \geq \phi^\star(\thetab)^\top \thetab  - \epsilon
        = 
        \sqrt{d} \mu - \epsilon.
        \numberthis \label{eq:proof_lower_subopt_gap}
    \end{align*}
    Moreover, by Jensen’s inequality, we have
    \begin{align*}
        \| \phi_{\pi} \|_2^2 \leq 
        \EE_{\phi \sim \pi} \left[ \| \phi \|_2^2 \right] = 1.
    \end{align*}
    Let $\Delta = \phi_{\pi} - \phi^\star(\thetab)$.
    Then, we get
    \begin{align*}
        &1 \geq \| \phi_{\pi} \|_2^2
        = \| \phi^\star(\thetab) + \Delta  \|_2^2
        = 1 + \|\Delta  \|_2^2
        + 2 \phi^\star(\thetab)^\top \Delta 
        \\
        \Longleftrightarrow \,\,
        &\phi^\star(\thetab)^\top \Delta  
        \leq - \frac{1}{2}\|\Delta  \|_2^2
        \\
        \Longleftrightarrow \,\,
        & \thetab^\star \Delta  
        \leq  - \frac{\sqrt{d} \mu}{2}\|\Delta  \|_2^2.
        \tag{Eqn.~\eqref{eq:proof_lower_phi_star}}
    \end{align*}
    Hence, if a policy $\pi$ is $\epsilon$-optimal for a parameter $\thetab$, then the following bound holds:
    \begin{align*}
        &-\epsilon \leq - \frac{\sqrt{d} \mu}{2}\|\Delta  \|_2^2.
        \tag{Eqn.~\eqref{eq:proof_lower_subopt_gap}}
        \\
        \Longleftrightarrow \,\,
        & \|\Delta  \|_2^2 \leq \frac{2 \epsilon}{\sqrt{d} \mu},
        \quad
        \text{where }\, \thetab = \sqrt{d} \mu (\phi_\pi - \Delta ).
    \end{align*}
    We now assume that we are given an $\epsilon$-optimal policy $\widehat{\pi}$.
    Define $\widehat{\phi}:= \phi_{\widehat{\pi}}$ and the following estimator
    \begin{align*}
        \widehat{\thetab}
        = \begin{cases}
            \thetab' 
            &\quad \text{if } \exists \thetab' \in \Theta \text{ with }
            \thetab' = \sqrt{d} \mu (\widehat{\phi} - \Delta')
            \text{ for some } 
            \Delta' \in \RR^d, \|\Delta' \|_2^2 \leq \frac{2 \epsilon}{\sqrt{d} \mu};
            \\
            \text{any }  \thetab' \in \Theta 
            &\quad \text{otherwise}.
        \end{cases}
    \end{align*}
    If $\widehat{\pi}$ is indeed $\epsilon$-optimal for some $\thetab \in \Theta$, then the first condition is satisfied, and we have:
    \begin{align*}
        \big\| \widehat{\thetab} - \thetab \big\|_2
        =  \big\| 
            \sqrt{d} \mu (\widehat{\phi} - \Delta') 
            - \sqrt{d} \mu (\widehat{\phi} - \Delta)
        \big\|_2
        \leq 2 \sqrt{d} \mu \sqrt{\frac{2 \epsilon}{\sqrt{d} \mu}}
        = \sqrt{8 \sqrt{d} \mu \epsilon}.
        \numberthis \label{eq:proof_lower_epsilon_optiaml_condition}
    \end{align*}
    We denote $\Ecal$ as the event that $\widehat{\pi}$ is $\epsilon$-optimal for $\thetab \in \Theta$.
    Then, we get
    \begin{align*}
        \EE_{\thetab} \left[ \big\| \widehat{\thetab} - \thetab \big\|_2^2 \right]
        &= \EE_{\thetab} \left[ 
        \big\| \widehat{\thetab} - \thetab \big\|_2^2 \cdot \II\{ \Ecal \}
        + \big\| \widehat{\thetab} - \thetab \big\|_2^2 \cdot \II\{ \Ecal^c \}
        \right]
        \\
        &\leq 
        8 \sqrt{d} \mu \epsilon
        +
        \EE_{\thetab} \left[ 
        \big\| \widehat{\thetab} - \thetab \big\|_2^2 \cdot \II\{ \Ecal^c \}
        \right]
        \tag{Eqn.~\eqref{eq:proof_lower_epsilon_optiaml_condition}}
        \\
        &\leq 8 \sqrt{d} \mu \epsilon
        + 2 d \mu^2 \cdot
        P_{\thetab} [\Ecal^c].
        \tag{$\max \{\| \widehat{\thetab} \|_2^2, \| \thetab \|_2^2 \} \leq d \mu^2$}
    \end{align*}
    On the other hand, by Theorem~\ref{thm:lower_bound_adap_PL_PLE}, there exists a parameter $\thetab \in \Theta$ such that, if we collect $T$ samples and set $\mu = \sqrt{d / (8 K^2 T)}$, then the following lower bound holds:
    \begin{align*}
        \EE_{\thetab} \left[ \big\| \widehat{\thetab} - \thetab \big\|_2^2 \right]
        \geq \frac{d^2}{32 K^2 T}.
    \end{align*}
    To satisfy both inequalities, we require:
    \begin{align*}
        &\frac{ 2\sqrt{2} d \epsilon}{\sqrt{K^2 T}} 
        + \frac{d^2}{4 K^2 T} 
        \cdot
        P_{\thetab} [\Ecal^c]
        \geq
        \frac{d^2}{32 K^2 T}
        \\
        \Longleftrightarrow \,\,
        &  P_{\thetab} [\Ecal^c]
        \geq \frac{1}{8}
        - \frac{4 \sqrt{2} K \sqrt{T} \epsilon}{d}.
    \end{align*}
    It follows that if 
    \begin{align*}
        \frac{1}{8}
        - \frac{4 \sqrt{2} K \sqrt{T} \epsilon}{d} \geq 0.1
        \,\, \Longleftrightarrow \,\,
         \frac{0.025^2}{32} \cdot 
        \frac{d^2}{K^2 \epsilon^2} \geq T,
    \end{align*}
    then we have that $P_{\thetab} [\Ecal^c] \geq 0.1$.
    In words, this means that with constant probability, any algorithm must either collect more than $c \cdot \frac{d^2}{K^2 \epsilon^2}$ samples, or output a policy that is not $\epsilon$-optimal.
    This implies that  $T = \Omega(\frac{d^2}{K^2 \epsilon^2})$ samples are necessary to guarantee an $\epsilon$-optimal policy.
    Equivalently, after $T$ rounds, the suboptimality gap $\epsilon$ is lower bounded as
    \begin{align*}
        \SubOpt(T) = \Omega \left( \frac{d}{K \sqrt{T}} \right).
    \end{align*}
    This concludes the proof of Theorem~\ref{thm:lower_bound}.
\end{proof}

\subsection{Proof of Lemmas for Theorem~\ref{thm:lower_bound}}
\subsubsection{Proof of Lemma~\ref{lemma:KL_PL_upperbound}}
\label{subsubsec:proof_lemma:KL_PL_upperbound}
\begin{proof} [Proof of Lemma~\ref{lemma:KL_PL_upperbound}]
    By the definition of KL divergence, we have
    \begin{align*}
        D_{\text{KL}}
        \big(\PP_{\thetab}  (\cdot | S) \|\PP_{\thetab'}  (\cdot | S)   \big)
        = \EE_{\sigma \sim \PP_{\thetab}  (\cdot | S) }
        \left[
            \sum_{j=1}^{|S|}
            \left(
                \phi_{\sigma_j}^\top \left( \thetab - \thetab' \right)
                - \log \frac{ \sum_{k=j}^{|S|} e^{\phi_{\sigma_k}^\top \thetab} }{\sum_{k=j}^{|S|} e^{\phi_{\sigma_k}^\top \thetab'} }
            \right)
        \right].
        \numberthis \label{eq:lower_KL_1}
    \end{align*}
    Fix a stage $j$ and a ranking $\sigma$. 
    We define 
    \begin{align*}
        p_{k'} (\thetab) := \frac{ \exp\left( \phi_{\sigma_{k'}}^\top \thetab \right) }{ \sum_{k=j}^{|S|} \exp\left( \phi_{\sigma_k}^\top \thetab \right)},
        \quad
        \text{where } k' \in \{ j, \dots, |S| \},
    \end{align*}
    which corresponds to the Multinomial Logit (MNL) probability of selecting action $\sigma_{k'}$ at position $j$, given the parameter $\thetab$ and the choice set $S$.
    Moreover, we define
    \begin{align*}
        f(\thetab) := \log \left(\sum_{k=j}^{|S|} e^{\phi_{\sigma_k}^\top \thetab} \right).
    \end{align*}
    Then, by applying the mean value form of Taylor’s theorem, there exists $\bar{\thetab} = (1-c) \thetab + c \thetab' $ for some $c \in (0,1)$ such that
    \begin{align*}
        - \log \frac{ \sum_{k=j}^{|S|} e^{\phi_{\sigma_k}^\top \thetab} }{\sum_{k=j}^{|S|} e^{\phi_{\sigma_k}^\top \thetab'} }
        &=  f(\thetab')
        -  f(\thetab)
        \\
        &= \nabla_{\thetab} f(\thetab)^\top  \left( \thetab' - \thetab  \right)
        + \frac{1}{2}  \left( \thetab' - \thetab  \right)^\top 
        \nabla_{\thetab}^2 f(\bar{\thetab})
         \left( \thetab' - \thetab  \right)
        \tag{Taylor's theorem}
        \\
        &\leq \nabla_{\thetab} f(\thetab)^\top  \left( \thetab' - \thetab  \right)
        + \frac{1}{2} 
         \sum_{k=j}^{|S|} p_k(\bar{\thetab}) \left( \phi_{\sigma_k}^\top
        ( \thetab' - \thetab ) \right)^2
        \\
        &\leq \sum_{k=j}^{|S|}
        p_{k} (\thetab)
        \phi_{\sigma_k}^\top \left( \thetab' - \thetab  \right)
         + \frac{1}{2} 
        \sum_{a \in S }\left( \phi_{a}^\top
                ( \thetab' - \thetab ) \right)^2
        ,
        \numberthis \label{eq:lower_KL_taylor}
    \end{align*}
    where the first inequality holds because
    \begin{align*}
        \nabla_{\thetab}^2 f(\bar{\thetab})
        &= \sum_{k=j}^{|S|} p_k(\bar{\thetab})  \phi_{\sigma_k} \phi_{\sigma_k}^\top
        - \left( \sum_{k=j}^{|S|} p_k(\bar{\thetab})  \phi_{\sigma_k} \right)
        \left( \sum_{k=j}^{|S|} p_k(\bar{\thetab})  \phi_{\sigma_k} \right)^\top
        \preceq \sum_{k=j}^{|S|} p_k(\bar{\thetab})  \phi_{\sigma_k} \phi_{\sigma_k}^\top.
    \end{align*}
    Plugging Equation~\eqref{eq:lower_KL_taylor} into Equation~\eqref{eq:lower_KL_1}, we get
    \begin{align*}
        D_{\text{KL}}&
        \big(\PP_{\thetab}  (\cdot | S) \|\PP_{\thetab'}  (\cdot | S)   \big)
        \\
        &\leq \EE_{\sigma \sim \PP_{\thetab}  (\cdot | S) }
        \Bigg[
            \sum_{j=1}^{|S|}
            \bigg(
                \phi_{\sigma_j}^\top \left( \thetab - \thetab' \right)
                -
                \sum_{k=j}^{|S|}
                 p_{k} (\thetab)
                \phi_{\sigma_k}^\top \left( \thetab - \thetab'  \right)
                 + \frac{1}{2} 
                \sum_{a \in S }\left( \phi_{a}^\top
                ( \thetab' - \thetab ) \right)^2
            \bigg)
        \Bigg]
        \\
        &= 
        \EE_{\sigma \sim \PP_{\thetab}  (\cdot | S) }
        \Bigg[ \sum_{j=1}^{|S|} 
            \underbrace{\EE_{\sigma_j}
            \bigg[
                    \phi_{\sigma_j}^\top \left( \thetab - \thetab' \right)
                    -
                    \sum_{k=j}^{|S|}
                     p_{k} (\thetab)
                    \phi_{\sigma_k}^\top \left( \thetab - \thetab'  \right)
                \Bigm\vert \sigma_{1}, \dots, \sigma_{j-1}
            \bigg]}_{= 0}
        \Bigg]
        \tag{Tower rule}
        \\
        &+ \frac{|S|}{2} 
                \sum_{a \in S }\left( \phi_{a}^\top
                ( \thetab' - \thetab ) \right)^2
        \\
        &\leq \frac{K}{2} 
                \sum_{a \in S }\left( \phi_{a}^\top
                ( \thetab' - \thetab ) \right)^2,
        \tag{$|S| \leq K$}
    \end{align*}
    which concludes the proof.
\end{proof}
\section{Additional Discussions}
\label{app_sec:discussions}
This section provides further discussion of our approach.
In Subsection~\ref{app_subsec:regret}, we show how it extends to the regret-minimization setting and improves the existing $\BigO(e^B)$ dependence.
In Subsection~\ref{app_subsec:active_learning}, we describe how the approach also applies to the active-learning setting considered in~\citep{das2024active}.
\subsection{Avoiding $e^B$ Scaling in Regret Minimization}
\label{app_subsec:regret}
In this subsection, we show that our technique for eliminating the $\BigO(e^B)$ dependence in the leading term—primarily by dividing the total rounds into warm-up and non–warm-up phases—can also be applied in the regret-minimization setting (e.g.,~\citealt{bengs2022stochastic}).

First, we formally define the cumulative regret as:
\begin{align*}
    \Regret
    := \sum_{t=1}^T \Big( 
            r_{\thetab^\star} \left( x_t, \pi^\star(x_t) \right)
            - \max_{a \in S_t} r_{\thetab^\star} ( x_t, a )
    \Big),
\end{align*}
where the contexts $x_t$ can be given arbitrarily (in contrast to the main paper, where they are drawn from a fixed distribution).
We assume the linear rewards (Assumption~\ref{assum:linear_reward}) as in the main paper.
Moreover, we use the RB update rule (Procedure~\ref{alg:update_RB}) in this subsection.

Next, we define
\begin{align*}
    g_t(S) := \frac{1}{|S|} \sum_{a \in S}
         \dot{\mu}\left( (\phi(x_t, a) - \phi(x_t, \widehat{a}_t))^\top \widehat{\thetab}_t \right)
        \left\| \phi(x_t, a)
            - \phi(x_t, \widehat{a}_t)
        \right\|_{H_t^{-1}}^2,
\end{align*}
where $\widehat{a}_t := \argmax_{a \in \Acal} \phi(x_t, a)^\top \widehat{\thetab}_t$.
The quantity measures the average uncertainty between the feature vectors of actions in $S$ and that of the current best action $\widehat{a}_t$.

We then select the assortment $S_t$ in a greedy manner, following the procedure in Equation~\eqref{eq:S_t_selection_greedy}.
We initialize $S =  \{ \widehat{a}_t\}$, and subsequently add actions one by one.
Specifically, at each step, we select
\begin{align*}
    a^\star \in \argmax_{a\in\Acal\setminus S}\ 
    \Delta_t(a\mid S),
    \qquad
    \text{where }\,
    \Delta_t(a\mid S):= g_t(S  \cup \{a\}) - g_t(S),
    \numberthis \label{eq:S_t_regret}
\end{align*}
and include $a^\star$ in $S$ if $\Delta_t(a\mid S) \geq 0$. 
The process continues until either 
$|S|=K$ or no action yields a non-negative gain.
Since $g_t(\{\widehat{a}_t\}) = 0$, this rule guarantees that the selected set satisfies $|S| \geq 2$.
\begin{theorem} [Regret upper bound]
\label{thm:regret}
    In the same setting as Theorem~\ref{thm:main_RB}, suppose $T \ge e^{K/d}$.
    Then, with probability at least $1-\delta$,
    the assortment selection rule in Equation~\eqref{eq:S_t_regret}, combined with the RB loss update strategy (Procedure~\ref{alg:update_RB}), guarantees
    \begin{align*}
        \Regret = \BigOTilde \left(
            d
            \sqrt{
                \sum_{t=1}^T \frac{1}{|S_t|}
            }
            + \frac{d^2}{\kappa }
        \right).
    \end{align*}
\end{theorem}

\begin{proof}[Proof of Theorem~\ref{thm:regret}]
By the definition of the cumulative regret, we have
\begin{align*}
    \Regret &:=
      \sum_{t=1}^T \left( 
            \phi(x_t, \pi^\star(x_t))
            -\max_{a \in S_t} \phi(x_t, a)
        \right)^\top \thetab^\star
    \\
    &\leq \sum_{t=1}^T
       \left( 
            \phi(x_t, \pi^\star(x_t))
            - \phi(x_t, \widehat{a}_t)
        \right)^\top \thetab^\star
    \tag{$\widehat{a}_t \in S_t$}
    \\
    &\leq
    \sum_{t=1}^T
        \left( 
            \phi(x_t, \pi^\star(x_t))
            - \phi(x_t, \widehat{a}_t)
        \right)^\top 
        \left(
            \thetab^\star
            - \widehat{\thetab}_t
        \right).
    \tag{ $\widehat{a}_t := \argmax_{a \in \Acal} \phi(x_t, a)^\top \widehat{\thetab}_t$}
\end{align*}
Recall the definition of the set of \textit{warm-up rounds}, i.e.,
$
    \WarmupRounds= \big\{
        t \in [T]:
        \max_{a, a' \in \Acal} \| \phi(x_t, a) - \phi(x_t, a') \|_{H_t^{-1} } \geq 
        1/\beta_{T}(\delta)
    \big\}.
$
Then, we get
\begin{align*}
     \Regret 
     &\leq 
      4B |\WarmupRounds| 
      + 
      \sum_{t \notin \WarmupRounds}
        \left( 
            \phi(x_t, \pi^\star(x_t))
            - \phi(x_t, \widehat{a}_t)
        \right)^\top 
        \left(
            \thetab^\star
            - \widehat{\thetab}_t
        \right)
     \\
     &\leq 
     \BigO \left(\frac{ B}{ \kappa } \beta_{T+1}(\delta)^2
         d \log \left(
            1 + \frac{ T}{d \lambda}
         \right)
         \right)
     + \sum_{t \notin \WarmupRounds}
        \left\| 
            \phi(x_t, \pi^\star(x_t))
            - \phi(x_t, \widehat{a}_t)
        \right\|_{H_t^{-1}}
        \left\|
            \thetab^\star
            - \widehat{\thetab}_t
        \right\|_{H_t}
     \tag{Lemma~\ref{lemma:bound_Tw} and Cauchy-Schwartz ineq.}
     \\
     &\leq 
    \BigO \left(\frac{ B}{ \kappa } \beta_{T+1}(\delta)^2
         d \log \left(
            1 + \frac{ T}{d \lambda}
         \right)
         \right)
     + \beta_{T}(\delta) \sum_{t \notin \WarmupRounds}
        \left\| 
            \phi(x_t, \pi^\star(x_t))
            - \phi(x_t, \widehat{a}_t)
        \right\|_{H_t^{-1}}
    \tag{Proposition~\ref{prop:online_CB_RB}}
     .
\end{align*}
For simplify the presentation, let $\psi_{t}(a,a') = \phi(x_t, a) - \phi(x_t, a')$.
Then, for $t \notin \WarmupRounds$, we have
\begin{align*}
    \frac{1}{\dot{\mu}\left(
        \psi_{t}(\pi^\star(x_t), \widehat{a}_t)^\top \thetab^\star
    \right)}
    &= \frac{\left(1 + e^{ \psi_{t}(\pi^\star(x_t), \widehat{a}_t)^\top \thetab^\star} \right)^2}{e^{ \psi_{t }(\pi^\star(x_t), \widehat{a}_t)^\top \thetab^\star}}
    \\
    &\leq \left(1 + e^{ \psi_{t}( \pi^\star(x_t), \widehat{a}_t)^\top \thetab^\star} \right)^2
    \tag{$\psi_{t}(\pi^\star(x_t), \widehat{a}_t)^\top \thetab^\star \geq 0$}
    \\
    &\leq \left(1 + e^{ \psi_{t}(\pi^\star(x_t), \widehat{a}_t)^\top (\thetab^\star - \widehat{\thetab}_t )} \right)^2
    \tag{ $\widehat{a}_t := \argmax_{a \in \Acal} \phi(x_t, a)^\top \widehat{\thetab}_t$}
    \\
    &\leq  \left(1 + e^{\| \psi_{t}(\pi^\star(x_t), \widehat{a}_t)^\top \|_{H_t^{-1}} \|\thetab^\star - \widehat{\thetab}_t \|_{H_t}} \right)^2
    \\
    &\leq  \left(1 + e^{\frac{\beta_t(\delta)}{\beta_T(\delta)}} \right)^2
    \leq (1+e)^2.
    \tag{$t \notin \WarmupRounds$ and Proposition~\ref{prop:online_CB_RB}}
\end{align*}
Therefore, the same line of reasoning used in the proof of Theorem \ref{thm:main_RB} applies here as well.
\begin{align*}
    \sum_{t \notin   \WarmupRounds} &\|   
            \phi(x_t, \pi^\star(x_t))
            - \phi(x_t, \widehat{a}_t)
        \|_{H_t^{-1}}
    \\
    &\leq e(1+e)^2
            \sum_{t \notin   \WarmupRounds}
            \frac{1}{|S^\star_t |} 
             \sum_{a \in S^\star_t }
             \dot{\mu}\big( \psi_t(a,\widehat{a}_t)^\top \widehat{\thetab}_t \big)
             \left\| 
                        \psi_t(a,\widehat{a}_t)
                    \right\|_{H_t^{-1}}
        \tag{$S^\star_t := \{ \pi^\star(x_t), \widehat{a}_t \}$}
    \\
    &\leq 
        e(1+e)^2
         \sqrt{\sum_{t = 1}^T
            \frac{1}{|S_t|}
        }
        \sqrt{
        \sum_{t \notin   \WarmupRounds}
        \frac{|S_t|}{|S^\star_t|}
            \sum_{a \in S^\star_t }
             \dot{\mu}\big( \psi_t(a,\widehat{a}_t)^\top \widehat{\thetab}_t \big)
             \left\| 
                        \psi_t(a,\widehat{a}_t)
                    \right\|_{H_t^{-1}}^2
        }
        \tag{Cauchy-Schwartz ineq., and $\WarmupRounds \subseteq [T]$}
        \\
        &\leq
         e(1+e)^2 
         \sqrt{\sum_{t = 1}^T
            \frac{1}{|S_t|}
        }
        \sqrt{
        \sum_{t \notin   \WarmupRounds}
        \frac{\cancel{|S_t|}}{\cancel{|\textcolor{blue}{S_t}|}}
            \sum_{a \in \textcolor{blue}{S_t}  }
             \dot{\mu}\big( \psi_t(a,\widehat{a}_t)^\top \widehat{\thetab}_t \big)
             \left\| 
                        \psi_t(a,\widehat{a}_t)
                    \right\|_{H_t^{-1}}^2
        }
        \\
        &\leq
         \sqrt{2}e^2(1+e)^2 
         \sqrt{\sum_{t = 1}^T
            \frac{1}{|S_t|}
        }
        \sqrt{
            \sum_{t \notin   \WarmupRounds}
            \sum_{a \in S_t  }
             \dot{\mu}\big( \psi_t(a,\widehat{a}_t)^\top \widehat{\thetab}_t \big)
             \left\| 
                        \psi_t(a,\widehat{a}_t)
                    \right\|_{\textcolor{red}{\Sigma_t^{-1}}}^2
        }
        \tag{Lemma~\ref{lemma:H_t_Lambda}}
        \\
        &=  \BigO \!\left(
             \sqrt{\sum_{t = 1}^T
                \frac{1}{|S_t|}
            }
            \cdot
                \sqrt{d \log \left( KT \right)}
        \right)
        \tag{Lemma~\ref{lemma:EPL}}
        ,
\end{align*}
where, in the last inequality, we introduce 
$
 \Sigma_t =  \!\!\sum_{s \in [t-1] \setminus \Tcal^w} 
        \!\sum_{a \in S_s}            
            \dot{\mu}
            \left(
                \psi_t(a,\widehat{a}_t)^\top
                \widehat{\thetab}_{s}
            \right)
            \psi_t(a,\widehat{a}_t)
            \psi_t(a,\widehat{a}_t)^\top
            + \lambda \Ib_d,
$
and use the relation $H_t \succeq \frac{1}{2e^2} \Lambda_t \succeq \frac{1}{2e^2} \Sigma_t$ ($\Lambda_t$ is defined in Equation~\eqref{eq:lambda_t_def}).
In the final equality, we apply the fact that if $\lambda = d \log T$ and $ T \geq e^{K/d}$, then, for any $t$, 
\begin{align*}
    \dot{\mu}\big( \psi_t(a,\widehat{a}_t)^\top \widehat{\thetab}_t \big)
         \left\| 
                    \psi_t(a,\widehat{a}_t)
                \right\|_{\Sigma_t^{-1}}^2
    \leq \frac{2K}{\lambda} \leq 2.
\end{align*}
Therefore, using $\beta_{T}(\delta) = \mathcal{O}\big( B \sqrt{d \log (KT)} + B\sqrt{\lambda} \big)$, we obtain, with probability at least $1-\delta$, 
\begin{align*}
    \Regret = \BigOTilde \left(
            d
            \sqrt{
                \sum_{t=1}^T \frac{1}{|S_t|}
            }
            + \frac{d^2}{\kappa}
        \right).
\end{align*}
This completes the proof of Theorem~\ref{thm:regret}.
\end{proof}
\subsection{Extension to Active Learning Setting}
\label{app_subsec:active_learning}
In this subsection, we consider a different setting—referred to as the \textit{active learning setting}—where the learner has access to the entire context set $\Xcal$, and the objective is to minimize the following \textit{worst-case suboptimality gap}, defined as:
\begin{align*}
    \WosrtSubOpt(T) 
        := 
        \max_{x \in \Xcal} \left[
            r_{\thetab^\star} \left( x, \pi^\star(x) \right)
            - r_{\thetab^\star} \left( x, \widehat{\pi}(x) \right)
        \right].
\end{align*}
This setting has received increasing attention in recent work~\citep{muldrew2024active, mehta2023sample, scheid2024optimal, das2024active, mukherjee2024optimal, thekumparampil2024comparing, kveton2025active}.
However, most existing approaches focus exclusively on pairwise preference feedback.
\citet{mukherjee2024optimal} study an online learning-to-rank problem where, for each context, a fixed set of $K$ actions is provided, and the goal is to recover the true ranking based on feedback over these $K$ actions.
In contrast, we consider a more general setting in which, for each context, a set of $N$ actions is available. 
The learner selects at most $K$ actions from this set and receives ranking feedback over the selected subset.
\citet{thekumparampil2024comparing} investigate the problem of ranking $N \geq K$ items using partial rankings over $K$ candidates, but under a context-free setting.
In contrast, we study a stochastic contextual setting, where contexts are drawn from an unknown (and fixed) distribution.

In the active learning setting, the algorithm jointly selects the context $x_t$—which is no longer given but actively chosen—and the assortment $S_t$ by maximizing the average uncertainty objective. 
For each candidate context $x$, it first constructs the assortment $S_t(x)$ by solving Equation~\eqref{eq:S_t_selection_greedy}.
It then selects
 $x_t = \argmax_{x \in \Xcal} f_t(S_t(x))$, and sets $S_t = S_t(x_t)$.
The rest of the algorithm proceeds in the same manner as Algorithm~\ref{alg:main}.

For simplicity, we focus on the RB loss and its corresponding update rule (Procedure~\ref{alg:update_RB}).
Then, we can obtain the following bound on the worst-case suboptimality gap.
\begin{theorem}
\label{thm:upper_active}
    Under the same setting as Theorem~\ref{thm:main_RB}, with probability at least $1-\delta$, the selection rule for $(x_t, S_t)$ described above achieves:
    \begin{align*}
        \WosrtSubOpt(T) = 
        \BigOTilde \left(
                \frac{d}{T}
                \sqrt{
                    \sum_{t=1}^T \frac{1}{|S_t|}
                }
                + \frac{d^2}{\kappa T}
            \right).
    \end{align*}
\end{theorem}
\begin{proof} [Proof of Theorem~\ref{thm:upper_active}]
    By the definition of the worst-case suboptimality gap, we have
    \begin{align*}
        \WosrtSubOpt(T)
        &= \max_{x \in \Xcal} \left[
            \big(
                \phi \left( x, \pi^\star(x) \right)
                - \phi \left( x, \widehat{\pi}_T(x) \right)
            \big)^\top 
            \thetab^\star
        \right]
        \\
        &\leq \max_{x \in \Xcal} \left[
            \big(
                \phi \left( x, \pi^\star(x) \right)
                - \phi \left( x, \widehat{\pi}_T(x) \right)
            \big)^\top 
            \left( \thetab^\star
                - \widehat{\thetab}_{T+1}
            \right)
        \right]
        \tag{$\widehat{\pi}_T(x) = \argmax_{a \in \Acal}\phi(x,a)^\top \widehat{\thetab}_{T+1} $}
        \\
        &= \frac{1}{T}
        \sum_{t=1}^T
        \max_{x \in \Xcal} \left[
            \big(
                \phi \left( x, \pi^\star(x) \right)
                - \phi \left( x, \widehat{\pi}_T(x) \right)
            \big)^\top 
            \left( \thetab^\star
                - \widehat{\thetab}_{T+1}
            \right)
        \right]
        .
    \end{align*}
     We adopt the same definitions for $\WarmupRounds
    := \{
        t \in [T]:
        \max_{a, a' \in \Acal} \| \phi(x_t, a) - \phi(x_t, a') \|_{H_t^{-1} } \geq 
        1/\beta_{T+1}(\delta)
    \}$.
    Thus, we get
    \begin{align*}
        \frac{1}{T}
        \sum_{t=1}^T&
        \max_{x \in \Xcal} \left[
            \big(
                \phi \left( x, \pi^\star(x) \right)
                - \phi \left( x, \widehat{\pi}_T(x) \right)
            \big)^\top 
            \left( \thetab^\star
                - \widehat{\thetab}_{T+1}
            \right)
        \right]
        \\
        &\leq 
        \BigO \left(\frac{ B}{ \kappa T} \beta_{T+1}(\delta)^2
         d \log \left(
            1 + \frac{ T}{d \lambda}
         \right)
         \right)
         \tag{Lemma~\ref{lemma:bound_Tw}}
         \\&+
         \frac{1}{T}
        \sum_{t \notin \WarmupRounds}
        \max_{x \in \Xcal} \left[
            \big(
                \phi \left( x, \pi^\star(x) \right)
                - \phi \left( x, \widehat{\pi}_T(x) \right)
            \big)^\top 
            \left( \thetab^\star
                - \widehat{\thetab}_{T+1}
            \right)
        \right].
        \numberthis \label{eq:active_proof_major_term_low_EP}
    \end{align*}
    To further bound the last term of Equation~\eqref{eq:active_proof_major_term_low_EP}, we get
    \begin{align*}
        \frac{1}{T}
        &\sum_{t \notin \WarmupRounds}
        \max_{x \in \Xcal} \left[
            \big(
                \phi \left( x, \pi^\star(x) \right)
                - \phi \left( x, \widehat{\pi}_T(x) \right)
            \big)^\top 
            \left( \thetab^\star
                - \widehat{\thetab}_{T+1}
            \right)
        \right]
        \\
        &\leq 
         \frac{1}{T} \sum_{t \notin \WarmupRounds}
        \max_{x \in \Xcal} \left[
            \left\| 
                \phi \left( x, \pi^\star(x) \right)
                - \phi \left( x, \widehat{\pi}_T(x) \right)
            \right\|_{H_{T+1}^{-1} }
            \right]
            \left\|
                \thetab^\star
                - \widehat{\thetab}_{T+1}
            \right\|_{H_{T+1}}
        \tag{Hölder's ineq.}
        \\
        &\leq \frac{  \beta_{T+1} (\delta)}{T}  \sum_{t \notin \WarmupRounds}
        \max_{x \in \Xcal} \left[
            \left\| 
                 \phi \left( x, \pi^\star(x) \right)
                - \phi \left( x, \widehat{\pi}_T(x) \right)
            \right\|_{H_t^{-1} }
        \right]
        \tag{$H_{T+1} \succeq H_t$ and Corollary~\ref{cor:online_CB_PL}, with prob. $1-\delta$}
        .
    \end{align*}
    For simplicity, let $x_t^\star = \argmax_x \left\| 
                 \phi \left( x, \pi^\star(x) \right)
                - \phi \left( x, \widehat{\pi}_T(x) \right)
            \right\|_{H_t^{-1} }$.
    It therefore suffices to bound $\sum_t \left\| 
                 \phi \left( x_t^\star , \pi^\star(x) \right)
                - \phi \left( x_t^\star, \widehat{\pi}_T(x) \right)
            \right\|_{H_t^{-1} }$.
    From here, the same argument as in the proof of Theorem~\ref{thm:main_RB} applies, with the only modification that the procedure now selects both $x_t$ and $S_t$.
    This completes the proof of Theorem~\ref{thm:upper_active}.
\end{proof}

\begin{figure*}[t]
    \centering
    \begin{subfigure}[b]{0.325\textwidth}
        \includegraphics[width=\textwidth, trim=0mm 0mm 0mm 0mm, clip]{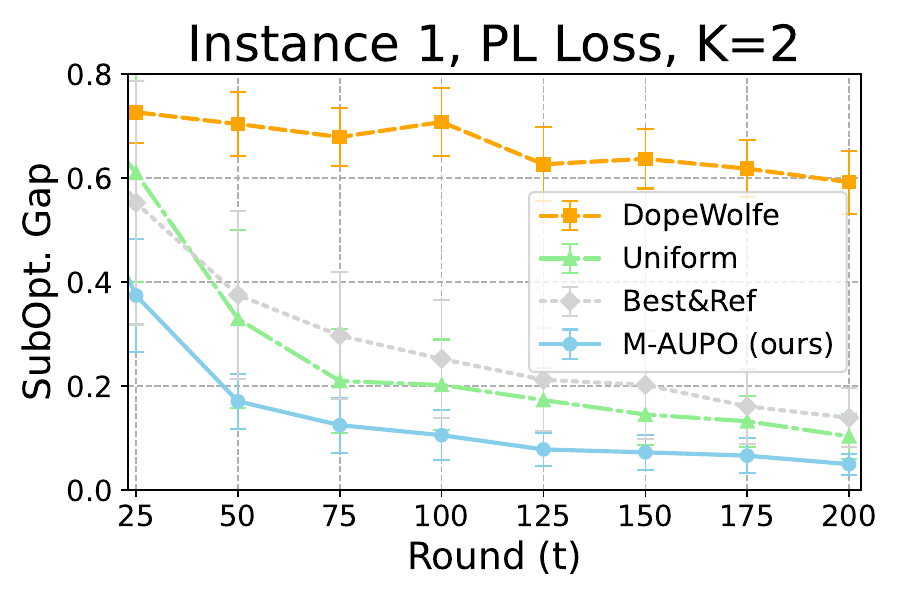}
        \label{fig:varying_K=2_pl_loss_vs_baselines}
    \end{subfigure}
    \hfill
    \begin{subfigure}[b]{0.325\textwidth}
        \includegraphics[width=\textwidth, trim=0mm 0mm 0mm 0mm, clip]{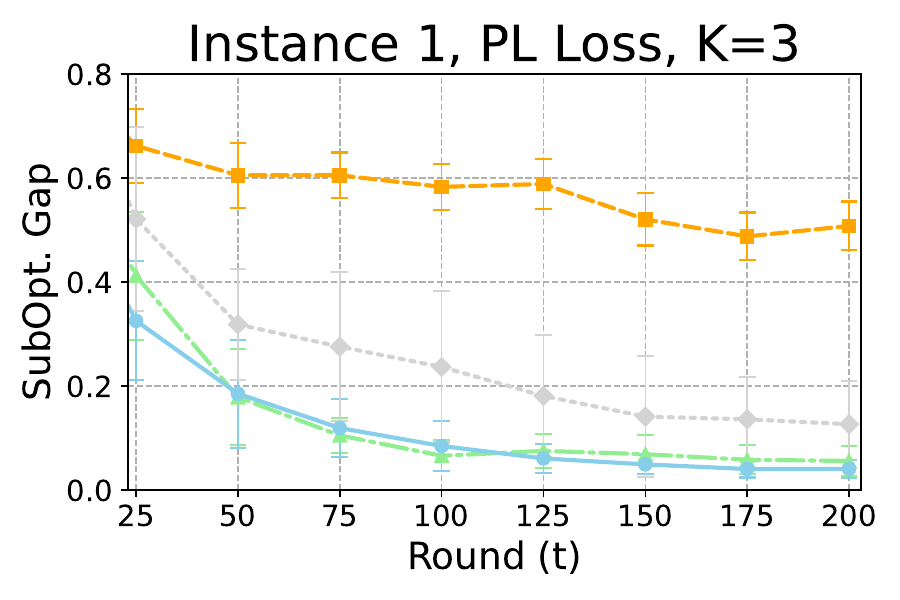}
        \label{fig:varying_K=3_pl_loss_vs_baselines}
    \end{subfigure}
    \hfill
    \begin{subfigure}[b]{0.325\textwidth}
        \includegraphics[width=\textwidth, trim=0mm 0mm 0mm 0mm, clip]{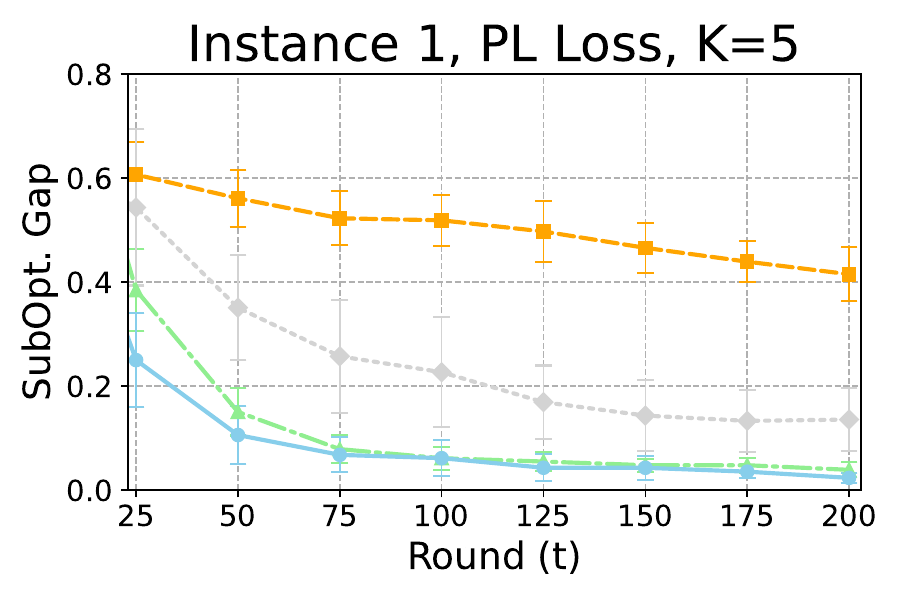}
        \label{fig:varying_K=5_pl_loss_vs_baselines}
    \end{subfigure}
    \\
    \begin{subfigure}[b]{0.325\textwidth}
        \includegraphics[width=\textwidth, trim=0mm 0mm 0mm 0mm, clip]{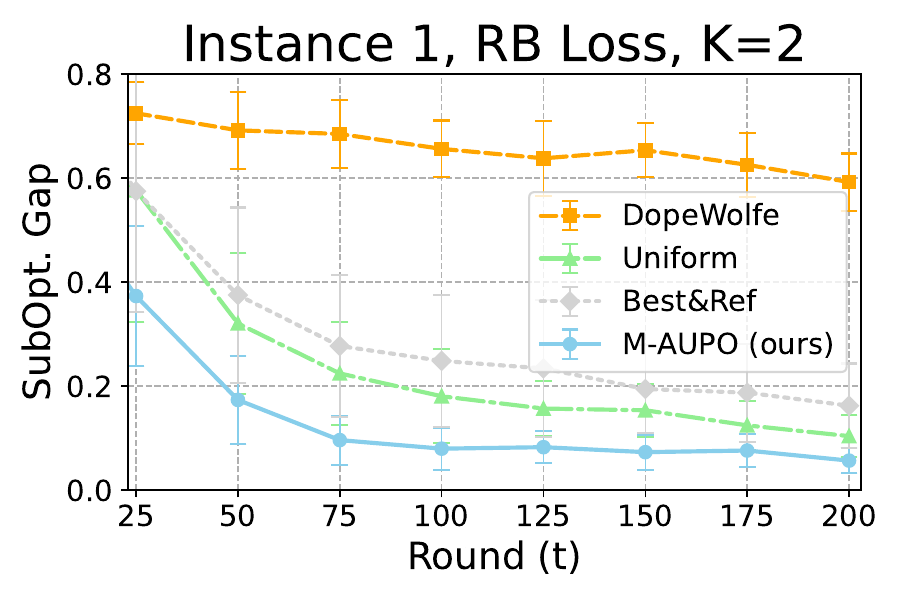}
        \label{fig:varying_K=2_rb_loss_vs_baselines}
    \end{subfigure}
    \hfill
    \begin{subfigure}[b]{0.325\textwidth}
        \includegraphics[width=\textwidth, trim=0mm 0mm 0mm 0mm, clip]{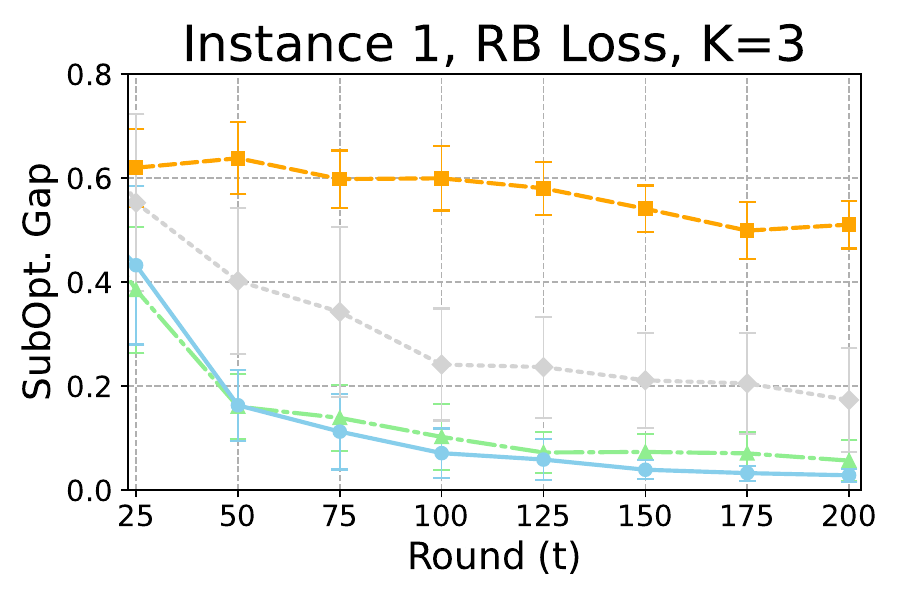}
        \label{fig:varying_K=3_rb_loss_vs_baselines}
    \end{subfigure}
    \hfill
    \begin{subfigure}[b]{0.325\textwidth}
        \includegraphics[width=\textwidth, trim=0mm 0mm 0mm 0mm, clip]{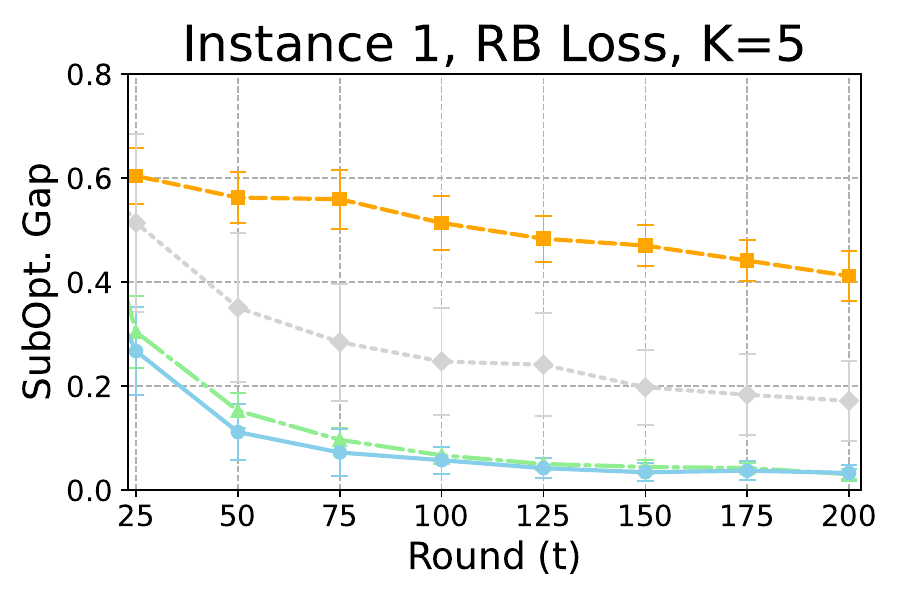}
        \label{fig:varying_K=5_rb_loss_vs_baselines}
    \end{subfigure}
    \caption{Performance comparisons for Instance 1 (Stochastic contexts) with $K = 2$, $3$, and $5$, evaluated under the PL loss (first row) and RB loss (second row).}
    \label{fig:app_synthetic_experiment_comparison_instance1}
\end{figure*}
%
\section{Experimental Details and Additional Results}
\label{app_sec:experimat_details}
%
%
\subsection{Synthetic Data}
\label{app_subsec:exp_synthetic}
\textbf{Setup.}
In the synthetic data experiment, we sample the true but unknown parameter $\thetab^\star \in \mathbb{R}^d$ from a $d$-dimensional standard normal distribution, i.e., $\thetab^\star \sim \mathcal{N}(0, I_d)$, and then normalize it to ensure $\|\thetab^\star\|_2 \leq 1$.
We consider four different types of context sets $\Xcal$:
\begin{enumerate}
    \item \textbf{Instance 1 (Stochastic contexts)}: For each $x \in \Xcal$, the feature vectors $\phi(x, \cdot)$ are sampled from a standard normal distribution and then normalized to satisfy $\|\phi(x, \cdot)\|_2 \leq 1$. Here, $|\Xcal| = 100$.
    
    \item \textbf{Instance 2 (Non-contextual)}: A single shared context is used for all rounds, i.e., $\Xcal = \{x_1\}$ and $|\Xcal| = 1$. 
    The corresponding feature vectors $\phi(x_1, \cdot)$ are sampled from a standard normal distribution and then normalized to satisfy $\|\phi(x_1, \cdot)\|_2 \leq 1$.
    
    \item \textbf{Instance 3 (Hard-to-learn contexts)}: For each $x \in \Xcal$, the feature vectors $\phi(x, \cdot)$ are constructed such that most of them are approximately orthogonal to the true parameter $\thetab^\star$. Here, $|\Xcal| = 100$.
    
    \item \textbf{Instance 4 (Skewed stochastic contexts)}: For each $x \in \Xcal$, the feature vectors $\phi(x, \cdot)$ are sampled in a skewed or biased manner and then normalized to satisfy $\|\phi(x, \cdot)\|_2 \leq 1$. Here, $|\Xcal| = 100$.
    This is our main experimental setup in Section~\ref{subsec:exp_synthetic}.
\end{enumerate}
Additionally, we set the feature dimension to $d = 5$ and the number of available actions to $|\Acal| = N = 100$.
The suboptimality gap is measured every 25 rounds.
All results are averaged over 20 independent runs with different random seeds, and standard errors are reported to indicate variability.
The experiments are run on a Xeon(R) Gold 6226R CPU @ 2.90GHz (16 cores).

\textbf{Baselines.}
We evaluate our proposed algorithm, \AlgName{}, against three baselines:
(i) \texttt{DopeWolfe}~\citep{thekumparampil2024comparing}, a method designed for non-contextual $K$-subset selection;
(ii) \texttt{Uniform}, which selects assortments of size $K$ uniformly at random; and
(iii) \texttt{Best\&Ref}, which forms a pair of actions ($|S_t|=2$) by combining one action from the current policy with another from a reference policy (e.g., uniform random or SFT), following the setup in Online GSHF~\citep{xiong2023iterative} and XPO~\citep{xie2025exploratory}.

When using the PL loss–based update in our algorithm, \AlgName{}, the exact expectation over rankings is computationally expensive. 
To mitigate this, we approximate the expectation via Monte Carlo sampling with $5$ samples. 
Consequently, the computational cost of the PL-based approach is higher than that of the RB-based approach (see Table~\ref{tab:comput_cost}).

\citet{thekumparampil2024comparing} propose a D-optimal design approach for the Plackett-Luce objective to efficiently select informative subsets of items for comparison.
Recognizing the computational complexity inherent in this method, they introduce a randomized Frank-Wolfe algorithm, named \texttt{DopeWolfe}, which approximates the optimal design by solving linear maximization sub-problems on randomly chosen variables. 
This approach reduces computational overhead while maintaining effective learning performance.
However, their approach is specifically tailored to the single-context setting (e.g., \textbf{Instance 2}) and may not generalize well to the multiple-context scenarios (e.g., \textbf{Instances 1, 3,} and \textbf{4}).
While their original implementation updates the model parameters using a maximum likelihood estimation (MLE) procedure, we instead adopt an online update strategy (as described in Procedures~\ref{alg:update_PL} and~\ref{alg:update_RB}) to ensure a fair comparison across all methods.
For sampling size parameter $R$, we set $R = \min\{\binom{N}{K},  100,000 \}$.

The uniform random selection strategy, \texttt{Uniform}, selects $K$ actions uniformly at random from the available action set $\Acal$ at each round, without utilizing any uncertainty or reward-based information.

\texttt{Best\&Ref} constructs an action pair ($|S_t|=2$) by combining two distinct sources of actions. 
The first action is chosen to maximize the current reward estimate, while the second is sampled from a reference policy—such as a uniform random policy or a supervised fine-tuned (SFT) model.
This pairing mechanism follows the framework introduced in Online GSHF~\citep{xiong2023iterative} and XPO~\citep{xie2025exploratory}.
In our experiments, we use the uniform random policy as the reference.

\textbf{Performance measure.}
Since computing the exact suboptimality gap is challenging under a general distribution $\rho$, we instead evaluate the \textit{realized regret}, which serves as a slightly relaxed proxy for the suboptimality gap.
\begin{align*}
    \SubOpt(T)
    &\lesssim \frac{1}{T} \sum_{t=1}^T 
        \big(
                \phi \left( x_t, \pi^\star(x_t) \right)
                - \phi \left( x_t, \widehat{\pi}_T(x_t) \right)
            \big)^\top 
            \thetab^\star
        + \underbrace{\BigOTilde\left(\frac{1}{\sqrt{T}}\right)}_{\text{incurred by MDS terms}}
    \\
    &\leq \underbrace{\frac{1}{T} \sum_{t=1}^T 
        \big(
                \phi \left( x_t, \pi^\star(x_t) \right)
                - \phi \left( x_t, \pi_t(x_t) \right)
            \big)^\top 
            \thetab^\star}_{=:\textit{realized regret}}
        + \BigOTilde\left(\frac{1}{\sqrt{T}}\right),
\end{align*}
where we define $\pi_t(x) := \argmax_a \phi(x,a)^\top \widehat{\thetab}_t$,
and let
$ \widehat{\pi}_T$ denote the best policy among $\{ \pi_t \}_{t=1}^T$, possibly selected using a validation set.

\begin{figure*}[t]
    \centering
    \begin{subfigure}[b]{0.325\textwidth}
        \includegraphics[width=\textwidth, trim=0mm 0mm 0mm 0mm, clip]{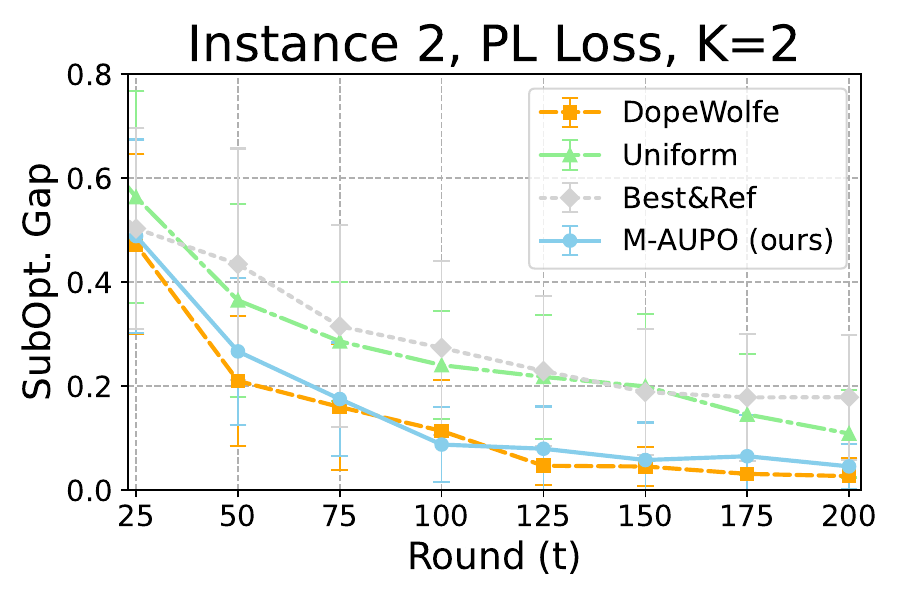}
        \label{fig:fixed_K=2_pl_loss_vs_baselines}
    \end{subfigure}
    \hfill
    \begin{subfigure}[b]{0.325\textwidth}
        \includegraphics[width=\textwidth, trim=0mm 0mm 0mm 0mm, clip]{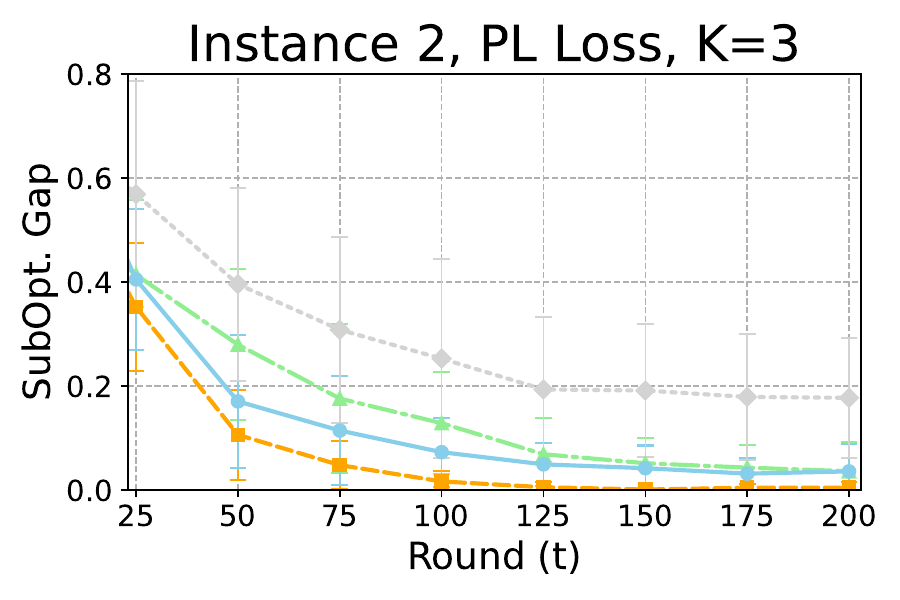}
        \label{fig:fixed_K=3_pl_loss_vs_baselines}
    \end{subfigure}
    \hfill
    \begin{subfigure}[b]{0.325\textwidth}
        \includegraphics[width=\textwidth, trim=0mm 0mm 0mm 0mm, clip]{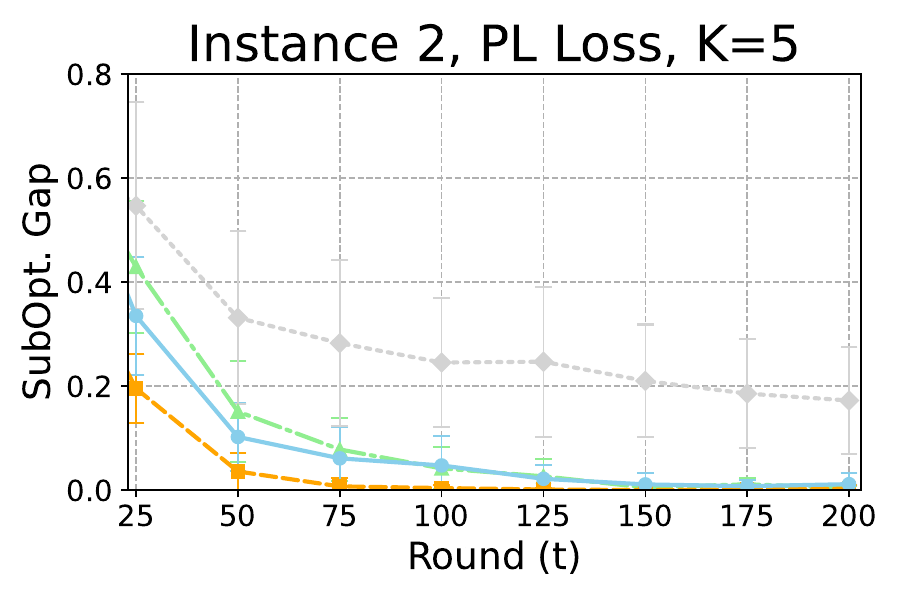}
        \label{fig:fixed_K=5_pl_loss_vs_baselines}
    \end{subfigure}
    \\
    \begin{subfigure}[b]{0.325\textwidth}
        \includegraphics[width=\textwidth, trim=0mm 0mm 0mm 0mm, clip]{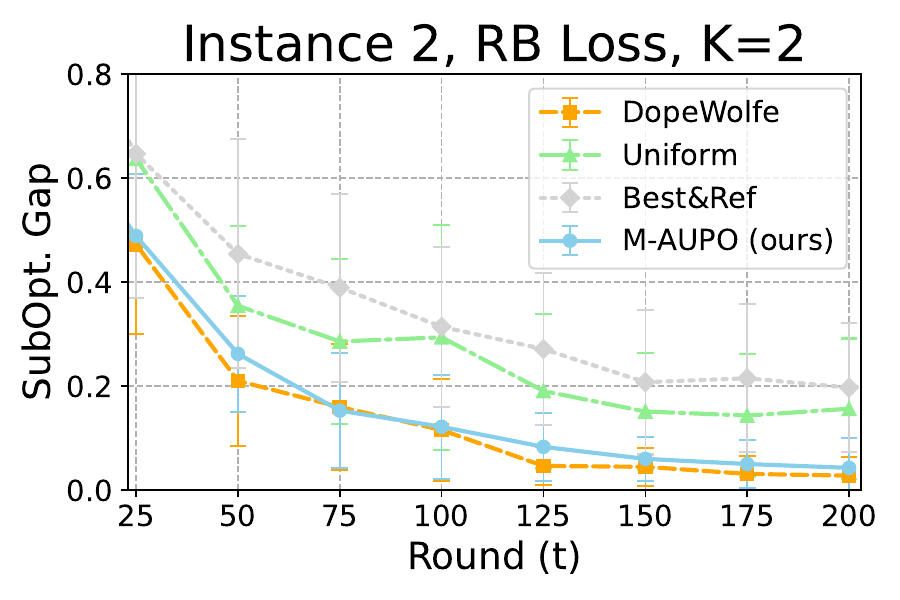}
        \label{fig:fixed_K=2_rb_loss_vs_baselines}
    \end{subfigure}
    \hfill
    \begin{subfigure}[b]{0.325\textwidth}
        \includegraphics[width=\textwidth, trim=0mm 0mm 0mm 0mm, clip]{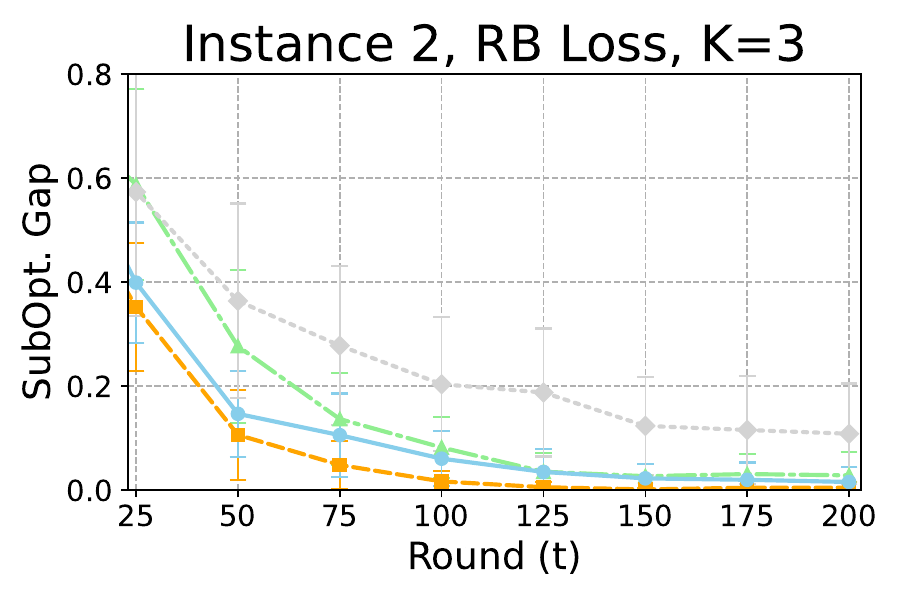}
        \label{fig:fixed_K=3_rb_loss_vs_baselines}
    \end{subfigure}
    \hfill
    \begin{subfigure}[b]{0.325\textwidth}
        \includegraphics[width=\textwidth, trim=0mm 0mm 0mm 0mm, clip]{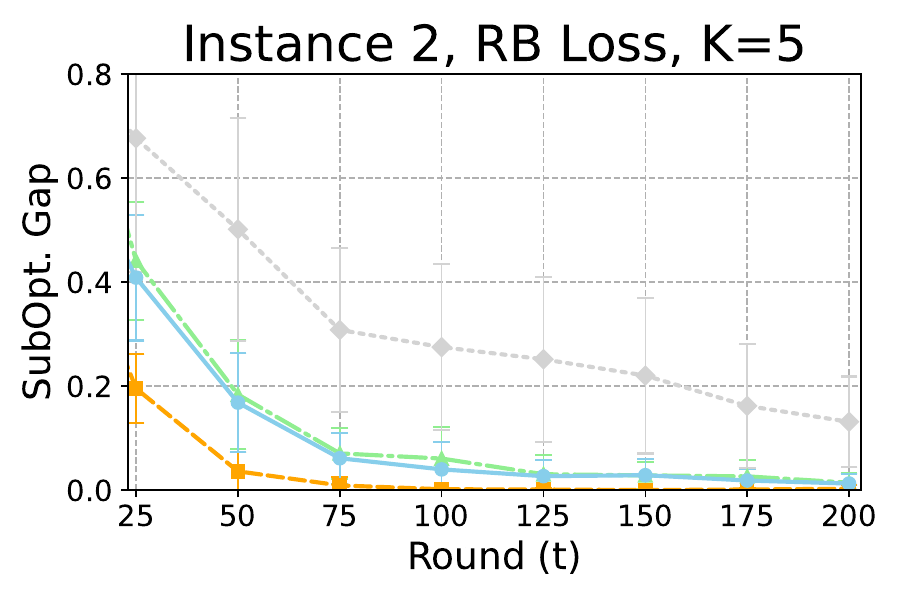}
        \label{fig:fixed_K=5_rb_loss_vs_baselines}
    \end{subfigure}
    \caption{Performance comparisons for Instance 2 (Non-contextual) with $K = 2$, $3$, and $5$, evaluated under the PL loss (first row) and RB loss (second row).}
    \label{fig:app_synthetic_experiment_comparison_instance2}
\end{figure*}

\begin{figure*}[t]
    \centering
    \begin{subfigure}[b]{0.325\textwidth}
        \includegraphics[width=\textwidth, trim=0mm 0mm 0mm 0mm, clip]{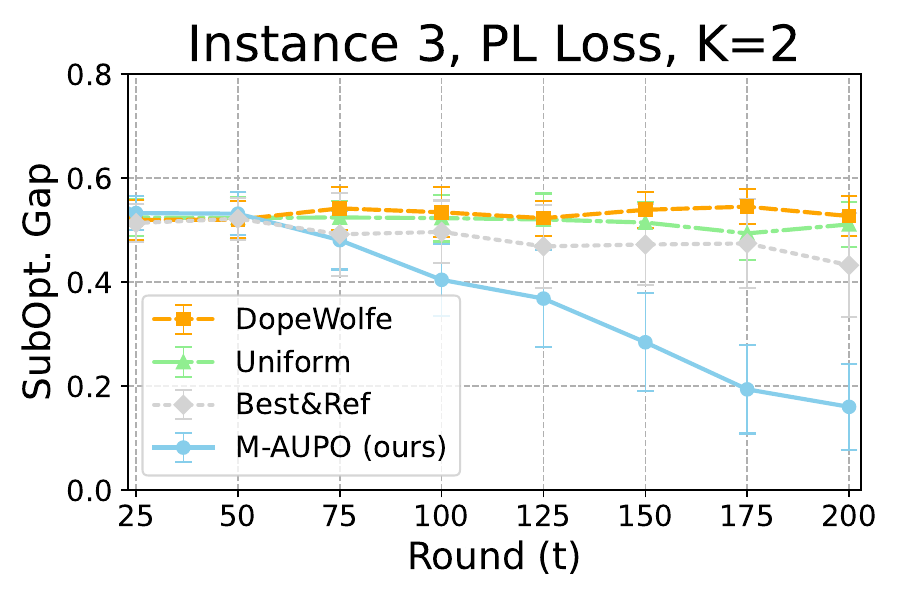}
        \label{fig:hard_instance_K=2_pl_loss_vs_baselines}
    \end{subfigure}
    \hfill
    \begin{subfigure}[b]{0.325\textwidth}
        \includegraphics[width=\textwidth, trim=0mm 0mm 0mm 0mm, clip]{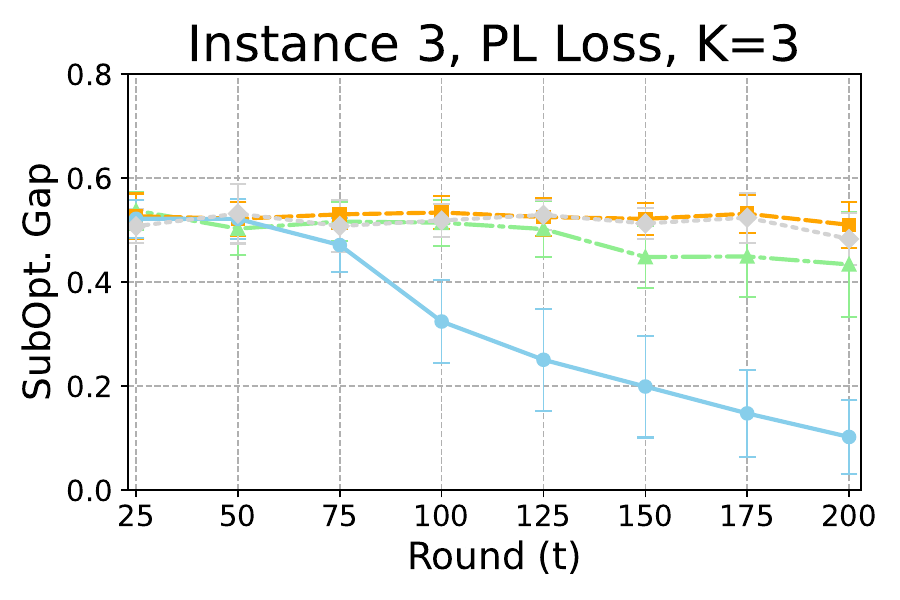}
        \label{fig:hard_instance_K=3_pl_loss_vs_baselines}
    \end{subfigure}
    \hfill
    \begin{subfigure}[b]{0.325\textwidth}
        \includegraphics[width=\textwidth, trim=0mm 0mm 0mm 0mm, clip]{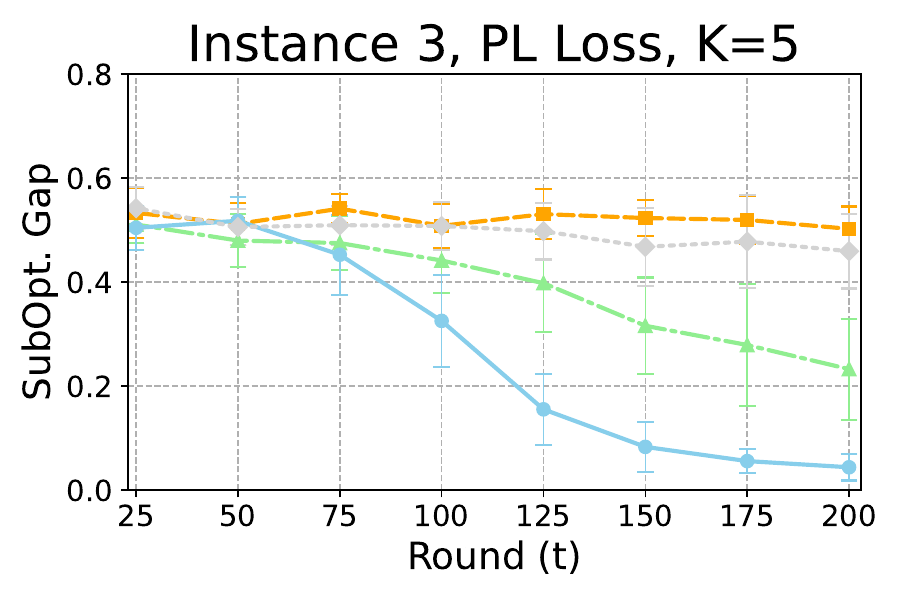}
        \label{fig:hard_instance_K=5_pl_loss_vs_baselines}
    \end{subfigure}
    \\
    \begin{subfigure}[b]{0.325\textwidth}
        \includegraphics[width=\textwidth, trim=0mm 0mm 0mm 0mm, clip]{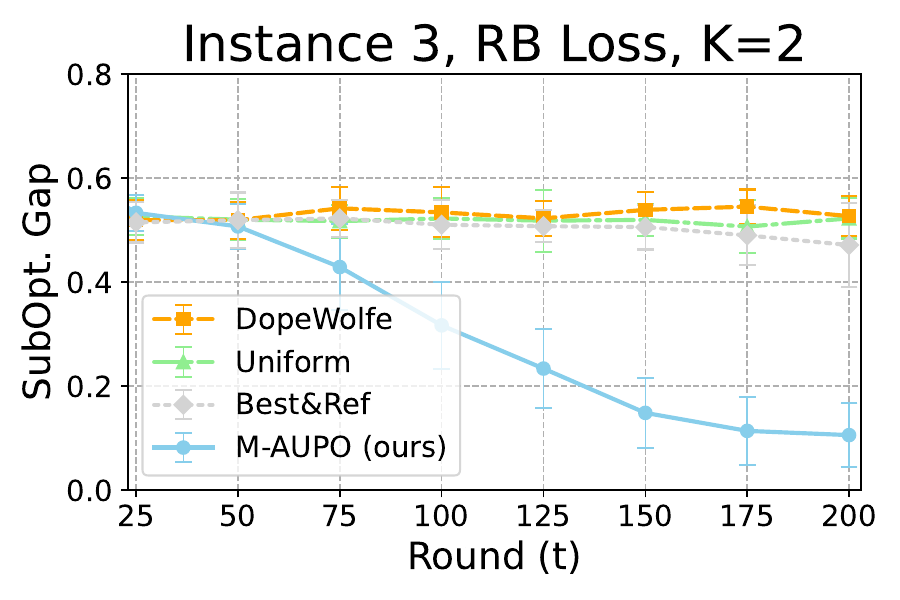}
        \label{fig:hard_instance_K=2_rb_loss_vs_baselines}
    \end{subfigure}
    \hfill
    \begin{subfigure}[b]{0.325\textwidth}
        \includegraphics[width=\textwidth, trim=0mm 0mm 0mm 0mm, clip]{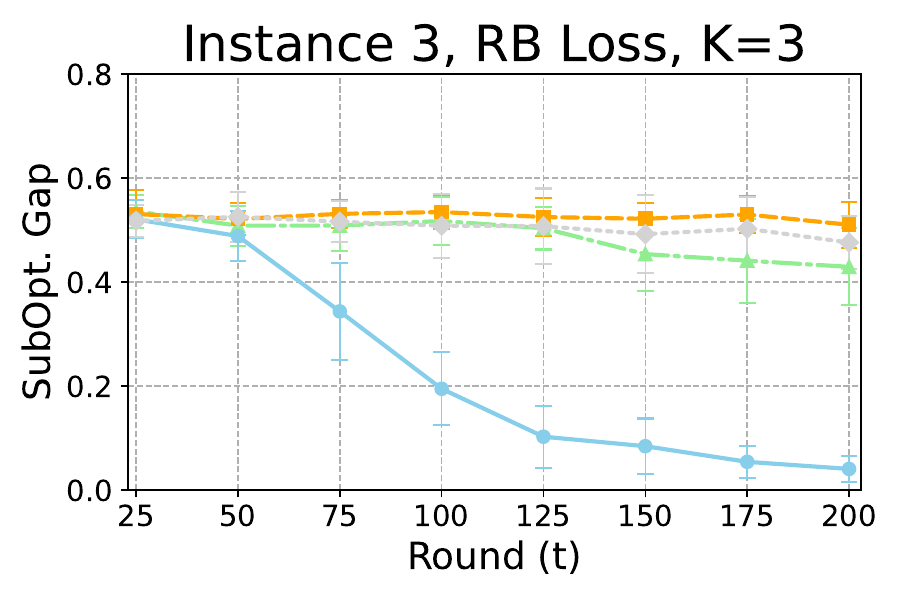}
        \label{fig:hard_instance_K=3_rb_loss_vs_baselines}
    \end{subfigure}
    \hfill
    \begin{subfigure}[b]{0.325\textwidth}
        \includegraphics[width=\textwidth, trim=0mm 0mm 0mm 0mm, clip]{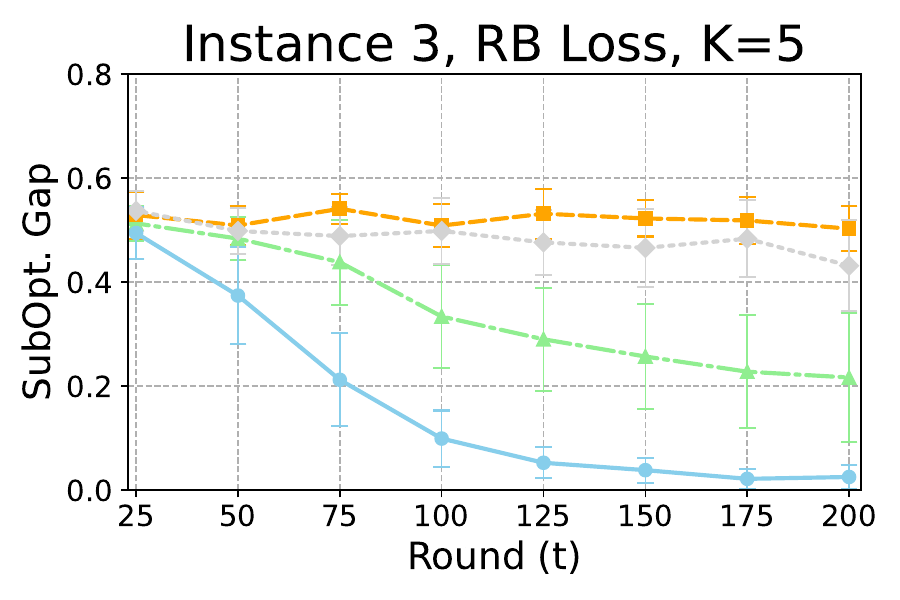}
        \label{fig:hard_instance_K=5_rb_loss_vs_baselines}
    \end{subfigure}
    \caption{Performance comparisons for Instance 3 (Hard-to-learn contexts) with $K = 2$, $3$, and $5$, evaluated under the PL loss (first row) and RB loss (second row).}
    \label{fig:app_synthetic_experiment_comparison_instance3}
\end{figure*}
\begin{figure*}[t]
    \centering
    \begin{subfigure}[b]{0.325\textwidth}
        \includegraphics[width=\textwidth, trim=0mm 0mm 0mm 0mm, clip]{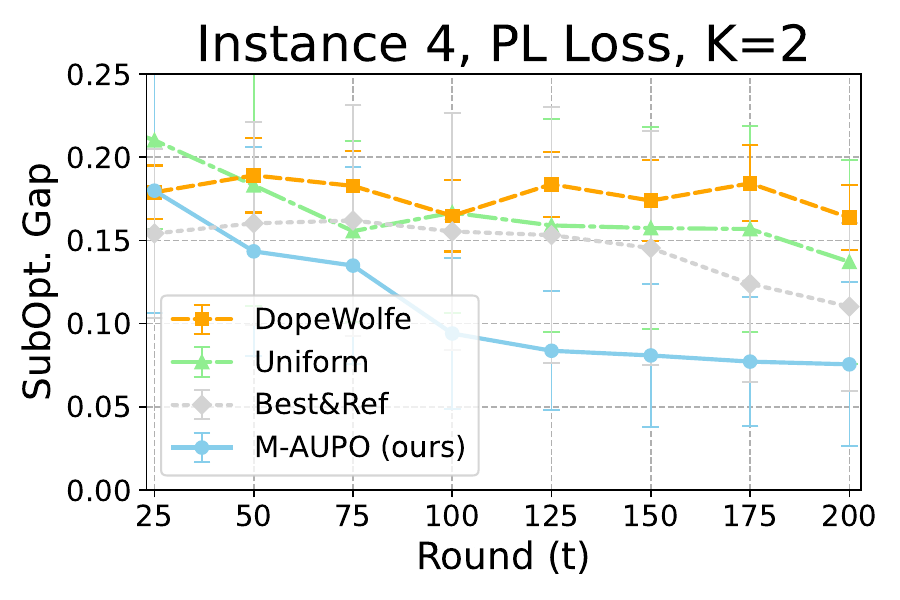}
        \label{fig:skewed_but_varying_K=2_pl_loss_vs_baselines}
    \end{subfigure}
    \hfill
    \begin{subfigure}[b]{0.325\textwidth}
        \includegraphics[width=\textwidth, trim=0mm 0mm 0mm 0mm, clip]{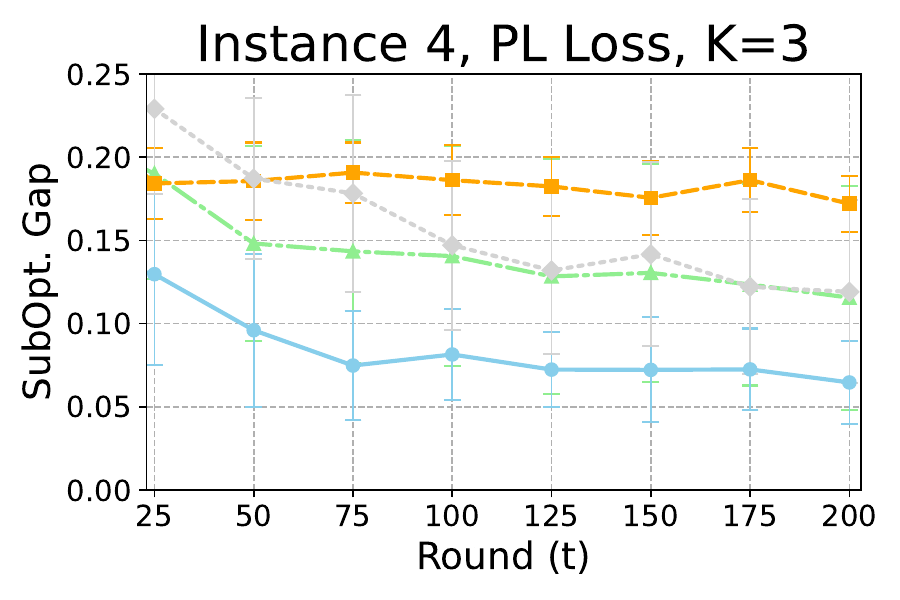}
        \label{fig:skewed_but_varying_K=3_pl_loss_vs_baselines}
    \end{subfigure}
    \hfill
    \begin{subfigure}[b]{0.325\textwidth}
        \includegraphics[width=\textwidth, trim=0mm 0mm 0mm 0mm, clip]{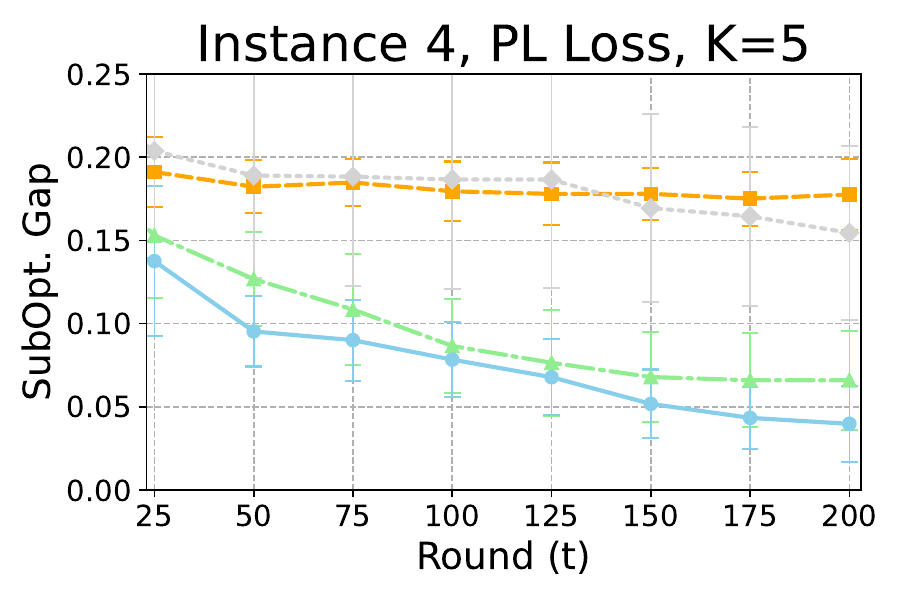}
        \label{fig:skewed_but_varying_K=5_pl_loss_vs_baselines}
    \end{subfigure}
    \\
    \begin{subfigure}[b]{0.325\textwidth}
        \includegraphics[width=\textwidth, trim=0mm 0mm 0mm 0mm, clip]{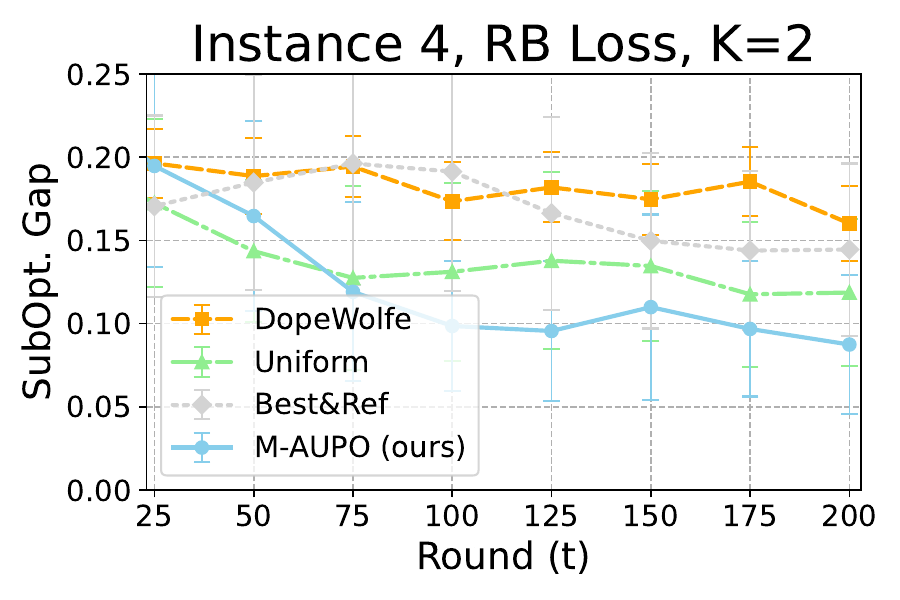}
        \label{fig:skewed_but_varying_K=2_rb_loss_vs_baselines}
    \end{subfigure}
    \hfill
    \begin{subfigure}[b]{0.325\textwidth}
        \includegraphics[width=\textwidth, trim=0mm 0mm 0mm 0mm, clip]{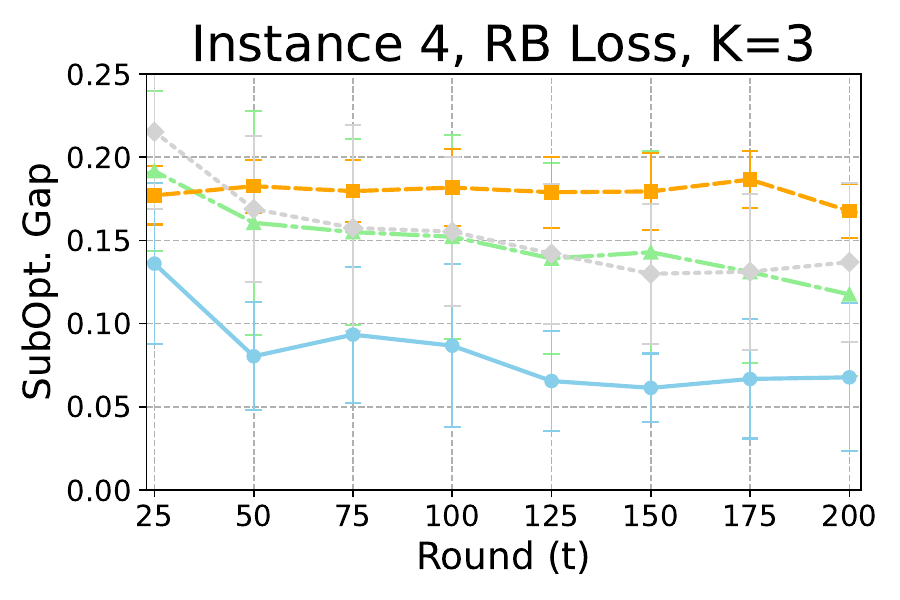}
        \label{fig:skewed_but_varying_K=3_rb_loss_vs_baselines}
    \end{subfigure}
    \hfill
    \begin{subfigure}[b]{0.325\textwidth}
        \includegraphics[width=\textwidth, trim=0mm 0mm 0mm 0mm, clip]{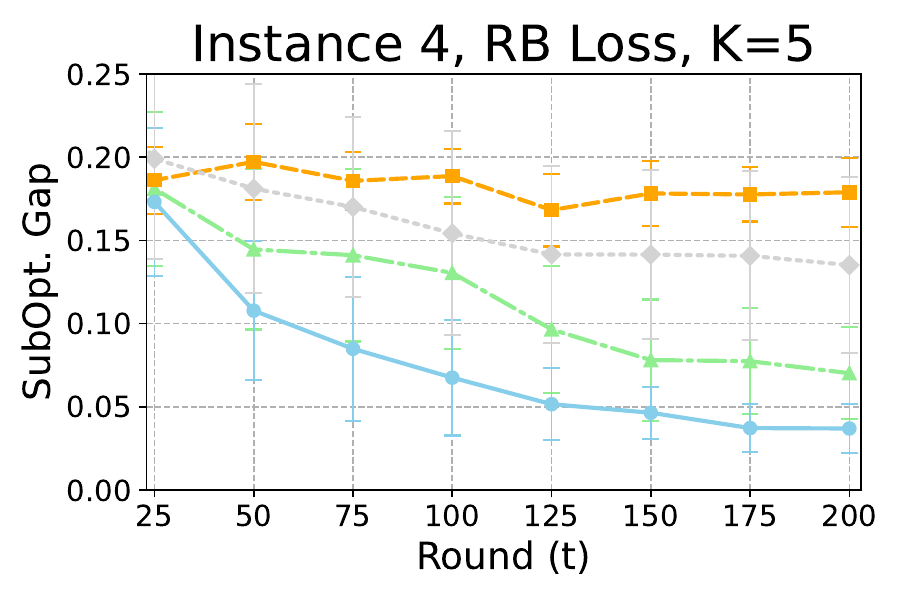}
        \label{fig:skewed_but_varying_K=5_rb_loss_vs_baselines}
    \end{subfigure}
    \caption{Performance comparisons for Instance 4 (skewed stochastic contexts) with $K = 2$, $3$, and $5$, evaluated under the PL loss (first row) and RB loss (second row).}
    \label{fig:app_synthetic_experiment_comparison_instance4}
\end{figure*}

\textbf{Results.}
We present performance comparisons in Figures~\ref{fig:app_synthetic_experiment_comparison_instance1} through~\ref{fig:app_synthetic_experiment_comparison_instance4}, corresponding to Instances 1 through 4, respectively.
Overall, our algorithm, \AlgName{}, consistently outperforms other baseline methods.
The only exception is in Instance 2 (Figure~\ref{fig:app_synthetic_experiment_comparison_instance2}), a special case of the non-contextual setting, where \AlgName{} performs slightly worse than \texttt{DopeWolfe}.
This is an expected outcome, as \texttt{DopeWolfe} leverages a D-optimal design strategy, which is known to be highly effective in the single-context setting.
However, it is important to note that \texttt{DopeWolfe} completely fails in more general contextual scenarios (Figures~\ref{fig:app_synthetic_experiment_comparison_instance1}, \ref{fig:app_synthetic_experiment_comparison_instance3}, and \ref{fig:app_synthetic_experiment_comparison_instance4}), and its computational cost is significantly higher than that of our approach (see Table~\ref{tab:comput_cost}).
\begin{table*}[h!]
\centering
    \begin{tabular}{rrrrrr}
    \toprule
    $K$ & DopeWolfe & Uniform & \texttt{Best\&Ref} & M-AUPO (PL) & M-AUPO (RB) \\
    \midrule
    2   & 7.28 s      & 0.10 s & 0.10s  & 1.09 s & 1.04 s  \\
    3   & 99.6 s     & 0.18 s  & 0.10s  & 2.35 s & 1.06 s  \\
    5   & 150.5 s    & 0.35 s  & 0.10s  & 2.71 s & 1.10 s  \\
    7   & 218.8 s    & 0.58 s  & 0.10s  & 4.10 s & 1.13 s  \\
    10  & 331.1 s    & 0.99 s  & 0.10s  & 6.89 s & 1.31 s  \\
    \bottomrule
    \end{tabular}
\caption{Runtime comparison over 200 rounds (seconds)}
\label{tab:comput_cost}
\end{table*}

The uniform random assortment selection strategy, \texttt{Uniform}, shows reasonable performance—though consistently inferior to \AlgName{}—in Instances 1, 2, and 4, as presented in Figures~\ref{fig:app_synthetic_experiment_comparison_instance1}, \ref{fig:app_synthetic_experiment_comparison_instance2}, and \ref{fig:app_synthetic_experiment_comparison_instance4}, respectively.
In contrast, for Instance 3 (Figure~\ref{fig:app_synthetic_experiment_comparison_instance3}), where most features are uninformative (being nearly orthogonal to the true parameter) and thus careful action selection becomes crucial, \texttt{Uniform} performs substantially worse than \AlgName{}.

The \texttt{Best\&Ref} algorithm performs consistently worse than our algorithm and does not benefit from larger $K$, since it always selects only a pair of actions.

Moreover, the suboptimality gap consistently decreases with larger $K$ across the three algorithms—\AlgName{}, \texttt{Uniform}, and \texttt{DopeWolfe}—whereas \texttt{Best\&Ref} shows no such improvement, as it always selects only two actions regardless of $K$.
For \AlgName{}, this trend is consistent with our theoretical results (Theorems~\ref{thm:main_PL} and \ref{thm:main_RB}).
In contrast, the improvement observed for \texttt{DopeWolfe} suggests that its current theoretical guarantees may be loose, as their bound actually deteriorates with increasing $K$ (recall that their theoretical guarantee worsens for larger $K$).
This suggests that tighter bounds could be obtained by incorporating techniques similar to those introduced in our work.

Table~\ref{tab:synthetic_assortment_size} presents the average assortment size $|S_t|$ of \AlgName{} for various values of the maximum assortment size $K$.
In most cases, the algorithm selects the full $K$ actions, i.e., $|S_t| = K$.
An exception occurs when $K$ is large (e.g., 30 or more), which may be impractical in real-world applications due to the increased annotation burden on human labelers.
\begin{table*}[h!]
\centering
    \begin{tabular}{cl|rrrrrrr}
    \toprule
    \multicolumn{2}{c|}{$K$}     & 2 & 3 & 5 & 7 & 10 & 30   & 50    \\
    \midrule
    PL loss, &$\!\!\!\!\!|S_t|$  & 2.00 & 3.00 & 5.00 & 7.00 & 10.00 & 16.21 & 18.78 \\
    RB loss, &$\!\!\!\!\!|S_t|$  & 2.00 & 3.00 & 5.00 & 7.00 & 10.00 & 19.87 & 22.37 \\
    \bottomrule
    \end{tabular}
\caption{Assortment size $|S_t|$ of \AlgName{} with varying $K$ in the synthetic experiment}
\label{tab:synthetic_assortment_size}
\end{table*}


\subsection{Real-World Dataset}
\label{app_subsec:exp_LLM_dataset}
\textbf{Setup.}
In our real-world dataset experiments, we evaluate performance on two widely used benchmark datasets: TREC Deep Learning (TREC-DL) and NECTAR.
The TREC-DL dataset provides 100 candidate answers for each query, offering a rich and diverse set of responses suitable for learning from listwise feedback.
In contrast, the NECTAR dataset presents a more concise setup, with only 7 candidate answers per question.
From each dataset, we randomly sample $|\Xcal| = 5000$ prompts, each paired with its corresponding set of candidate actions—100 for TREC-DL and 7 for NECTAR.

We use the Gemma-2B language model~\citep{team2024gemma} to construct the feature representation $\phi(x, a)$.
To obtain $\phi(x, a)$, we first concatenate the input prompt $x$ and the candidate response $a$ into a single sequence, which is then fed into Gemma-2B.
The resulting feature vector is extracted from the last hidden layer of the model and has a dimensionality of $d = 2048$.
We then apply $\ell_1$ normalization to enhance numerical stability and ensure consistent scaling.
For each round $t$, we sample the context index from an exponential distribution with rate $\lambda = 0.1$, which assigns higher probability to smaller indices and thus biases the selection toward earlier contexts.
To generate ranking feedback and evaluate the suboptimality gap, we use the Mistral-7B reward model~\citep{jiang2023mistral7b} as the ground-truth reward function.

We measure the suboptimality gap every 2,500 rounds throughout the training process and report the average performance over 10 independent runs, each with a different random seed.
Along with the average, we also include the standard error to indicate variability across runs.
In these experiments, we report results under the RB loss only, due to its superior performance, as demonstrated in the synthetic data experiments.
The experiments are conducted on a Xeon(R) Gold 6226R CPU @ 2.90GHz (16 cores) and a single GeForce RTX 3090 GPU.

\textbf{Baselines.}
We use the same set of baselines as in the synthetic data experiments.
For \texttt{DopeWolfe}~\citep{thekumparampil2024comparing}, we set the sampling size parameter $R$ as $R = \min\{\binom{N}{K}, 1000\}$.
Although a small value of $R \leq 1000$ may introduce significant approximation error—since the theoretically minimal-error choice is $R = \mathcal{O}\big(\binom{N}{K}\big)$—we adopt this smaller value in our experiment to reduce computational overhead.

\begin{figure*}[t]
    \centering
    \begin{subfigure}[b]{0.315\textwidth}
        \includegraphics[width=\textwidth, trim=0mm 0mm 0mm 0mm, clip]{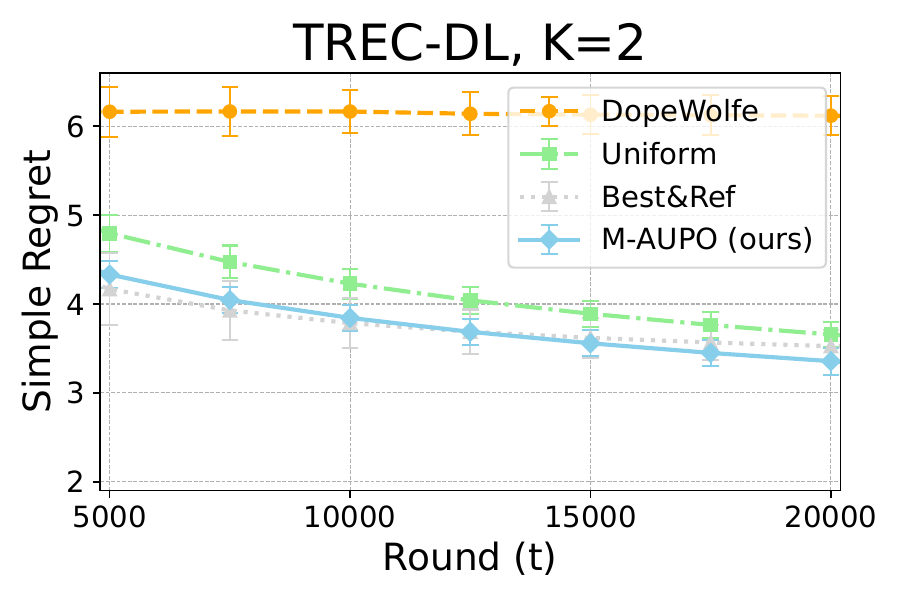}
        \label{fig:trec_comparison_K=2}
    \end{subfigure}
    \hspace{0.01\textwidth}
    \begin{subfigure}[b]{0.315\textwidth}
        \includegraphics[width=\textwidth, trim=0mm 0mm 0mm 0mm, clip]{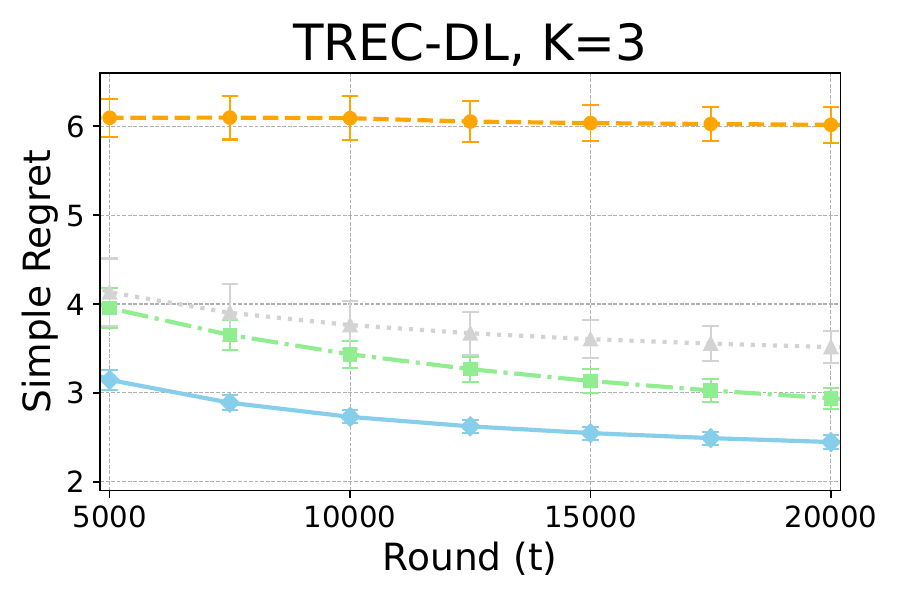}
        \label{fig:trec_comparison_K=3}
    \end{subfigure}
    \hspace{0.01\textwidth}
    \begin{subfigure}[b]{0.315\textwidth}
        \includegraphics[width=\textwidth, trim=0mm 0mm 0mm 0mm, clip]{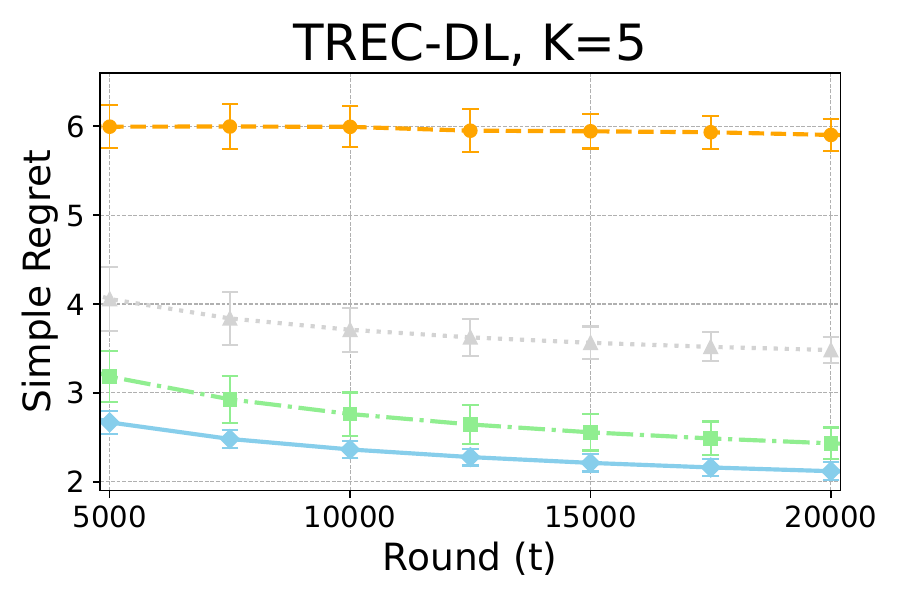}
        \label{fig:trec_comparison_K=5}
    \end{subfigure}
    \\
    \begin{subfigure}[b]{0.325\textwidth}
        \includegraphics[width=\textwidth, trim=0mm 0mm 0mm 0mm, clip]{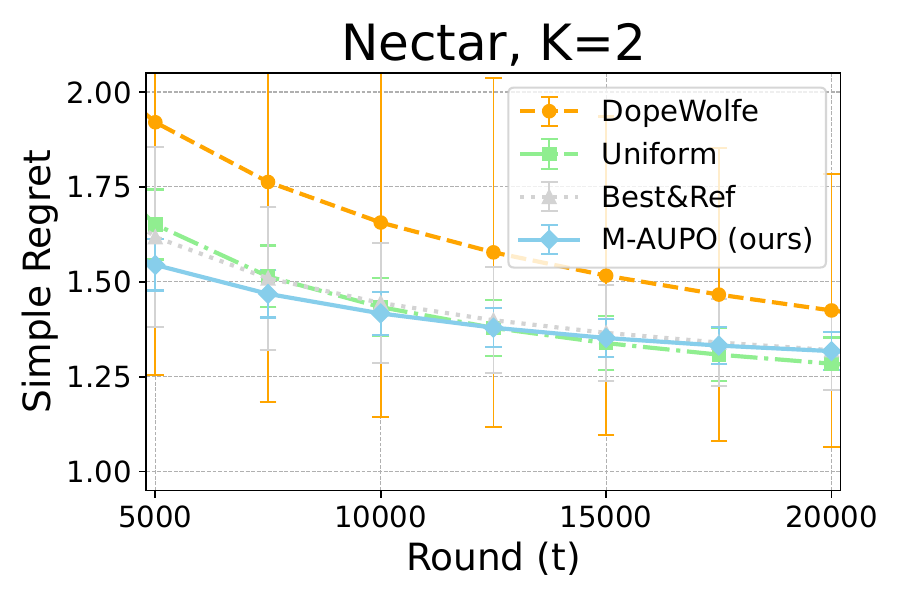}
        \label{fig:nectar_comparison_K=2}
    \end{subfigure}
    \hfill
    \begin{subfigure}[b]{0.325\textwidth}
        \includegraphics[width=\textwidth, trim=0mm 0mm 0mm 0mm, clip]{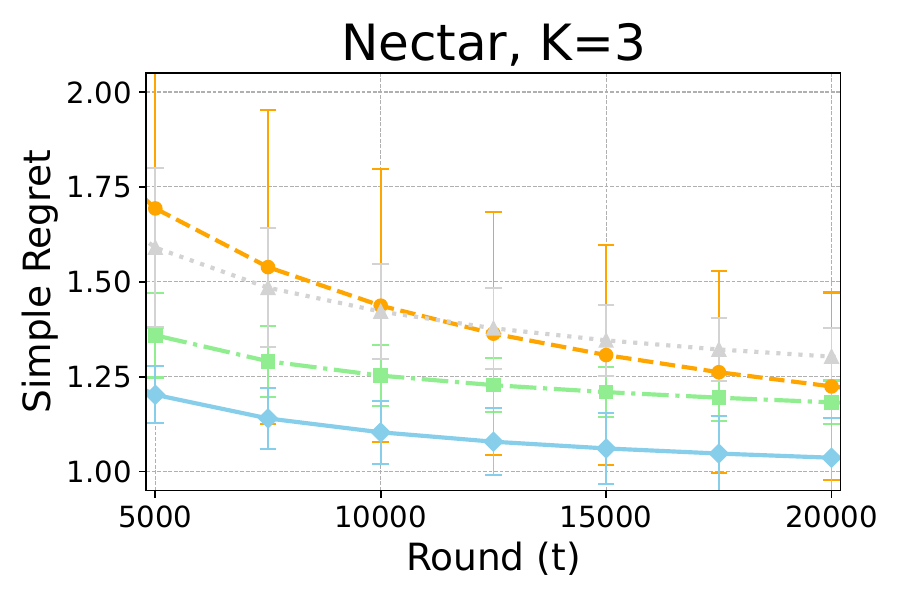}
        \label{fig:nectar_comparison_K=3}
    \end{subfigure}
    \hfill
    \begin{subfigure}[b]{0.325\textwidth}
        \includegraphics[width=\textwidth, trim=0mm 0mm 0mm 0mm, clip]{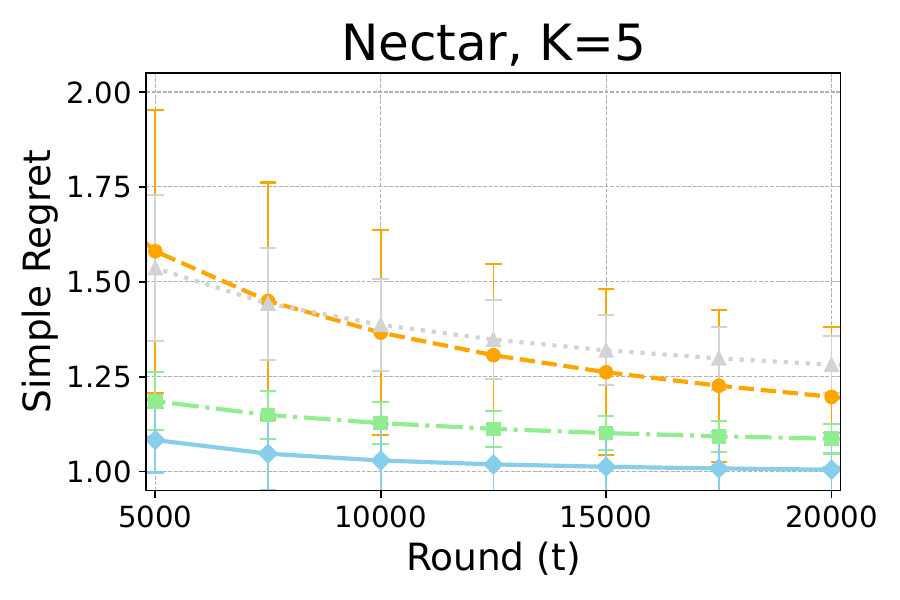}
        \label{fig:nectar_comparison_K=5}
    \end{subfigure}
    \caption{Performance comparisons on the TREC-DL dataset (top row) and the NECTAR dataset (bottom row) for varying values of $K = 2$, $3$, and $5$.}
    \label{fig:app_llm_experiment_comparison_trec}
\end{figure*}
\textbf{Performance measure.}
We measure the realized regret as in the synthetic experiment.

\textbf{Results.}
We present performance comparisons in Figure~\ref{fig:app_llm_experiment_comparison_trec}.
Our algorithm, \AlgName{}, consistently outperforms all baselines by a significant margin.
As in the synthetic data experiments, the suboptimality gap for all methods decreases as $K$ increases.
Notably, \texttt{DopeWolfe} performs particularly poorly on the TREC-DL dataset.
This may be attributed to the use of a small sampling size $R$, which is insufficient compared to the full subset space of size $\binom{N}{K} = \mathcal{O}(N^K) \gg 1000 \geq R$.
This result highlights an important practical limitation of \texttt{DopeWolfe}:  despite its use of approximate optimization to reduce runtime, the method still depends on combinatorial sampling to perform well, which becomes computationally infeasible in large-scale settings.
In contrast, our algorithm, \AlgName{}, achieves consistently strong performance while incurring only polynomial computational cost, demonstrating superior scalability and practicality for real-world deployment.

Table~\ref{tab:LLM_assortment_size} reports the actual assortment size $|S_t|$ selected by \AlgName{} on both datasets.
In the TREC-DL experiment, $|S_t|$ is nearly equal to $K$ for all values of $K$, as the number of available actions is large ($N = 100$).
In contrast, in the NECTAR experiment, where the number of available actions is much smaller ($N = 7$), the actual assortment size $|S_t|$ is often smaller than $K$, especially when $K = N$.
This reduction occurs because the limited action space constrains the potential informativeness of larger assortments—for example, it becomes difficult to achieve high average uncertainty when there are too few actions to choose from.

\begin{table*}[h!]
\centering
    \begin{tabular}{cl|rrrr}
    \toprule
    \multicolumn{2}{c|}{$K$}     & 2 & 3 & 5 & 7    \\
    \midrule
    TREC-DL dataset, &$\!\!\!\!\!|S_t|$  & 2.00 & 3.00 & 5.00 & 6.98  \\
    NECTAR dataset, &$\!\!\!\!\!|S_t|$  & 2.00 & 2.99 & 4.45 & 4.76  \\
    \bottomrule
    \end{tabular}
\caption{Assortment size $|S_t|$ of \AlgName{} with varying $K$ in the real-world dataset experiment}
\label{tab:LLM_assortment_size}
\end{table*}

\end{document}